\documentclass[11pt]{article}

\usepackage{setspace}
\usepackage[final]{acl}

\usepackage{times}
\usepackage{latexsym}
\usepackage{balance}

\usepackage[T1]{fontenc}

\usepackage[utf8]{inputenc}

\usepackage{microtype}

\usepackage{inconsolata}

\usepackage{graphicx}

\usepackage{xcolor} 
\usepackage{amsmath}
\usepackage{amsfonts}
\usepackage{amssymb}
\usepackage{booktabs}
\usepackage{subfigure}
\usepackage{multirow}
\usepackage{hyperref}
\usepackage{siunitx}

\usepackage{enumitem}
\setlist{leftmargin=5.5mm}

\usepackage{colortbl}

\newcommand{\note}[1]{}
\definecolor{grey}{RGB}{200,200,200}
\definecolor{forestgreen}{RGB}{34,139,34}

\long\def\shorten#1{}

\usepackage{colortbl}

\let\svthefootnote\thefootnote
\newcommand\freefootnote[1]{%
    \let\thefootnote\relax%
    \footnotetext{#1}%
    \let\thefootnote\svthefootnote%
}

\newcommand{\nd}{\phantom{$^{\dagger}$}}

\usepackage{enumitem}

\usepackage[ruled,vlined]{algorithm2e}
\usepackage{amsmath}
\title{Layer-wise Target Propagation: Efficient Component Attribution through Target Centric Propagation}

\author{
\textbf{Lasse Jantsch\textsuperscript{*}}
\textbf{Dong-Jae Koh\textsuperscript{*}}
\textbf{Seonghyeon Lee$^{\dagger}$}
\textbf{Young-Kyoon Suh$^{\dagger}$}
\\
School of Computer Science and Engineering, Kyungpook National University, South Korea
\\
\texttt{\{lassejantsch, djkoh, sh0416, yksuh\}@knu.ac.kr}
}

\begin{document}
\maketitle
\freefootnote{\textsuperscript{*} Equal Contributions}
\freefootnote{$^{\dagger}$ Corresponding authors}
\begin{abstract}
Understanding the internal mechanisms of transformer-based large language models (LLMs) is crucial for their reliable deployment and effective operation. While recent efforts have yielded a plethora of attribution methods attempting to balance faithfulness and computational efficiency, dense component attribution remains prohibitively expensive. In this work, we introduce \emph{Layer-wise Target Propagation} (LTP), a novel framework that faithfully traces information flow on the frozen transformer in one forward and one backward pass without requiring counterfactual examples. LTP analytically decomposes and linearizes the computational structure of the Transformers into distinct pathways along which it propagates a targeted unembedding vector to receive the effective representation at each residual position. This target-centric propagation achieves $O(1)$ time complexity with respect to the number of model components, scaling to long input sequences and dense component attribution. Extensive experiments on standard interpretability benchmarks demonstrate that LTP achieves state-of-the-art faithfulness and unprecedented efficiency compared to existing baselines.
\end{abstract}
\section{Introduction}~\label{sec:introduction}
Transformer-based large language models (LLMs) are the dominant model paradigm in AI. However, their internal mechanisms remain poorly understood. A central challenge lies in \emph{attribution}~\citep{pmlr-v70-sundararajan17a}, identifying which components are responsible for a model's predictions.

While a plethora of attribution approaches have been applied to LLMs~\cite{Ferrando2024APO}, these methods often fall short in real-world settings. 
Along with this research line, fully decomposing the computational graph of the transformers, i.e., \textit{decomposition-based approaches}, becomes a promising research direction in the attribution \hbox{research} realm~\citep{Hong2025DePassUF}.

Motivated by the limited efficiency and inaccurate approximations in previous work, we propose \emph{Layer-wise Target Propagation} (LTP), a novel decomposition-based framework for efficient input and component attribution. Specifically, LTP efficiently computes the attribution through a single forward and backward pass by propagating the unembedding vector of the target token. As our method leverages general forward and backward passes, it can be universally applied to \emph{any} transformer variant. In this paper, we focus on the SwiGLU transformer, as most modern competitive LLMs adopt this architecture~\citep{Touvron2023Llama2O}. Figure \ref{fig:dpa_framework} shows an overview of our approach.

Our experiments demonstrate the superior faithfulness and efficiency of LTP on standard tasks for both input and component attribution. 
Our contributions are summarized as follows:

\begin{figure*}[!t]
    \centering
    \includegraphics[width=1.03\textwidth]{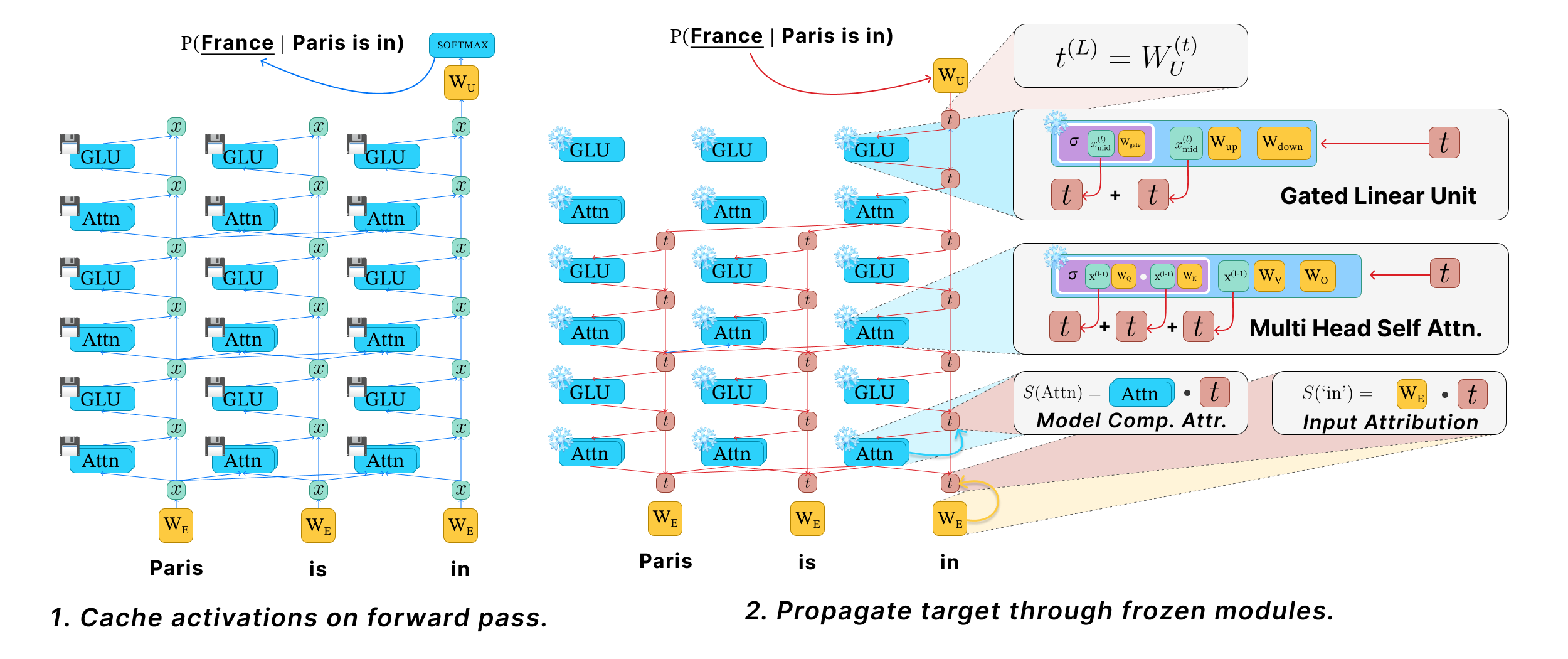}
    \caption{Overview of the \emph{Layer-wise Target Propagation} (LTP) framework for efficient input and component attribution. \emph{LTP} operates in two stages: (1) \emph{One-pass forward execution}, in which the model processes the input once while caching activations required for the back propagation; and (2) \emph{top-down target propagation}, where the unembedding vector of the targeted token is recursively propagated backward through Transformer modules to identify an effective target at each residual position.~\label{fig:dpa_framework}}
    \vspace{-1ex}
\end{figure*}

\vspace{-1mm}
\begin{itemize}
    \item We introduce \emph{Layer-wise Target Propagation} (LTP), a novel decomposition-based framework that efficiently assigns logit scores in a single forward and backward pass, without requiring counterfactual examples.
    \vspace{-1mm}
    \item By leveraging the bilinear structure of the SwiGLU Transformers, we show that the computational structure is decomposable into distinct control and content pathways.
    \vspace{-1mm}
    \item Through extensive experiments on standard interpretability tasks, we demonstrate that LTP achieves superior faithfulness and efficiency compared to state-of-the-art baselines, effectively identifying causal mechanisms.
\end{itemize}
All artifacts are publicly available on GitHub.\footnote{\url{https://github.com/lab-paper-code/LTP}}
\section{Related Work}~\label{sec:related_work}
Attribution research aims to localize the sources of model behavior by identifying responsible inputs or internal components~\citep{Ferrando2024APO}.

\paragraph{Input attribution.}
A first family of input attribution methods derives importance from model gradients and their propagated variants. They treat the first-order Taylor expansion as a sensitivity signal~\citep{Simonyan2013DeepIC, Denil2014ExtractionOS, Li2015VisualizingAU}, as in \textit{Input x Gradient}~\citep{Denil2014ExtractionOS}, which scales the gradient by the embedding to account for input magnitude. As raw gradients suffer from saturation and \hbox{shattering}~\citep{shrikumar2019learning, balduzzi2018shattered}, later work \hbox{denoises} the gradients through \textit{Integrated Gradients}~\citep{pmlr-v70-sundararajan17a} or \textit{SmoothGrad}~\citep{smilkov2017smoothgrad}. 

A second family traces how the signal flows through attention. Raw attention weights were the earliest such signal, but proved unreliable as indicators~\citep{Jain2019AttentionIN}. Subsequent work made the signal more faithful by weighting attention with value-vector norms~\citep{kobayashi-etal-2020-attention, kobayashi-etal-2021-incorporating} and composing it across layers via rollout~\citep{abnar-zuidema-2020-quantifying, ferrando-etal-2022-measuring, modarressi-etal-2022-globenc}. Building on this trajectory, decomposition-based methods analyze the model's linear structure, yielding faithful attribution of attention-mediated effects~\citep{modarressi-etal-2023-decompx, Kobayashi2023AnalyzingFB, Hong2025DePassUF}.
Our method adopts this decomposition view, but rather than propagating the decomposed states forward, it propagates the target backward through the linearized structure.

\paragraph{Component-level attribution.}
Attribution can also target internal components rather than inputs. \textit{Direct Logit Attribution} (DLA)~\citep{elhage2021mathematical, nostalgebraist2020logit} exploits the additive residual stream to read off a component's \emph{direct} contribution to the output logit and applies to both attention heads~\citep{Ferrando2023ExplainingHT} and neurons~\citep{Geva2022TransformerFL}. DLA, however, does not capture the indirect effects of downstream layers, thereby reducing interpretability in early layers.

Activation patching removes this blind spot by overwriting a component and measuring the resulting output change, thereby capturing both direct and indirect effects through counterfactual forward passes~\citep{Vig2020InvestigatingGB, Meng2022LocatingAE, Wang2022InterpretabilityIT, Zhang2023TowardsBP}. This faithfulness comes at a steep cost due to the separate forward pass per component. Our method targets exactly this gap, recovering the direct and indirect effects that DLA misses through a single backward pass, without the per-component forward passes that make patching intractable.

These component-level signals, in turn, are the building blocks of automated circuit discovery, which searches for the component subsets responsible for a behavior. Existing pipelines instantiate this search on top of causal intervention~\citep{Conmy2023TowardsAC}, gradients~\citep{nanda2023attribution, syed-etal-2024-attribution}, LRP~\citep{jafari_relp_2025}, and contextual mixing~\citep{ferrando-voita-2024-information}.
\section{Background}~\label{sec:transformer_architecture}
In this section, we review the recent variants of Transformer components and establish the notation used throughout the paper. A more general overview is provided in Appendix~\ref{app:transformer_architecture}.

\subsection{Multi-Head Self-Attention (MHSA)}
Different from the original MHSA module~\citep{Radford2019LanguageMA}, recent variants such as the LLaMA2 architecture~\citep{Touvron2023Llama2O,grattafiori2024llama3herdmodels} employ different normalization and positional encoding techniques.

First, given an input embedding sequence $\mathbf{X} \in \mathbb{R}^{T \times d}$, the modern MHSA module first normalizes the embedding sequence using RMSNorm~\citep{Zhang2019RootMS}. 
These normalized embeddings $\tilde{\mathbf{X}}$ are projected into queries, keys, and values across different heads:
\begin{equation*}
    \tilde{\mathbf{X}}\mathbf{W}^{(h)}_Q, \tilde{\mathbf{X}}\mathbf{W}^{(h)}_K, \tilde{\mathbf{X}}\mathbf{W}^{(h)}_V
\end{equation*}
where $\mathbf{W}^{(h)}_Q, \mathbf{W}^{(h)}_K, \mathbf{W}^{(h)}_V$ are the query, key, and value projections for head $h$.
The attention scores are computed using the scaled dot product with the query and key, followed by the softmax function:
\[
\resizebox{0.96\linewidth}{!}{%
$\displaystyle
\tilde{\alpha}^{(h)} = \mathrm{SoftMax}\!\left(
\frac{
R_Q\left(\tilde{\mathbf{X}}\mathbf{W}^{(h)}_Q\right)\,
R_K\left(\tilde{\mathbf{X}}\mathbf{W}^{(h)}_K\right)^\top
}{\sqrt{d_h}}
\right),
$}
\]
where $R(\cdot)$ denotes the application of RoPE~\citep{Su2021RoFormerET}. These attention scores are used for the weighted sum of value vectors $V$, and the output of MHSA is formed by concatenating head-wise outputs and projecting them to the embedding dimension:
\[
\text{MHSA}(\mathbf{X}) = \mathrm{Concat} \big(\tilde{\mathbf{X}} \mathbf{W}^{(i)}_V\big)\mathbf{W}_O,
\]
where $\mathbf{W}_O$ is the projection matrix. The output of the MHSA module is added to the residual stream.

\subsection{Feed-forward Network (FFN)}
In modern architecture, the FFN uses the SwiGLU activation, which separates computation into a gating and a value pathway. Given a normalized input $\tilde{\mathbf{X}}$, GLU activation is defined as:
\[
\text{GLU}(\mathbf{X}) =
\Big(
\sigma(\tilde{\mathbf{X}}\mathbf{W}_G)
\odot
(\tilde{\mathbf{X}}\mathbf{W}_U)
\Big)\mathbf{W}_D,
\]
\noindent where $\sigma(\cdot)$ denotes the SiLU (Swish) activation, $\odot$ is element-wise multiplication, and
$\mathbf{W}_G, \mathbf{W}_U \in \mathbb{R}^{d \times d_{\text{ffn}}}$ and
$\mathbf{W}_D \in \mathbb{R}^{d_{\text{ffn}} \times d}$ are the gate, up, and down projection matrices, respectively.

\subsection{Decomposition of Transformer Modules} 
One interesting finding is the linear separability of the transformer blocks~\citep{modarressi-etal-2023-decompx,kobayashi-etal-2021-incorporating}, allowing for decomposition into independent model component contributions; attention heads and neurons for MHSA and GLU, respectively. For MHSA, the output can be expressed as the sum of head-wise contributions:
\[
\text{MHSA}(\mathbf{X}) =
\sum_{h=1}^{H}
\alpha^{(h)}
\big(\tilde{\mathbf{X}}\mathbf{W}_V^{(h)}\big)
\mathbf{W}_O^{(h)},
\]
where $\mathbf{W}_V^{(h)}$ and $\mathbf{W}_O^{(h)}$ denote the value and output projections of head $h$. Similarly, the GLU output can be decomposed into contributions from individual neurons:
\[
\text{GLU}(\mathbf{X}) =
\sum_{i=1}^{d_{\text{ffn}}}
\Big(
\sigma(\tilde{\mathbf{X}}\mathbf{w}_G^{(i)})
\odot
(\tilde{\mathbf{X}}\mathbf{w}_U^{(i)})
\Big)\mathbf{w}_D^{(i)},
\]
\noindent where $\mathbf{w}_G^{(i)}$ and $\mathbf{w}_U^{(i)}$ denote the $i$-th columns of $\mathbf{W}_G$ and $\mathbf{W}_U$, and $\mathbf{w}_D^{(i)}$ denotes the $i$-th row of $\mathbf{W}_D$. Each term corresponds to the contribution of a single GLU neuron to the residual stream.

\section{Layer-wise Target Propagation}~\label{sec:dual_path_attribution}
In this section, we formalize the dense component attribution problem and propose \emph{Layer-wise Target Propagation} (LTP) as our solution.

\subsection{Problem Statement}~\label{sec:problem_statement}
Our objective is to \emph{quantify how much a given model component} $\mathbf{x}$ \emph{contributes to a target model output $\mathbf{t}$}. Formally, we seek to assign an attribution score $S_t(\mathbf{x})$ that measures the relevance of $\mathbf{x}$ to a target token prediction $\mathbf{t}$. Let $\mathbf{x}^{(0)}_i \in \mathbb{R}^d$ denote the embedding vector at token position $i$, and let $\mathbf{t}_j \in \mathbb{R}^d$ be the $j$-th row of the unembedding matrix $W_{UE}$, corresponding to the direction in latent space that increases the logit $logit^{(j)}_i$ of token $t^{(j)} \in \mathcal{V}$ at position $i$. Owing to the additive structure of the residual stream, $\mathbf{x}^{(0)}_i$ contributes directly to the output logit, yielding the direct attribution score:
\[
S_{t,\text{direct}}(\mathbf{x}^{(0)}_i) = (\mathbf{x}^{(0)}_i)^\top \mathbf{t}_j.
\]
While each embedding at position $i'$ contributes directly to the logit at its own position, it has no direct contribution to the logits at later positions $i > i'$; such cross-position influence is mediated entirely through downstream layers~\citep{Belrose2023ElicitingLP}.
The full attribution can be formalized as:
\[
S_t(\mathbf{X}^{(0)}) = f(\mathbf{x}^{(0)}_i)^\top \mathbf{t}_j,
\]

\noindent where $f(\cdot)$ denotes the downstream processing of $\mathbf{x}^{(0)}_i$ with an otherwise frozen network. While faithful~\citep{Hong2025DePassUF}, this approach requires a separate forward pass per component to evaluate $f(\mathbf{x}^{(0)}_i)$, making dense attribution prohibitively expensive.




\subsection{Target Propagation Framework}~\label{sec:target_propagation_framework}
To overcome this limitation, we invert the attribution perspective: rather than passing each component forward, we propagate the target backward. Specifically, we reformulate the attribution score as:
\[
S_t(\mathbf{x}^{(0)}_i) = (\mathbf{x}^{(0)}_i)^\top f^{-1}(\mathbf{t}_j) = (\mathbf{x}^{(0)}_i)^\top \mathbf{t}^{(0)}_j,
\]

where $\mathbf{t}^{(0)}_j$ denotes the \emph{effective target} at layer $0$. This effective target is computed recursively via:
\begin{align*}
\mathbf{t}^{(l)}_{\text{mid}} &= \mathbf{t}^{(l+1)} + g^{(l)}_{\text{GLU}}\big(\mathbf{t}^{(l+1)}\big), \\
\mathbf{t}^{(l)}_j &= \mathbf{t}^{(l)}_{\text{mid}} + g^{(l)}_{\text{MHSA}}\big(\mathbf{t}^{(l)}_{\text{mid}}\big),
\end{align*}
\noindent where $g^{(l)}_{\text{GLU}}(\cdot)$ and $g^{(l)}_{\text{MHSA}}(\cdot)$ denote linearized inverse transformations of the $\text{GLU}^{(l)}$ and $\text{MHSA}^{(l)}$ blocks, respectively. Under this formulation, attribution requires only a single forward pass to cache activations and a single backward pass to propagate the effective target.

LTP implements this framework by linearizing and inverting MHSA and GLU via their constituent pathways while keeping the rest of the module frozen. To prevent the explosion of the target norm and sustain completeness\footnote{While LTP generally supports completeness, our choice of softmax linearization breaks it for the sake of a robust gradient.}, each inversion is further regularized by a weighting parameter $\mu$. The following sections instantiate $g^{(l)}_{\text{GLU}}(\cdot)$ and $g^{(l)}_{\text{MHSA}}(\cdot)$ at the neuron and attention-head level. Detailed derivations and their complexity analyses are provided in Appendix~\ref{app:dpa_derivations} and \ref{app:complexity_analysis}, respectively.

\subsection{Neuron Target Propagation}
Consider a GLU neuron $n$ with activation $\alpha^{(n)}$. To propagate the effective target backward through the GLU, we approximate the inverse transformation $\mathbf{g}_{\text{GLU}}^{(n)}(\cdot)$ as a weighted combination of the up- and gate-projection paths, assuming frozen activations:
\[
\mathbf{g}_{\text{GLU}}^{(n)}(\mathbf{t}) 
= \mu_{\text{up}} \cdot \mathbf{g}_{\text{up}}^{(n)}(\mathbf{t}) 
+ \mu_{\text{gate}} \cdot \mathbf{g}_{\text{gate}}^{(n)}(\mathbf{t}),
\]
where $\mu_{\text{up}} + \mu_{\text{gate}} = 1$.

\paragraph{Up-projection.}
With frozen activations, the \hbox{up-projection-pathway} is linear and corresponds to the \hbox{composition} of the up- and down-projection weights,
$\mathbf{W}_{UD}^{(n)} = \mathbf{w}^{(n)}_U \mathbf{w}^{(n)}_D$.
The propagated \hbox{target} is given by:
\vspace{-1ex}
\[
\mathbf{g}_{\text{up}}^{(n)}(\mathbf{t}) 
= \underbrace{\frac{\mathbf{w}_{U}^{(n)} \odot \boldsymbol{\gamma}}{\sigma}}_{\text{effective weight } \tilde{\mathbf{w}}_{U}^{(n)}}
\cdot \alpha^{(n)} \cdot 
\underbrace{\big((\mathbf{w}_{D}^{(n)})^\top \mathbf{t}\big)}_{\text{gradient scalar } \lambda_{\text{GLU}}},
\]
where $\boldsymbol{\gamma}$ and $\sigma$ denote the RMSNorm parameters. For clarity, we refer to such normalized weights as \emph{effective weights} $\tilde{\mathbf{w}}$.

\paragraph{Gate-projection.}
Due to nonlinearity, the gate-projection pathway must be linearized to be inverted. We approximate the local gradient using the ratio $\alpha^{(n)} / s^{(n)}$, where $s^{(n)}$ denotes the pre-activation value:
\vspace{-2ex}
\[
\mathbf{g}_{\text{gate}}^{(n)}(\mathbf{t}) 
= \tilde{\mathbf{w}}_{G}^{(n)} \cdot \frac{\alpha^{(n)}}{s^{(n)}} \cdot \lambda_{\text{GLU}}.
\]

\subsection{Attention Head Target Propagation}
For an attention head $h$, target propagation is more involved due to the token mixing \hbox{softmax} activation. We decompose the inverse MHSA transformation into contributions from the query, key, and value paths and combine them via weighted aggregation:
\[
\mathbf{g}_{\text{head}}^{(h)}(\mathbf{t}) 
= \mu_Q \cdot \mathbf{g}_{Q}^{(h)}(\mathbf{t}) 
+ \mu_K \cdot \mathbf{g}_{K}^{(h)}(\mathbf{t}) 
+ \mu_V \cdot \mathbf{g}_{V}^{(h)}(\mathbf{t}),
\]
where $\mu_Q + \mu_K + \mu_V = 1$.

\paragraph{Value path.}
The value path is linear, given fixed attention weights, and is defined as:
\[
\mathbf{g}_{V}^{(h)}(\mathbf{t}) 
= \tilde{\mathbf{W}}_{V}^{(h)} \cdot 
\underbrace{(\mathbf{W}_{O}^{(h)})^\top \mathbf{t}}_{\text{grad. vec. } \boldsymbol{\lambda}_{\text{MHSA}}}
\cdot \,\, \boldsymbol{\alpha}^{(h)}.
\]

\paragraph{Query and key paths.}
Propagating targets through the query and key paths requires tracing gradients through the softmax operation. Unlike GLU activations, softmax outputs depend jointly on all attention scores and remain \hbox{non-zero} even for zero inputs, making naive linearization unstable. Following \citet{achtibat2024attnlrp}, we adopt a Taylor-based decomposition of the softmax, which provides a stable—though \hbox{non-conservative-approximation}:
\vspace{-1.5ex}
\begin{align*}
\mathbf{g}_{Q,i}^{(h)}(\mathbf{t}) 
&= \tilde{\mathbf{W}}_{Q}^{(h)} 
\sum_{j \le i} \delta_{ij} \cdot 
R^{-1}_{Q,i} 
\left(\frac{R_{K,j}(\tilde{\mathbf{x}}_j \mathbf{W}_K^{(h)})}{\sqrt{d_h}}\right)^\top, \\
\mathbf{g}_{K,j}^{(h)}(\mathbf{t}) 
&= \tilde{\mathbf{W}}_{K}^{(h)} 
\sum_{i \ge j} \delta_{ij} \cdot 
R^{-1}_{K,j} 
\left(\frac{R_{Q,i}(\tilde{\mathbf{x}}_i \mathbf{W}_Q^{(h)})}{\sqrt{d_h}}\right)^\top,
\end{align*}
where the scalar interaction term $\delta_{ij}$ captures the sensitivity of the softmax output:
\[
\delta_{ij} 
= \alpha_{ij} \big(\tilde{\mathbf{x}}_j \mathbf{W}_V^{(h)} - \boldsymbol{\mu}^{(h)}_i \big)
\bullet \boldsymbol{\lambda}_{\text{MHSA}}.
\]
Here, $\boldsymbol{\mu}^{(h)}_i$ denotes the expected intermediate activation vector of head $h$ at token position $i$.

\begin{figure*}[!t]
    \centering
        \includegraphics[width=1.9\columnwidth]{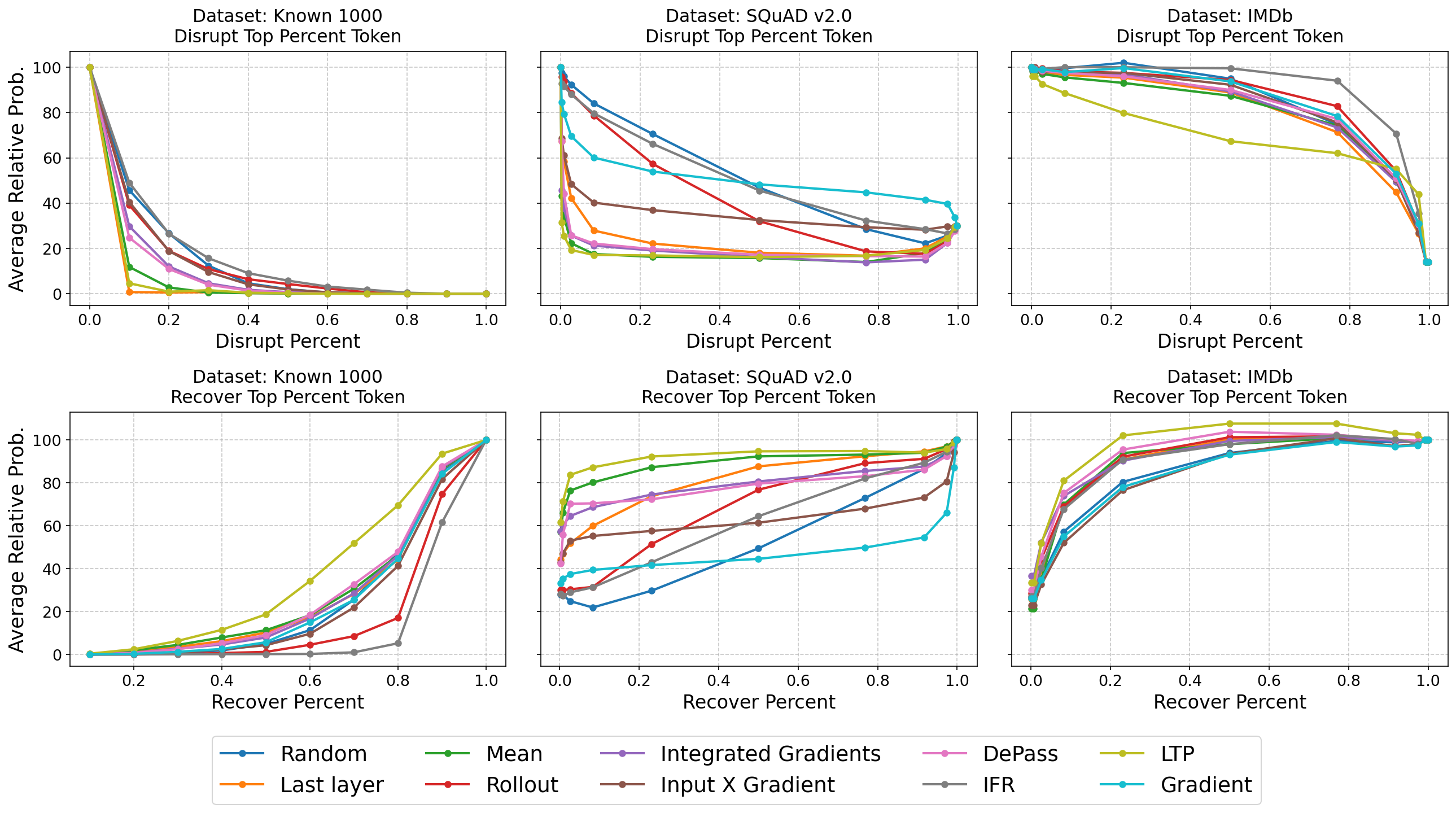}
    \vspace{-.1in}
    \caption{Performance comparison with baseline attribution methods on \textbf{Llama-3.1-8B-Instruct} under different top input token masking strategies. Our approach shows \emph{lower} Top-$k$ disruption and \emph{higher} Top-$k$ recovery, indicating more accurate identification of important components.~\label{fig:input_masking}}
    \vspace{-1ex}
\end{figure*}

\begin{table*}[!t]
    \centering
    \footnotesize
    \renewcommand{\arraystretch}{0.65}

\resizebox{\linewidth}{!}{%
    \begin{tabular}{l@{\hskip 14pt}ccc@{\hskip 14pt}ccc@{\hskip 14pt}ccc}
        \toprule
        \multirow{2.5}{*}{\textbf{Methods}}
        & \multicolumn{3}{c}{\textbf{Known 1000}}
        & \multicolumn{3}{c}{\textbf{SQuAD v2.0}}
        & \multicolumn{3}{c}{\textbf{IMDb}} \\
        \cmidrule(lr){2-4}\cmidrule(lr){5-7}\cmidrule(lr){8-10}
        & dis.\,$\downarrow$ & rec.\,$\uparrow$ & total\,$\uparrow$
        & dis.\,$\downarrow$ & rec.\,$\uparrow$ & total\,$\uparrow$
        & dis.\,$\downarrow$ & rec.\,$\uparrow$ & total\,$\uparrow$ \\
        \midrule
        Random
        & 14.22\nd & 22.78\nd & 8.56\nd
        & 50.64\nd & 52.39\nd & 1.75\nd
        & 84.44\nd & 85.93\nd & 1.49\nd \\
        \midrule
        \multicolumn{10}{c}{\textit{Attention-based}} \\
        \midrule
        Last layer
        & \textbf{5.26}\nd & 24.83$^{\dagger}$ & 19.57$^{\dagger}$
        & 21.96\nd & 82.04\nd & 60.08\nd
        & 79.68$^{\dagger}$ & 91.57\nd & 11.89$^{\dagger}$ \\
        Mean
        & 6.57\nd & 25.78\nd & 19.21\nd
        & \textbf{17.06}\nd & 89.46$^{\dagger}$ & 72.40$^{\dagger}$
        & 79.89\nd & 91.42\nd & 11.53\nd \\
        Rollout
        & 13.26\nd & 15.72\nd & 2.46\nd
        & 40.58\nd & 69.43\nd & 28.85\nd
        & 85.33\nd & 92.47\nd & 7.14\nd \\
        \midrule
        \multicolumn{10}{c}{\textit{Gradient-based}} \\
        \midrule
        Gradient
        & 12.14\nd & 22.94\nd & 10.80\nd
        & 49.95\nd & 46.68\nd & -3.27\nd
        & 84.52\nd & 84.58\nd & 0.06\nd \\
        Input X Gradient
        & 12.58\nd & 21.27\nd & 8.69\nd
        & 34.07\nd & 63.22\nd & 29.15\nd
        & 82.65\nd & 84.34\nd & 1.69\nd \\
        Integrated Gradients
        & 9.92\nd & 24.49\nd & 14.57\nd
        & 17.92\nd & 79.53\nd & 61.61\nd
        & 81.29\nd & 92.41$^{\dagger}$ & 11.12\nd \\
        \midrule
        \multicolumn{10}{c}{\textit{Contextual Mixing-based}} \\
        \midrule
        IFR
        & 16.15\nd & 11.88\nd & -4.27\nd
        & 50.16\nd & 62.51\nd & 12.35\nd
        & 91.63\nd & 91.12\nd & -0.51\nd \\
        \midrule
        \multicolumn{10}{c}{\textit{Decomposition-based}} \\
        \midrule
        DePass
        & 9.17\nd & 25.45\nd & 16.28\nd
        & 19.29\nd & 78.45\nd & 59.16\nd
        & 85.49\nd & 69.65\nd & -15.84\nd \\

        \textbf{LTP (Ours)}
        & 5.79$^{\dagger}$ & \textbf{33.88}\nd & \textbf{28.09}\nd
        & 17.73$^{\dagger}$ & \textbf{92.51}\nd & \textbf{74.78}\nd
        & \textbf{69.34}\nd & \textbf{99.51}\nd & \textbf{30.17}\nd \\
        \bottomrule
    \end{tabular}
}
    \vspace{-2mm}
    \caption{
    Faithfulness evaluation of crucial token attribution on \textbf{Llama-3.1-8B-Instruct} across Known 1000 (factual knowledge), SQuAD v2.0 (reading comprehension), and IMDb (sentiment analysis).
    Lower disruption (\textit{dis.}) and higher recovery (\textit{rec.}) indicate more faithful token localization; \textit{total} denotes their aggregate AUC score.
    \textbf{Bold} indicates the best result, and $^{\dagger}$ the second best.
    LTP consistently achieves the highest faithfulness across datasets.
    }
    \label{tab:input_faithfulness}
    \vspace{-1mm}
\end{table*}

\section{Experiments}~\label{sec:experiments}
We assess the effectiveness of LTP under two \hbox{research} questions:
\vspace{-1mm}
\begin{itemize}
    \item RQ1. How \emph{faithfully} does LTP attribute input tokens and internal component importance?
    \vspace{-2mm}
    \item RQ2. How \emph{effective} are balanced hyperparameter selections for gaining optimal performance?
\end{itemize}
\vspace{-1mm}
To answer these questions, we conduct \hbox{attribution} \emph{faithfulness} and \emph{sensitivity} experiments.

\subsection{Attribution Faithfulness}

We quantify how well LTP identifies components that are causally relevant to a model’s prediction.

\subsubsection{Experimental setup}

\paragraph{Datasets and Models.}
We evaluate LTP on several modern LLM checkpoints, \hbox{including} Llama-3.1-8B-Instruct~\citep{grattafiori2024llama3herdmodels}, \hbox{Llama-2-7B-Chat}~\citep{Touvron2023Llama2O}, \hbox{Qwen3-4B-Instruct}~\citep{yang2025qwen3technicalreport}, Mistral-7B-Instruct~\citep{Jiang2023Mistral7}, Qwen2.5-32B-Instruct~\citep{Yang2024Qwen25TR}, and GPT-2 Large~\citep{Radford2019LanguageMA}. Three representative datasets are used to cover different attribution scenarios: 
(1)~\textit{Known 1000}~\citep{Meng2022LocatingAE} for factual sequence completion,
(2)~\textit{SQuAD v2.0}~\citep{Rajpurkar2018KnowWY} for context-supported question answering, and
(3)~\textit{IMDb}~\citep{maas-EtAl:2011:ACL-HLT2011} for sentiment analysis. We also leverage the Indirect Object Identification (IOI) task~\citep{Wang2022InterpretabilityIT}, a widely used benchmark for circuit discovery in the component attribution experiment. See Appendix \ref{app:dataset_configurations} for additional details.

\paragraph{Metrics.}
Following prior work, we adopt two complementary metrics: \emph{disruption} and \emph{recovery}. Both replace the embeddings of a selected set of tokens with zero vectors and measure the resulting change in the probability of the correct next token.
For \emph{disruption}, we zero out the top-$K$ tokens judged most important and keep the rest intact; a faithful attribution should then cause a large drop in the probability, since the tokens driving the prediction have been removed.
For \emph{recovery}, we instead zero out all tokens \emph{except} the top-$K$ most important; here, a faithful attribution should preserve the probability since the retained tokens alone should suffice. These mirror the comprehensiveness and sufficiency criteria of \citet{deyoung-etal-2020-eraser}.

\subsubsection{Input Attribution Experiments}

\paragraph{Baselines.}
We compare LTP against a broad range of attribution baselines, including gradient-based methods (\textit{Gradient} at input, \textit{ Input$\times$Gradient}~\citep{Denil2014ExtractionOS}, and \textit{\hbox{Integrated} Gradients}~\citep{pmlr-v70-sundararajan17a}, \hbox{attention-based} methods (\textit{Last-Layer Attention} and \textit{Mean Attention} on the last \hbox{token}, as well as \textit{Attention Rollout}~\citep{abnar-zuidema-2020-quantifying}, the context-mixing method \textit{Information Flow Routes}~\citep{ferrando-voita-2024-information}, and the decomposition-based method \hbox{\textit{DePass}~\citep{Hong2025DePassUF}.}

\paragraph{Implementation details.}
For the Known 1000 dataset, we assess faithfulness by ablating individual input tokens, excluding special tokens. For the SQuAD v2.0 and IMDb datasets, we employ a context- or review-level ablation strategy, in which tokens within the supporting paragraph or review are ablated while the remaining prompt remains intact. This setup mitigates artifacts caused by truncating incomplete sentences immediately preceding the target token, yielding a cleaner faithfulness signal. Additional setup information for reproducibility is located in Appendix \ref{app:reproducibility}.

For token-level attribution, we compute effective targets $\mathbf{t}^{(0)}_i$ by propagating the unembedding vector of the correct answer token backward to the input layer. Token importance is then obtained via the dot product between $\mathbf{t}^{(0)}_i$ and the corresponding embedding vectors.

\begin{figure*}[!tp]
    \centering
        \includegraphics[width=2\columnwidth]{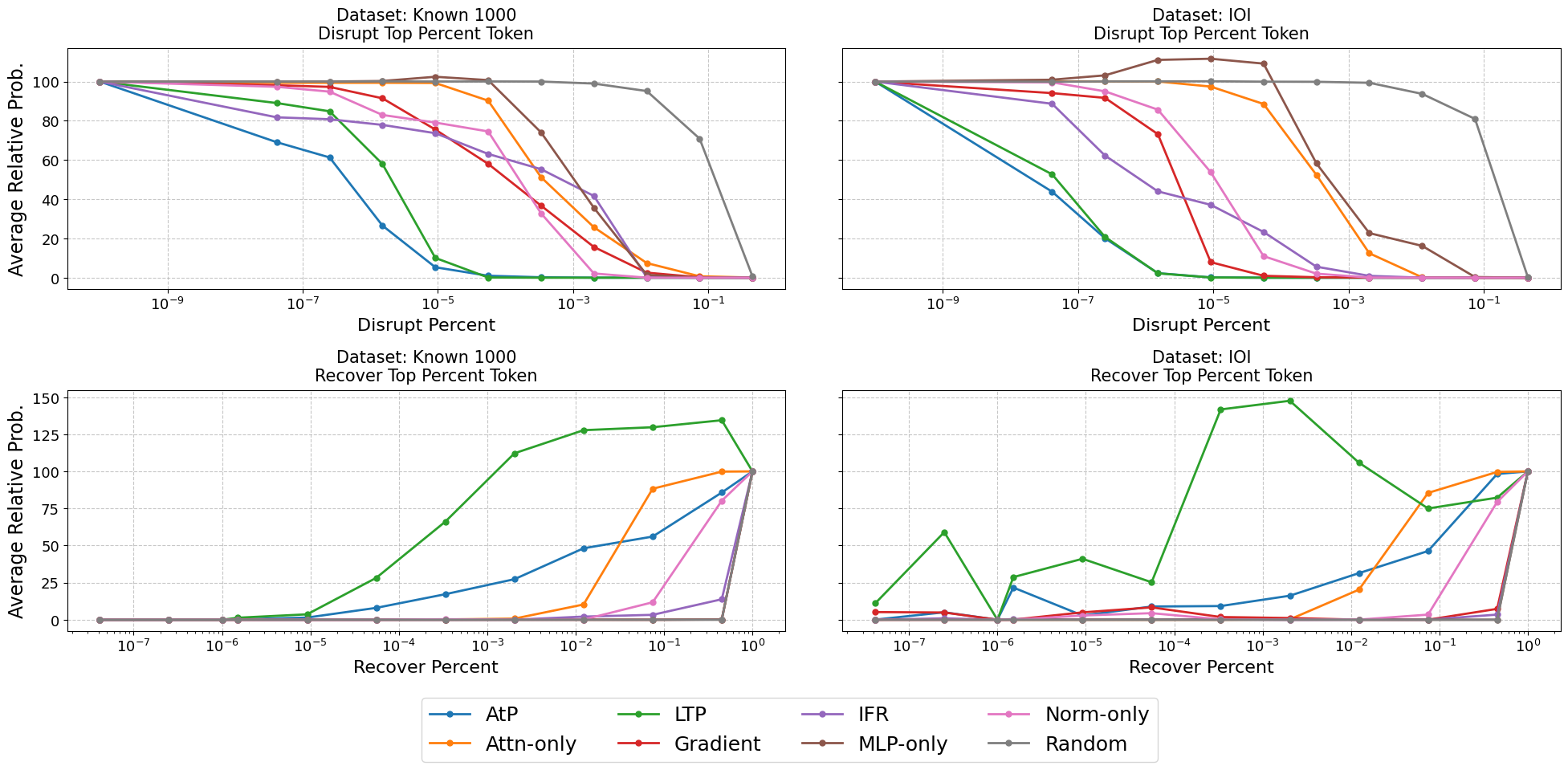}
    \vspace{-2mm}
    \caption{Performance comparison with baseline attribution methods on \textbf{Llama-3.1-8B-Instruct} under different top model component masking strategies. Our approach shows lower disruption and higher recovery, indicating more accurate identification of important components.}~\label{fig:model_masking}
    \vspace{-4mm}
\end{figure*}

\paragraph{Results.}
As shown in Figure~\ref{fig:input_masking} and Table~\ref{tab:input_faithfulness}, LTP attains the highest total faithfulness across all three datasets. LTP is the only method that remains \emph{faithful} across two qualitatively different regimes. On \hbox{\emph{localized}} tasks, where the prediction hinges on a single answer-bearing span (Known 1000, SQuAD v2.0), several baselines are competitive. On the \hbox{\emph{distributed}} IMDb task, where sentiment is spread across many tokens, every competing method degrades sharply while LTP retains a large lead. We attribute this robustness to how LTP decomposes the computation: rather than scoring tokens by attention routing alone, it propagates the target through both the \emph{content} pathway (value and up projections) and the \emph{control} pathway (attention and gate scores), so that evidence dispersed across many tokens is still accumulated rather than lost. Finally, extended results on Llama-2, Qwen3, Qwen2.5, Mistral, and GPT-2 Large (Appendix~\ref{app:extended_results}) reproduce the same ordering, confirming that the gains are not specific to a single backbone.

\subsubsection{Component Attribution Experiment}
We further assess attribution faithfulness at the level of internal model components, focusing on individual attention heads and MLP neurons.

\paragraph{Baselines.}
We compare LTP with activation-based baselines: \textit{Attn-only}, which ranks attention heads by activation magnitude; \textit{MLP-only}, which ranks neurons by activation strength; and \textit{Norm}, which decomposes component contributions using the L2 norm. We also evaluate gradient-based methods, including \textit{Gradient}, \textit{AtP}~\citep{nanda2023attribution}, and contextual mixing-based method, \textit{Information Flow Routes} (IFR)~\citep{ferrando-voita-2024-information}.

\paragraph{Evaluation Setup.}
All attention heads and MLP neurons are ranked according to their attribution scores. We progressively ablate or recover the top-ranked components and measure the resulting change in the model’s prediction confidence. Additional setup information for reproducibility is located in Appendix~\ref{app:reproducibility}.

\begin{table}[!t]
    \centering
    \small
    \setlength{\tabcolsep}{2pt}
    \renewcommand{\arraystretch}{0.9}

    \begin{tabular}{lcccccc}
        \toprule
        \multirow{2.5}{*}{\textbf{Methods}}
        & \multicolumn{3}{c}{\textbf{Known 1000}}
        & \multicolumn{3}{c}{\textbf{IOI}} \\
        \cmidrule(lr){2-4}\cmidrule(lr){5-7}
        & dis.$\downarrow$ & rec.$\uparrow$ & tot.$\uparrow$
        & dis.$\downarrow$ & rec.$\uparrow$ & tot.$\uparrow$ \\
        \midrule

        Random
        & 19.77\nd & 27.55\nd & 7.78\nd
        & 21.85\nd & 27.54\nd & 5.69\nd \\

        \midrule
        \multicolumn{7}{c}{\textit{Baselines}} \\
        \midrule

        Attn-only
        & 0.66\nd & 93.43$^{\dagger}$ & 92.77$^{\dagger}$
        & 0.15\nd & \textbf{93.07}\nd & \textbf{92.92}\nd \\

        MLP-only
        & 0.37\nd & 27.58\nd & 27.21\nd
        & 0.87\nd & 27.53\nd & 26.66\nd \\

        Norm-only
        & 0.06$^{\dagger}$ & 67.25\nd & 67.19\nd
        & 0.01$^{\dagger}$ & 65.02\nd & 65.01\nd \\

        \midrule
        \multicolumn{7}{c}{\textit{Gradient-based}} \\
        \midrule

        Gradient
        & 0.28\nd & 27.62\nd & 27.34\nd
        & \textbf{0.00}\nd & 30.95\nd & 30.95\nd \\

        AtP
        & \textbf{0.00}\nd & 81.39\nd & 81.39\nd
        & \textbf{0.00}\nd & 84.39\nd & 84.39\nd \\

        \midrule
        \multicolumn{7}{c}{\textit{Contextual Mixing-based}} \\
        \midrule

        IFR
        & 0.32\nd & 34.72\nd & 34.40\nd
        & 0.02\nd & 29.15\nd & 29.13\nd \\

        \midrule
        \multicolumn{7}{c}{\textit{Decomposition-based}} \\
        \midrule

        \textbf{LTP (Ours)}
        & \textbf{0.00}\nd & \textbf{123.50}\nd & \textbf{123.50}\nd
        & \textbf{0.00}\nd & 86.87$^{\dagger}$ & 86.87$^{\dagger}$ \\

        \bottomrule
    \end{tabular}
    \vspace{-2mm}
    \caption{
    Component-level attribution performance on \textbf{Llama-3.1-8B-Instruct}.
    Lower disruption and higher recovery indicate more faithful localization.
    Bold denotes best; $^{\dagger}$ denotes second best.
    }
    \label{tab:component_faithfulness}
    \vspace{-6mm}
\end{table}

\paragraph{Results.}  
As shown in Figure~\ref{fig:model_masking} and Table~\ref{tab:component_faithfulness}, LTP achieves the highest total faithfulness on both IOI and Known 1000. Although \hbox{Attn-only} \hbox{obtains} slightly higher aggregate AUC on IOI, LTP's \hbox{S-shaped} curve isolates the circuit much earlier (around 0.001\%), raising the target probability above 140\% before declining as opposing nodes are reintroduced. This dip, which reduces LTP’s AUC, arises from (i)~attributing logits rather than \hbox{post-softmax} probabilities, thereby missing gains from suppressed competing logits, and (ii)~approximation errors in IOI's highly specialized circuit. The sharp early recovery nevertheless highlights LTP’s precision in identifying causal mechanisms. Extended results for generalizability are provided in Appendix~\ref{app:extended_results}.

\subsubsection{Computational Efficiency}
Beyond faithfulness, LTP is also substantially more computationally efficient in practice. As reported in Appendix~\ref{app:complexity_analysis} and Table~\ref{tab:empirical_measurements}, LTP completes attribution on Llama-3.1-8B-Instruct in 4.98 seconds on Known 1000, compared to 14.88 and 14.91 seconds for Activation Patching and DePass. The gap becomes much larger in the longer-context SQuAD v2.0 setting: LTP requires only 129.05 seconds, whereas Activation Patching and DePass require 5370.65 and 5452.28 seconds, respectively, yielding over a 40$\times$ speedup while maintaining comparable peak memory footprints. These empirical measurements show that LTP’s faithfulness gains come with strong practical scalability.

\subsection{Sensitivity Analysis}\label{sec:sensitivity_analysis}
In the sensitivity analysis, we evaluate how the attribution performance changes as we alter the weighting parameters $\mu$, which balance the contributions of different pathways. Table~\ref{tab:sensitivity_analysis} shows the configuration weights. Despite the Query-Key sensitivity analysis, a higher $p$ indicates a larger reliance on the content pathways.
\begin{itemize}
    \item \textbf{Control-Content Balance.} Varying $p$ shifts emphasis from pure content ($p \approx 1$) to pure control ($p \approx 0$), exposing how content–routing interactions affect faithfulness.
    \vspace{-1mm}
    \item \textbf{Query-Key Balance.} Within the attention control path, we fix $\mu_v = 0.5$ and redistribute the remaining weight between $\mu_q$ and $\mu_k$ to localize sensitivity to the `request' (query) versus `source' (key) of information.
    \vspace{-1mm}
    \item \textbf{Attention Balance.} With MLP weights fixed, we test whether attention-mediated effects are dominated by content being moved ($\mu_v$) or routing logic ($\mu_q, \mu_k$).
    \vspace{-1mm}
    \item \textbf{MLP Balance.} We focus on the Gated Linear Unit by fixing the attention mechanism to the balanced configuration and shifting the weights between $\mu_{gate}$ and $\mu_{up}$ to probe the gate's selective filtering behavior.
\end{itemize}

\begin{table}[!tp]
    \centering
    \small     
    \renewcommand{\arraystretch}{0.9}
    
    \begin{tabular}{l@{\hskip 16pt}ccc@{\hskip 16pt}cc}
        \toprule
        \raisebox{-0.5ex}{\textbf{Sensitivity}} &  $\mu_{q}$ & $\mu_{k}$ & $\mu_{v}$ & $\mu_{gate}$ & $\mu_{up}$ \\[1ex]
        \midrule
        Control-Content & $\frac{1 - p}{2}$ & $\frac{1 - p}{2}$ & $p$ & $1 - p$ & $p$ \\[1ex]
        Attention & $\frac{1 - p}{2}$ & $\frac{1 - p}{2}$ & $p$ & $0.5$ & $0.5$ \\[1ex]
        Query-Key & $\frac{p}{2}$ & $\frac{1 - p}{2}$ & $0.5$ & $0.5$ & $0.5$ \\[1ex]
        MLP & $0.25$ & $0.25$ & $0.5$ & $1 - p$ & $p$ \\
        \bottomrule
    \end{tabular}

    \caption{Hyperparameter configuration for the \hbox{sensitivity} analysis. $p \in [0, 1]$ is the sensitivity variable in our experiment.}
    \label{tab:sensitivity_analysis}
\end{table}

\begin{figure}[!tp]
    \includegraphics[width=1\columnwidth]{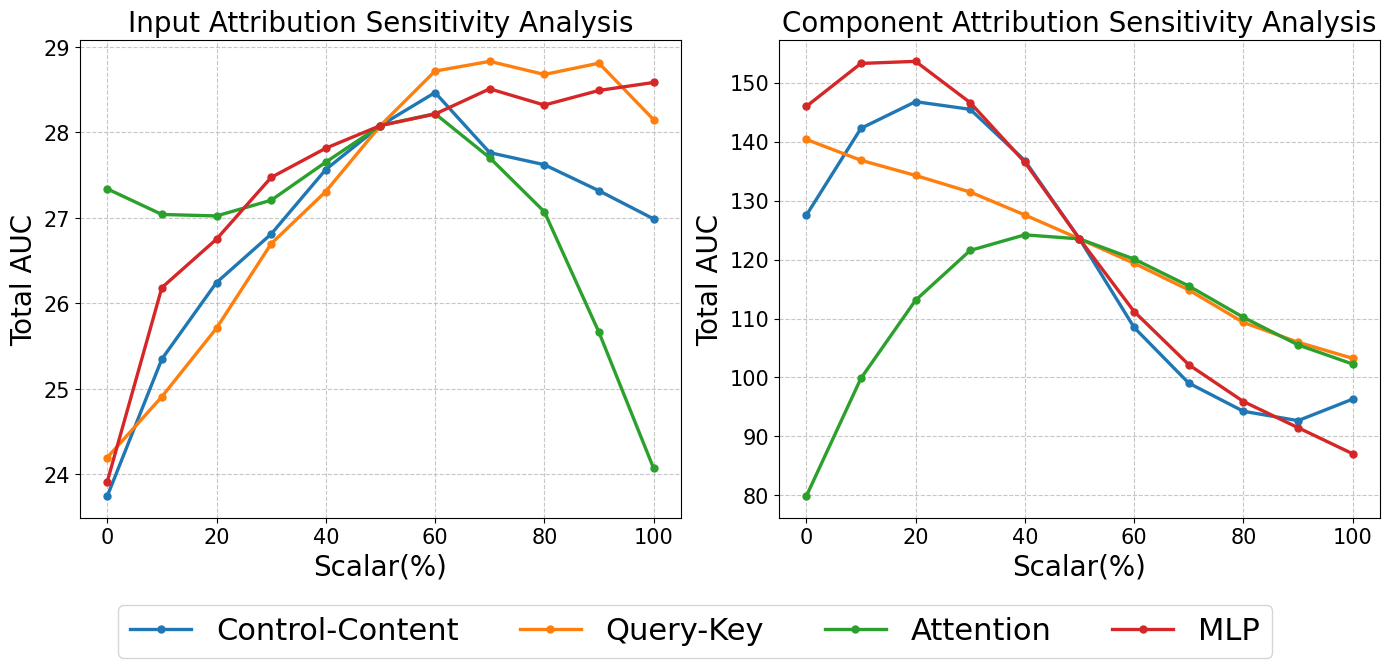}
    \vspace{-3.5ex}
    \caption{Sensitivity analysis of LTP scaling configurations on the Known 1000 dataset. We evaluate the impact of varying attribution parameters to prioritize different information pathways.~\label{fig:sensitivity_analysis}}
    \vspace{-2ex}
\end{figure}

\paragraph{Results.}
Figure~\ref{fig:sensitivity_analysis} summarizes the sensitivity \hbox{results}, with complementary qualitative \hbox{examples} deferred to Appendix~\ref{app:qualitative}. Input attribution peaks in the mid-range (50--70\%), indicating that \hbox{macro-level} token importance favors direct content pathways (e.g., value and up projections). In contrast, component attribution peaks at much smaller scales ($p \approx 0.2$) before a sharp decline, revealing that fine-grained fidelity hinges on control/routing pathways (gates and queries/keys), with $1-p \approx 0.8$ needed to capture individual functional roles. Considering both, $p=0.5$ provides the most stable trade-off by unifying the control and content pathways with equal weighting, which we use throughout the experiments in this paper. An extended sensitivity analysis on fundamentally different tasks and granularities, which further supports $p=0.5$ as a robust task-agnostic default, is provided in \hbox{Appendix}~\ref{app:extended_sensitivity}. We leave broader configuration sweeps to future work.

\section{Discussion}~\label{sec:discussion}
In this section, we discuss the scope and reliability of LTP beyond the main results. We first examine the implications of attributing logits rather than post-softmax probabilities. We then assess the causal validity of LTP by comparing its rankings against exact activation patching. Finally, we quantify the deviation from conservation introduced by the softmax linearizations.

\paragraph{Logit-level attribution.}
LTP attributes logits rather than post-softmax probabilities, as the \hbox{final} softmax poses a unique challenge: (1) the Deep Taylor Decomposition~\citep{Montavon2015ExplainingNC} distributes relevance via the gradient Jacobian, which vanishes for highly confident next-token predictions, and (2) the error introduced through the final softmax propagates through the entire network and is not bypassed by any linear connection, as is the case with the value projection in the attention block. By stopping at the logit level, however, the framework structurally misses suppressive mechanisms such as the negative name mover heads in IOI~\citep{Wang2022InterpretabilityIT}. \citet{achtibat2024attnlrp} suggest that these issues could be somewhat alleviated through temperature scaling; in line with their work, we defer the investigation of probability attribution to future work. Alternatively, inhibition circuits could be incorporated without propagating through the final softmax by using the logit residual as the target and attributing it to the component activations.

\paragraph{Agreement with activation patching.}
Our faithfulness evaluation relies on zero ablation to maintain comparability with previous relevant work, although such ablations may introduce \hbox{out-of-distribution} artifacts. To address this concern, we directly compare the LTP ranking with the causal effects obtained by exact activation patching on a component and sample subset. On IOI (\hbox{Llama-3.1-8B-Instruct;} 200 samples), we score all 1,024 attention heads with both LTP and exact activation patching. The two methods agree strongly: \hbox{Pearson} $r = 0.80$, a 70\% overlap among the \hbox{top-20} heads, and Spearman $\rho = 0.905$ with 100\% sign agreement among the causally effective heads. The more moderate full-population rank correlation ($\rho = 0.39$) arises from the weaker alignment of the ${\sim}900$ of the 1{,}024 heads with negligible causal effect, which are therefore more prone to order variations due to minimal noise.

\paragraph{Conservation in practice.}
LTP does not take conservation as a foundational axiom; rather, it is a property that comes with the approach. The choice of the softmax linearization breaks conservation through its implicit bias term, which consumes a small proportion of the propagated relevance. Empirical measurements on 200 examples of the Known 1000 dataset show that the biases in each layer account for ${\sim}2.3\%$ of the consumption, where the decrease is not monotone but slightly concave, with a higher loss in late and a lower loss in early layers. At the input, LTP exhibits a normed absolute mean residual (NAMR) of 0.73 with a standard deviation of 0.042, compared to 1.01 (0.087) for Input$\times$Gradient and 6.592 (9.839) for the gradient approach, indicating that LTP remains substantially closer to the target logit than gradient-based alternatives.


\section{Conclusion}~\label{sec:conclusion}
In this paper, we introduce \emph{Layer-wise Target Propagation} (LTP), a novel framework that recasts attribution as a target-centric backward process: LTP \hbox{propagates} the unembedding vector of the target token backward through the frozen transformer in a single forward and backward pass, recovering an effective target at every residual position without counterfactual examples. To make this propagation accurate, LTP decomposes and linearizes the transformer's computation into distinct pathways through which the target is traced. LTP achieves $O(1)$ time complexity with respect to the number of model components, demonstrating scalability for large models. Extensive experiments on standard interpretability benchmarks demonstrate that LTP achieves state-of-the-art faithfulness and \hbox{unprecedented} efficiency compared to existing baselines.


\section{Limitations}

While Layer-wise Target Propagation (LTP) offers a highly efficient framework for dense component attribution, its methodology introduces several structural and operational limitations:

    
\textbf{Approximation in High-Curvature Regimes:} To achieve top-down propagation, LTP relies on local linearization and first-order Taylor expansions. Because it inherently discards second-order cross-derivatives, the method's attribution faithfulness may degrade when analyzing circuits driven by highly coupled, synergistic, or heavily non-linear component interactions.
    
    
\textbf{Memory and Compute Overhead:} Although LTP drastically reduces the time complexity to $O(1)$ with respect to the number of components, caching the dense intermediate activations during the forward pass is memory-intensive. Therefore, despite its comparative efficiency, conducting LTP on exceptionally large models or ultra-long context windows still demands a significant memory footprint and computational overhead.

\section{Acknowledgments}
This work was supported by the National Research Foundation of Korea (NRF) grants funded by the Korean government (MSIT) (RS-2025-00513098 and RS-2025-16067408) and the BK21 FOUR project (AI-driven Convergence Software Education Research Program) funded by the Ministry of Education at Kyungpook National University (KNU) (4120240214871). The A100 GPU used in the experiments was provided by the Department of Data Convergence \hbox{Computing} at KNU, where \hbox{Young-Kyoon} Suh is an adjunct professor.

\bibliography{anthology-1,anthology-2, custom} 

\begin{thebibliography}{44}
\providecommand{\natexlab}[1]{#1}

\bibitem[{Abnar and Zuidema(2020)}]{abnar-zuidema-2020-quantifying}
Samira Abnar and Willem Zuidema. 2020.
\newblock \href {https://doi.org/10.18653/v1/2020.acl-main.385} {Quantifying attention flow in transformers}.
\newblock In \emph{Proceedings of the 58th Annual Meeting of the Association for Computational Linguistics}, pages 4190--4197, Online. Association for Computational Linguistics.

\bibitem[{Achtibat et~al.(2024)Achtibat, Hatefi, Dreyer, Jain, Wiegand, Lapuschkin, and Samek}]{achtibat2024attnlrp}
Reduan Achtibat, Sayed Mohammad~Vakilzadeh Hatefi, Maximilian Dreyer, Aakriti Jain, Thomas Wiegand, Sebastian Lapuschkin, and Wojciech Samek. 2024.
\newblock \href {https://arxiv.org/abs/2402.05602} {Attnlrp: Attention-aware layer-wise relevance propagation for transformers}.
\newblock \emph{Preprint}, arXiv:2402.05602.

\bibitem[{Balduzzi et~al.(2018)Balduzzi, Frean, Leary, Lewis, Ma, and McWilliams}]{balduzzi2018shattered}
David Balduzzi, Marcus Frean, Lennox Leary, JP~Lewis, Kurt Wan-Duo Ma, and Brian McWilliams. 2018.
\newblock \href {https://arxiv.org/abs/1702.08591} {The shattered gradients problem: If resnets are the answer, then what is the question?}
\newblock \emph{Preprint}, arXiv:1702.08591.

\bibitem[{Belrose et~al.(2023)Belrose, Furman, Smith, Halawi, Ostrovsky, McKinney, Biderman, and Steinhardt}]{Belrose2023ElicitingLP}
Nora Belrose, Zach Furman, Logan Smith, Danny Halawi, Igor~V. Ostrovsky, Lev McKinney, Stella Biderman, and Jacob Steinhardt. 2023.
\newblock \href {https://api.semanticscholar.org/CorpusID:257504984} {Eliciting latent predictions from transformers with the tuned lens}.
\newblock \emph{ArXiv}, abs/2303.08112.

\bibitem[{Conmy et~al.(2023)Conmy, Mavor-Parker, Lynch, Heimersheim, and Garriga-Alonso}]{Conmy2023TowardsAC}
Arthur Conmy, Augustine~N. Mavor-Parker, Aengus Lynch, Stefan Heimersheim, and Adri{\`a} Garriga-Alonso. 2023.
\newblock \href {https://api.semanticscholar.org/CorpusID:258418244} {Towards automated circuit discovery for mechanistic interpretability}.
\newblock \emph{ArXiv}, abs/2304.14997.

\bibitem[{Denil et~al.(2014)Denil, Demiraj, and de~Freitas}]{Denil2014ExtractionOS}
Misha Denil, Alban Demiraj, and Nando de~Freitas. 2014.
\newblock \href {https://api.semanticscholar.org/CorpusID:9121062} {Extraction of salient sentences from labelled documents}.
\newblock \emph{ArXiv}, abs/1412.6815.

\bibitem[{DeYoung et~al.(2020)DeYoung, Jain, Rajani, Lehman, Xiong, Socher, and Wallace}]{deyoung-etal-2020-eraser}
Jay DeYoung, Sarthak Jain, Nazneen~Fatema Rajani, Eric Lehman, Caiming Xiong, Richard Socher, and Byron~C. Wallace. 2020.
\newblock \href {https://doi.org/10.18653/v1/2020.acl-main.408} {{ERASER}: {A} benchmark to evaluate rationalized {NLP} models}.
\newblock In \emph{Proceedings of the 58th Annual Meeting of the Association for Computational Linguistics}, pages 4443--4458, Online. Association for Computational Linguistics.

\bibitem[{Elhage et~al.(2021)Elhage, Nanda, Olsson, Henighan, Joseph, Mann, Askell, Bai, Chen, Conerly, DasSarma, Drain, Ganguli, Hatfield-Dodds, Hernandez, Jones, Kernion, Lovitt, Ndousse, Amodei, Brown, Clark, Kaplan, McCandlish, and Olah}]{elhage2021mathematical}
Nelson Elhage, Neel Nanda, Catherine Olsson, Tom Henighan, Nicholas Joseph, Ben Mann, Amanda Askell, Yuntao Bai, Anna Chen, Tom Conerly, Nova DasSarma, Dawn Drain, Deep Ganguli, Zac Hatfield-Dodds, Danny Hernandez, Andy Jones, Jackson Kernion, Liane Lovitt, Kamal Ndousse, and 6 others. 2021.
\newblock A mathematical framework for transformer circuits.
\newblock \emph{Transformer Circuits Thread}.
\newblock Https://transformer-circuits.pub/2021/framework/index.html.

\bibitem[{Ferrando et~al.(2022)Ferrando, G{\'a}llego, and Costa-juss{\`a}}]{ferrando-etal-2022-measuring}
Javier Ferrando, Gerard~I. G{\'a}llego, and Marta~R. Costa-juss{\`a}. 2022.
\newblock \href {https://doi.org/10.18653/v1/2022.emnlp-main.595} {Measuring the mixing of contextual information in the transformer}.
\newblock In \emph{Proceedings of the 2022 Conference on Empirical Methods in Natural Language Processing}, pages 8698--8714, Abu Dhabi, United Arab Emirates. Association for Computational Linguistics.

\bibitem[{Ferrando et~al.(2023)Ferrando, G{\'a}llego, Tsiamas, and Costa-juss{\`a}}]{Ferrando2023ExplainingHT}
Javier Ferrando, Gerard~I. G{\'a}llego, Yiannis~(Ioannis) Tsiamas, and Marta~Ruiz Costa-juss{\`a}. 2023.
\newblock \href {https://api.semanticscholar.org/CorpusID:258832652} {Explaining how transformers use context to build predictions}.
\newblock In \emph{Annual Meeting of the Association for Computational Linguistics}.

\bibitem[{Ferrando et~al.(2024)Ferrando, Sarti, Bisazza, and Costa-juss{\`a}}]{Ferrando2024APO}
Javier Ferrando, Gabriele Sarti, Arianna Bisazza, and Marta~Ruiz Costa-juss{\`a}. 2024.
\newblock \href {https://api.semanticscholar.org/CorpusID:269484740} {A primer on the inner workings of transformer-based language models}.
\newblock \emph{ArXiv}, abs/2405.00208.

\bibitem[{Ferrando and Voita(2024)}]{ferrando-voita-2024-information}
Javier Ferrando and Elena Voita. 2024.
\newblock \href {https://doi.org/10.18653/v1/2024.emnlp-main.965} {Information flow routes: Automatically interpreting language models at scale}.
\newblock In \emph{Proceedings of the 2024 Conference on Empirical Methods in Natural Language Processing}, pages 17432--17445, Miami, Florida, USA. Association for Computational Linguistics.

\bibitem[{Geva et~al.(2022)Geva, Caciularu, Wang, and Goldberg}]{Geva2022TransformerFL}
Mor Geva, Avi Caciularu, Ke~Wang, and Yoav Goldberg. 2022.
\newblock \href {https://api.semanticscholar.org/CorpusID:247762385} {Transformer feed-forward layers build predictions by promoting concepts in the vocabulary space}.
\newblock \emph{ArXiv}, abs/2203.14680.

\bibitem[{Grattafiori et~al.(2024)Grattafiori, Dubey, Jauhri, Pandey, Kadian, Al-Dahle, Letman, Mathur, Schelten, Vaughan, Yang, Fan, Goyal, Hartshorn, Yang, Mitra, Sravankumar, Korenev, Hinsvark, Rao, Zhang, Rodriguez, Gregerson, Spataru, Roziere, Biron, Tang, Chern, Caucheteux, Nayak, Bi, Marra, McConnell, Keller, Touret, Wu, Wong, Ferrer, Nikolaidis, Allonsius, Song, Pintz, Livshits, Wyatt, Esiobu, Choudhary, Mahajan, Garcia-Olano, Perino, Hupkes, Lakomkin, AlBadawy, Lobanova, Dinan, Smith, Radenovic, Guzmán, Zhang, Synnaeve, Lee, Anderson, Thattai, Nail, Mialon, Pang, Cucurell, Nguyen, Korevaar, Xu, Touvron, Zarov, Ibarra, Kloumann, Misra, Evtimov, Zhang, Copet, Lee, Geffert, Vranes, Park, Mahadeokar, Shah, van~der Linde, Billock, Hong, Lee, Fu, Chi, Huang, Liu, Wang, Yu, Bitton, Spisak, Park, Rocca, Johnstun, Saxe, Jia, Alwala, Prasad, Upasani, Plawiak, Li, Heafield, Stone, El-Arini, Iyer, Malik, Chiu, Bhalla, Lakhotia, Rantala-Yeary, van~der Maaten, Chen, Tan, Jenkins, Martin, Madaan, Malo, Blecher,
  Landzaat, de~Oliveira, Muzzi, Pasupuleti, Singh, Paluri, Kardas, Tsimpoukelli, Oldham, Rita, Pavlova, Kambadur, Lewis, Si, Singh, Hassan, Goyal, Torabi, Bashlykov, Bogoychev, Chatterji, Zhang, Duchenne, Çelebi, Alrassy, Zhang, Li, Vasic, Weng, Bhargava, Dubal, Krishnan, Koura, Xu, He, Dong, Srinivasan, Ganapathy, Calderer, Cabral, Stojnic, Raileanu, Maheswari, Girdhar, Patel, Sauvestre, Polidoro, Sumbaly, Taylor, Silva, Hou, Wang, Hosseini, Chennabasappa, Singh, Bell, Kim, Edunov, Nie, Narang, Raparthy, Shen, Wan, Bhosale, Zhang, Vandenhende, Batra, Whitman, Sootla, Collot, Gururangan, Borodinsky, Herman, Fowler, Sheasha, Georgiou, Scialom, Speckbacher, Mihaylov, Xiao, Karn, Goswami, Gupta, Ramanathan, Kerkez, Gonguet, Do, Vogeti, Albiero, Petrovic, Chu, Xiong, Fu, Meers, Martinet, Wang, Wang, Tan, Xia, Xie, Jia, Wang, Goldschlag, Gaur, Babaei, Wen, Song, Zhang, Li, Mao, Coudert, Yan, Chen, Papakipos, Singh, Srivastava, Jain, Kelsey, Shajnfeld, Gangidi, Victoria, Goldstand, Menon, Sharma, Boesenberg,
  Baevski, Feinstein, Kallet, Sangani, Teo, Yunus, Lupu, Alvarado, Caples, Gu, Ho, Poulton, Ryan, Ramchandani, Dong, Franco, Goyal, Saraf, Chowdhury, Gabriel, Bharambe, Eisenman, Yazdan, James, Maurer, Leonhardi, Huang, Loyd, Paola, Paranjape, Liu, Wu, Ni, Hancock, Wasti, Spence, Stojkovic, Gamido, Montalvo, Parker, Burton, Mejia, Liu, Wang, Kim, Zhou, Hu, Chu, Cai, Tindal, Feichtenhofer, Gao, Civin, Beaty, Kreymer, Li, Adkins, Xu, Testuggine, David, Parikh, Liskovich, Foss, Wang, Le, Holland, Dowling, Jamil, Montgomery, Presani, Hahn, Wood, Le, Brinkman, Arcaute, Dunbar, Smothers, Sun, Kreuk, Tian, Kokkinos, Ozgenel, Caggioni, Kanayet, Seide, Florez, Schwarz, Badeer, Swee, Halpern, Herman, Sizov, Guangyi, Zhang, Lakshminarayanan, Inan, Shojanazeri, Zou, Wang, Zha, Habeeb, Rudolph, Suk, Aspegren, Goldman, Zhan, Damlaj, Molybog, Tufanov, Leontiadis, Veliche, Gat, Weissman, Geboski, Kohli, Lam, Asher, Gaya, Marcus, Tang, Chan, Zhen, Reizenstein, Teboul, Zhong, Jin, Yang, Cummings, Carvill, Shepard, McPhie,
  Torres, Ginsburg, Wang, Wu, U, Saxena, Khandelwal, Zand, Matosich, Veeraraghavan, Michelena, Li, Jagadeesh, Huang, Chawla, Huang, Chen, Garg, A, Silva, Bell, Zhang, Guo, Yu, Moshkovich, Wehrstedt, Khabsa, Avalani, Bhatt, Mankus, Hasson, Lennie, Reso, Groshev, Naumov, Lathi, Keneally, Liu, Seltzer, Valko, Restrepo, Patel, Vyatskov, Samvelyan, Clark, Macey, Wang, Hermoso, Metanat, Rastegari, Bansal, Santhanam, Parks, White, Bawa, Singhal, Egebo, Usunier, Mehta, Laptev, Dong, Cheng, Chernoguz, Hart, Salpekar, Kalinli, Kent, Parekh, Saab, Balaji, Rittner, Bontrager, Roux, Dollar, Zvyagina, Ratanchandani, Yuvraj, Liang, Alao, Rodriguez, Ayub, Murthy, Nayani, Mitra, Parthasarathy, Li, Hogan, Battey, Wang, Howes, Rinott, Mehta, Siby, Bondu, Datta, Chugh, Hunt, Dhillon, Sidorov, Pan, Mahajan, Verma, Yamamoto, Ramaswamy, Lindsay, Lindsay, Feng, Lin, Zha, Patil, Shankar, Zhang, Zhang, Wang, Agarwal, Sajuyigbe, Chintala, Max, Chen, Kehoe, Satterfield, Govindaprasad, Gupta, Deng, Cho, Virk, Subramanian, Choudhury,
  Goldman, Remez, Glaser, Best, Koehler, Robinson, Li, Zhang, Matthews, Chou, Shaked, Vontimitta, Ajayi, Montanez, Mohan, Kumar, Mangla, Ionescu, Poenaru, Mihailescu, Ivanov, Li, Wang, Jiang, Bouaziz, Constable, Tang, Wu, Wang, Wu, Gao, Kleinman, Chen, Hu, Jia, Qi, Li, Zhang, Zhang, Adi, Nam, Yu, Wang, Zhao, Hao, Qian, Li, He, Rait, DeVito, Rosnbrick, Wen, Yang, Zhao, and Ma}]{grattafiori2024llama3herdmodels}
Aaron Grattafiori, Abhimanyu Dubey, Abhinav Jauhri, Abhinav Pandey, Abhishek Kadian, Ahmad Al-Dahle, Aiesha Letman, Akhil Mathur, Alan Schelten, Alex Vaughan, Amy Yang, Angela Fan, Anirudh Goyal, Anthony Hartshorn, Aobo Yang, Archi Mitra, Archie Sravankumar, Artem Korenev, Arthur Hinsvark, and 542 others. 2024.
\newblock \href {https://arxiv.org/abs/2407.21783} {The llama 3 herd of models}.
\newblock \emph{Preprint}, arXiv:2407.21783.

\bibitem[{Hong et~al.(2025)Hong, Jiang, Tian, Qi, Sun, Ding, and Zhou}]{Hong2025DePassUF}
Xiangyu Hong, Che Jiang, Kai Tian, Biqing Qi, Youbang Sun, Ning Ding, and Bowen Zhou. 2025.
\newblock \href {https://api.semanticscholar.org/CorpusID:282246406} {Depass: Unified feature attributing by simple decomposed forward pass}.
\newblock \emph{ArXiv}, abs/2510.18462.

\bibitem[{Jafari et~al.(2025)Jafari, Eberle, Khakzar, and Nanda}]{jafari_relp_2025}
Farnoush~Rezaei Jafari, Oliver Eberle, Ashkan Khakzar, and Neel Nanda. 2025.
\newblock \href {https://doi.org/10.48550/arXiv.2508.21258} {{RelP}: Faithful and efficient circuit discovery in language models via relevance patching}.
\newblock \emph{Preprint}, arxiv:2508.21258 [cs].

\bibitem[{Jain and Wallace(2019)}]{Jain2019AttentionIN}
Sarthak Jain and Byron~C. Wallace. 2019.
\newblock \href {https://api.semanticscholar.org/CorpusID:67855860} {Attention is not explanation}.
\newblock In \emph{North American Chapter of the Association for Computational Linguistics}.

\bibitem[{Jiang et~al.(2023)Jiang, Sablayrolles, Mensch, Bamford, Chaplot, de~Las~Casas, Bressand, Lengyel, Lample, Saulnier, Lavaud, Lachaux, Stock, Scao, Lavril, Wang, Lacroix, and Sayed}]{Jiang2023Mistral7}
Albert~Qiaochu Jiang, Alexandre Sablayrolles, Arthur Mensch, Chris Bamford, Devendra~Singh Chaplot, Diego de~Las~Casas, Florian Bressand, Gianna Lengyel, Guillaume Lample, Lucile Saulnier, L{\'e}lio~Renard Lavaud, Marie-Anne Lachaux, Pierre Stock, Teven~Le Scao, Thibaut Lavril, Thomas Wang, Timoth{\'e}e Lacroix, and William~El Sayed. 2023.
\newblock \href {https://api.semanticscholar.org/CorpusID:263830494} {Mistral 7b}.
\newblock \emph{ArXiv}, abs/2310.06825.

\bibitem[{Kobayashi et~al.(2020)Kobayashi, Kuribayashi, Yokoi, and Inui}]{kobayashi-etal-2020-attention}
Goro Kobayashi, Tatsuki Kuribayashi, Sho Yokoi, and Kentaro Inui. 2020.
\newblock \href {https://doi.org/10.18653/v1/2020.emnlp-main.574} {Attention is not only a weight: Analyzing transformers with vector norms}.
\newblock In \emph{Proceedings of the 2020 Conference on Empirical Methods in Natural Language Processing (EMNLP)}, pages 7057--7075, Online. Association for Computational Linguistics.

\bibitem[{Kobayashi et~al.(2021)Kobayashi, Kuribayashi, Yokoi, and Inui}]{kobayashi-etal-2021-incorporating}
Goro Kobayashi, Tatsuki Kuribayashi, Sho Yokoi, and Kentaro Inui. 2021.
\newblock \href {https://doi.org/10.18653/v1/2021.emnlp-main.373} {{I}ncorporating {R}esidual and {N}ormalization {L}ayers into {A}nalysis of {M}asked {L}anguage {M}odels}.
\newblock In \emph{Proceedings of the 2021 Conference on Empirical Methods in Natural Language Processing}, pages 4547--4568, Online and Punta Cana, Dominican Republic. Association for Computational Linguistics.

\bibitem[{Kobayashi et~al.(2023)Kobayashi, Kuribayashi, Yokoi, and Inui}]{Kobayashi2023AnalyzingFB}
Goro Kobayashi, Tatsuki Kuribayashi, Sho Yokoi, and Kentaro Inui. 2023.
\newblock \href {https://api.semanticscholar.org/CorpusID:256459817} {Analyzing feed-forward blocks in transformers through the lens of attention maps}.
\newblock In \emph{International Conference on Learning Representations}.

\bibitem[{Li et~al.(2015)Li, Chen, Hovy, and Jurafsky}]{Li2015VisualizingAU}
Jiwei Li, Xinlei Chen, Eduard~H. Hovy, and Dan Jurafsky. 2015.
\newblock \href {https://api.semanticscholar.org/CorpusID:14099741} {Visualizing and understanding neural models in nlp}.
\newblock In \emph{North American Chapter of the Association for Computational Linguistics}.

\bibitem[{Maas et~al.(2011)Maas, Daly, Pham, Huang, Ng, and Potts}]{maas-EtAl:2011:ACL-HLT2011}
Andrew~L. Maas, Raymond~E. Daly, Peter~T. Pham, Dan Huang, Andrew~Y. Ng, and Christopher Potts. 2011.
\newblock \href {http://www.aclweb.org/anthology/P11-1015} {Learning word vectors for sentiment analysis}.
\newblock In \emph{Proceedings of the 49th Annual Meeting of the Association for Computational Linguistics: Human Language Technologies}, pages 142--150, Portland, Oregon, USA. Association for Computational Linguistics.

\bibitem[{Meng et~al.(2022)Meng, Bau, Andonian, and Belinkov}]{Meng2022LocatingAE}
Kevin Meng, David Bau, Alex Andonian, and Yonatan Belinkov. 2022.
\newblock \href {https://api.semanticscholar.org/CorpusID:255825985} {Locating and editing factual associations in gpt}.
\newblock In \emph{Neural Information Processing Systems}.

\bibitem[{Modarressi et~al.(2023)Modarressi, Fayyaz, Aghazadeh, Yaghoobzadeh, and Pilehvar}]{modarressi-etal-2023-decompx}
Ali Modarressi, Mohsen Fayyaz, Ehsan Aghazadeh, Yadollah Yaghoobzadeh, and Mohammad~Taher Pilehvar. 2023.
\newblock \href {https://doi.org/10.18653/v1/2023.acl-long.149} {{D}ecomp{X}: Explaining transformers decisions by propagating token decomposition}.
\newblock In \emph{Proceedings of the 61st Annual Meeting of the Association for Computational Linguistics (Volume 1: Long Papers)}, pages 2649--2664, Toronto, Canada. Association for Computational Linguistics.

\bibitem[{Modarressi et~al.(2022)Modarressi, Fayyaz, Yaghoobzadeh, and Pilehvar}]{modarressi-etal-2022-globenc}
Ali Modarressi, Mohsen Fayyaz, Yadollah Yaghoobzadeh, and Mohammad~Taher Pilehvar. 2022.
\newblock \href {https://doi.org/10.18653/v1/2022.naacl-main.19} {{G}lob{E}nc: Quantifying global token attribution by incorporating the whole encoder layer in transformers}.
\newblock In \emph{Proceedings of the 2022 Conference of the North American Chapter of the Association for Computational Linguistics: Human Language Technologies}, pages 258--271, Seattle, United States. Association for Computational Linguistics.

\bibitem[{Montavon et~al.(2015)Montavon, Lapuschkin, Binder, Samek, and M{\"u}ller}]{Montavon2015ExplainingNC}
Gr{\'e}goire Montavon, Sebastian Lapuschkin, Alexander Binder, Wojciech Samek, and Klaus-Robert M{\"u}ller. 2015.
\newblock \href {https://api.semanticscholar.org/CorpusID:266022338} {Explaining nonlinear classification decisions with deep taylor decomposition}.
\newblock \emph{ArXiv}, abs/1512.02479.

\bibitem[{Nanda(2023)}]{nanda2023attribution}
Neel Nanda. 2023.
\newblock \href {https://www.neelnanda.io/mechanistic-interpretability/attribution-patching} {Attribution patching: Activation patching at industrial scale}.

\bibitem[{nostalgebraist(2020)}]{nostalgebraist2020logit}
nostalgebraist. 2020.
\newblock \href {https://www.alignmentforum.org/posts/AcKRB8wDpdaN6v6ru/interpreting-gpt-the-logit-lens} {Interpreting gpt: the logit lens}.
\newblock \emph{AI Alignment Forum}.

\bibitem[{Radford et~al.(2019)Radford, Wu, Child, Luan, Amodei, and Sutskever}]{Radford2019LanguageMA}
Alec Radford, Jeff Wu, Rewon Child, David Luan, Dario Amodei, and Ilya Sutskever. 2019.
\newblock \href {https://api.semanticscholar.org/CorpusID:160025533} {Language models are unsupervised multitask learners}.

\bibitem[{Rajpurkar et~al.(2018)Rajpurkar, Jia, and Liang}]{Rajpurkar2018KnowWY}
Pranav Rajpurkar, Robin Jia, and Percy Liang. 2018.
\newblock \href {https://api.semanticscholar.org/CorpusID:47018994} {Know what you don’t know: Unanswerable questions for squad}.
\newblock \emph{ArXiv}, abs/1806.03822.

\bibitem[{Shrikumar et~al.(2019)Shrikumar, Greenside, and Kundaje}]{shrikumar2019learning}
Avanti Shrikumar, Peyton Greenside, and Anshul Kundaje. 2019.
\newblock \href {https://arxiv.org/abs/1704.02685} {Learning important features through propagating activation differences}.
\newblock \emph{Preprint}, arXiv:1704.02685.

\bibitem[{Simonyan et~al.(2013)Simonyan, Vedaldi, and Zisserman}]{Simonyan2013DeepIC}
Karen Simonyan, Andrea Vedaldi, and Andrew Zisserman. 2013.
\newblock \href {https://api.semanticscholar.org/CorpusID:1450294} {Deep inside convolutional networks: Visualising image classification models and saliency maps}.
\newblock \emph{CoRR}, abs/1312.6034.

\bibitem[{Smilkov et~al.(2017)Smilkov, Thorat, Kim, Viégas, and Wattenberg}]{smilkov2017smoothgrad}
Daniel Smilkov, Nikhil Thorat, Been Kim, Fernanda Viégas, and Martin Wattenberg. 2017.
\newblock \href {https://arxiv.org/abs/1706.03825} {Smoothgrad: removing noise by adding noise}.
\newblock \emph{Preprint}, arXiv:1706.03825.

\bibitem[{Su et~al.(2021)Su, Lu, Pan, Wen, and Liu}]{Su2021RoFormerET}
Jianlin Su, Yu~Lu, Shengfeng Pan, Bo~Wen, and Yunfeng Liu. 2021.
\newblock \href {https://api.semanticscholar.org/CorpusID:233307138} {Roformer: Enhanced transformer with rotary position embedding}.
\newblock \emph{ArXiv}, abs/2104.09864.

\bibitem[{Sundararajan et~al.(2017)Sundararajan, Taly, and Yan}]{pmlr-v70-sundararajan17a}
Mukund Sundararajan, Ankur Taly, and Qiqi Yan. 2017.
\newblock \href {https://proceedings.mlr.press/v70/sundararajan17a.html} {Axiomatic attribution for deep networks}.
\newblock In \emph{Proceedings of the 34th International Conference on Machine Learning}, volume~70 of \emph{Proceedings of Machine Learning Research}, pages 3319--3328. PMLR.

\bibitem[{Syed et~al.(2024)Syed, Rager, and Conmy}]{syed-etal-2024-attribution}
Aaquib Syed, Can Rager, and Arthur Conmy. 2024.
\newblock \href {https://doi.org/10.18653/v1/2024.blackboxnlp-1.25} {Attribution patching outperforms automated circuit discovery}.
\newblock In \emph{Proceedings of the 7th BlackboxNLP Workshop: Analyzing and Interpreting Neural Networks for NLP}, pages 407--416, Miami, Florida, US. Association for Computational Linguistics.

\bibitem[{Touvron et~al.(2023)Touvron, Martin, Stone, Albert, Almahairi, Babaei, lay Bashlykov, Batra, Bhargava, Bhosale, Bikel, Blecher, Ferrer, Chen, Cucurull, Esiobu, Fernandes, Fu, Fu, Fuller, Gao, Goswami, Goyal, Hartshorn, Hosseini, Hou, Inan, Kardas, Kerkez, Khabsa, Kloumann, Korenev, Koura, Lachaux, Lavril, Lee, Liskovich, Lu, Mao, Martinet, Mihaylov, Mishra, Molybog, Nie, Poulton, Reizenstein, Rungta, Saladi, Schelten, Silva, Smith, Subramanian, Tan, Tang, Taylor, Williams, Kuan, Xu, Yan, Zarov, Zhang, Fan, Kambadur, Narang, Rodriguez, Stojnic, Edunov, and Scialom}]{Touvron2023Llama2O}
Hugo Touvron, Louis Martin, Kevin~R. Stone, Peter Albert, Amjad Almahairi, Yasmine Babaei, Niko lay Bashlykov, Soumya Batra, Prajjwal Bhargava, Shruti Bhosale, Daniel~M. Bikel, Lukas Blecher, Cristian~Canton Ferrer, Moya Chen, Guillem Cucurull, David Esiobu, Jude Fernandes, Jeremy Fu, Wenyin Fu, and 49 others. 2023.
\newblock \href {https://api.semanticscholar.org/CorpusID:259950998} {Llama 2: Open foundation and fine-tuned chat models}.
\newblock \emph{ArXiv}, abs/2307.09288.

\bibitem[{Vig et~al.(2020)Vig, Gehrmann, Belinkov, Qian, Nevo, Singer, and Shieber}]{Vig2020InvestigatingGB}
Jesse Vig, Sebastian Gehrmann, Yonatan Belinkov, Sharon Qian, Daniel Nevo, Yaron Singer, and Stuart~M. Shieber. 2020.
\newblock \href {https://api.semanticscholar.org/CorpusID:227275068} {Investigating gender bias in language models using causal mediation analysis}.
\newblock In \emph{Neural Information Processing Systems}.

\bibitem[{Wang et~al.(2022)Wang, Variengien, Conmy, Shlegeris, and Steinhardt}]{Wang2022InterpretabilityIT}
Kevin Wang, Alexandre Variengien, Arthur Conmy, Buck Shlegeris, and Jacob Steinhardt. 2022.
\newblock \href {https://api.semanticscholar.org/CorpusID:253244237} {Interpretability in the wild: a circuit for indirect object identification in gpt-2 small}.
\newblock \emph{ArXiv}, abs/2211.00593.

\bibitem[{Yang et~al.(2025)Yang, Li, Yang, Zhang, Hui, Zheng, Yu, Gao, Huang, Lv, Zheng, Liu, Zhou, Huang, Hu, Ge, Wei, Lin, Tang, Yang, Tu, Zhang, Yang, Yang, Zhou, Zhou, Lin, Dang, Bao, Yang, Yu, Deng, Li, Xue, Li, Zhang, Wang, Zhu, Men, Gao, Liu, Luo, Li, Tang, Yin, Ren, Wang, Zhang, Ren, Fan, Su, Zhang, Zhang, Wan, Liu, Wang, Cui, Zhang, Zhou, and Qiu}]{yang2025qwen3technicalreport}
An~Yang, Anfeng Li, Baosong Yang, Beichen Zhang, Binyuan Hui, Bo~Zheng, Bowen Yu, Chang Gao, Chengen Huang, Chenxu Lv, Chujie Zheng, Dayiheng Liu, Fan Zhou, Fei Huang, Feng Hu, Hao Ge, Haoran Wei, Huan Lin, Jialong Tang, and 41 others. 2025.
\newblock \href {https://arxiv.org/abs/2505.09388} {Qwen3 technical report}.
\newblock \emph{Preprint}, arXiv:2505.09388.

\bibitem[{Yang et~al.(2024)Yang, Yang, Zhang, Hui, Zheng, Yu, Li, Liu, Huang, Dong, Wei, Lin, Yang, Tu, Zhang, Yang, Yang, Zhou, Lin, Dang, Lu, Bao, Yang, Yu, Li, Xue, Zhang, Zhu, Men, Lin, Li, Xia, Ren, Ren, Fan, Su, Zhang, Wan, Liu, Cui, Zhang, Qiu, Quan, and Wang}]{Yang2024Qwen25TR}
Qwen~An Yang, Baosong Yang, Beichen Zhang, Binyuan Hui, Bo~Zheng, Bowen Yu, Chengyuan Li, Dayiheng Liu, Fei Huang, Guanting Dong, Haoran Wei, Huan Lin, Jian Yang, Jianhong Tu, Jianwei Zhang, Jianxin Yang, Jiaxin Yang, Jingren Zhou, Junyang Lin, and 25 others. 2024.
\newblock \href {https://api.semanticscholar.org/CorpusID:274859421} {Qwen2.5 technical report}.
\newblock \emph{ArXiv}, abs/2412.15115.

\bibitem[{Zhang and Sennrich(2019)}]{Zhang2019RootMS}
Biao Zhang and Rico Sennrich. 2019.
\newblock \href {https://api.semanticscholar.org/CorpusID:113405151} {Root mean square layer normalization}.
\newblock \emph{ArXiv}, abs/1910.07467.

\bibitem[{Zhang and Nanda(2023)}]{Zhang2023TowardsBP}
Fred Zhang and Neel Nanda. 2023.
\newblock \href {https://api.semanticscholar.org/CorpusID:263131114} {Towards best practices of activation patching in language models: Metrics and methods}.
\newblock \emph{ArXiv}, abs/2309.16042.

\end{thebibliography}

\clearpage
\newpage
\appendix
\section{Transformer Architecture}~\label{app:transformer_architecture}
We consider a decoder-only Transformer composed of a stack of $L$ identical layers indexed by $l \in \{1, \dots, L\}$. Each layer consists of two primary sub-modules: a \emph{multi-head self-attention} (MHSA) block and a \emph{gated feed-forward network} (GLU).

Given an input token sequence $t = (t_1, \dots, t_T)$, tokens are mapped to vector embeddings $\mathbf{X}^{(0)} \in \mathbb{R}^{T \times d}$ using an embedding matrix $\mathbf{W}_E \in \mathbb{R}^{|\mathcal{V}| \times d}$, where $\mathcal{V}$ denotes the vocabulary. Representations propagate sequentially through the network \hbox{layers}. Under a pre-normalization configuration, the \hbox{update} rule for layer $l$ is given by:
\vspace{-0.5ex}
\begin{align*}
    \mathbf{X}^{(l)}_{\text{mid}} &= \mathbf{X}^{(l-1)} + \text{MHSA}\!\left(\text{LN}(\mathbf{X}^{(l-1)})\right), \\
    \mathbf{X}^{(l)} &= \mathbf{X}^{(l)}_{\text{mid}} + \text{GLU}\!\left(\text{LN}(\mathbf{X}^{(l)}_{\text{mid}})\right),
\end{align*}
\noindent where $\text{LN}(\cdot)$ denotes the normalization applied prior to each sub-module. For notational \hbox{convenience}, we refer to normalized activations as $\tilde{\mathbf{X}}$ when needed.

The final representation $\mathbf{X}^{(L)}$ is projected into the vocabulary space via the unembedding matrix $\mathbf{W}_{UE} \in \mathbb{R}^{d \times |\mathcal{V}|}$, yielding the logits used for \hbox{next-token} prediction.

A key property enabling attribution analysis is the additive structure induced by residual connections around each sub-module, commonly referred to as the \emph{residual stream}. This structure allows the final representation to be expressed as:
\vspace{-0.5ex}
\[
\mathbf{X}^{(L)} = \mathbf{X}^{(0)} + \sum_{l=1}^{L} \Big(
\text{MHSA}(\tilde{\mathbf{X}}^{(l-1)}) + \text{GLU}(\tilde{\mathbf{X}}^{(l)}_{\text{mid}})
\Big).
\]
As a result, the contribution of any module output can be directly interpreted by projecting it into the vocabulary space via the unembedding matrix~\citep{elhage2021mathematical, nostalgebraist2020logit}.

\subsection{Root-Mean-Squared Normalization (RMSNorm)}
Root-Mean-Squared Normalization (RMSNorm) normalizes activations by their root-mean-square magnitude and is defined as:
\begin{equation}
    \text{RMSNorm}(\mathbf{X}) = 
    \frac{\mathbf{X}}{\sqrt{\mathrm{Var}[\mathbf{X}] + \epsilon}} \odot \boldsymbol{\gamma},
\end{equation}
\noindent where $\epsilon$ is a small constant for numerical stability and $\boldsymbol{\gamma}$ denotes learnable scaling parameters.

\subsection{Rotary Positional Embeddings (RoPE)}
Rotary Positional Embeddings (RoPE) encode \hbox{positional} information by applying a rotation to query and key vectors. Given a vector $x \in \mathbb{R}^d$ at position $m$, RoPE defines the rotated representation as:
\begin{equation}
    R(x, m) = \mathcal{R}_{\Theta, m} x,
\end{equation}
\noindent where $\mathcal{R}_{\Theta, m} \in \mathbb{R}^{d \times d}$ is a block-diagonal rotation matrix composed of $d/2$ independent $2 \times 2$ rotation blocks:
\begin{equation}
    \mathcal{R}_{\Theta, m} =
    \begin{pmatrix}
        M_1 &        & 0 \\
            & \ddots &   \\
        0   &        & M_{d/2}
    \end{pmatrix}.
\end{equation}
Each block $M_j$ rotates a pair of dimensions by an angle $m\theta_j$:
\begin{equation}
    M_j =
    \begin{pmatrix}
        \cos(m\theta_j) & -\sin(m\theta_j) \\
        \sin(m\theta_j) & \cos(m\theta_j)
    \end{pmatrix}.
\end{equation}
The rotation frequencies are defined as a geometric progression:
\begin{equation}
    \theta_j = 10000^{-2(j-1)/d}, 
    \quad j \in \{1, \dots, d/2\}.
\end{equation}

An important property of RoPE is that the inner product between a query at position $m$ and a key at position $n$ depends only on their relative offset:
\begin{equation}
    (R(q, m))^\top R(k, n) = q^\top \mathcal{R}_{\Theta, n-m} k.
\end{equation}
\section{Complexity Analysis}~\label{app:complexity_analysis}
As large language models are increasingly deployed on tasks requiring long context windows, the computational and memory scalability of attribution methods becomes a critical bottleneck. In this section, we analyze the complexity of Layer-wise Target Propagation (LTP) and compare it against two established baselines: Activation Patching (AP) and DePass~\citep{Hong2025DePassUF}. We evaluate these methods both theoretically and empirically, demonstrating that LTP uniquely decoupling attribution cost from the number of components, thereby enabling highly efficient dense attribution even over long sequences.

\subsection{Theoretical Complexity Analysis}
As summarized in Table~\ref{tab:complexity_comparison}, we evaluate the computational complexity of Layer-wise Target Propagation (LTP) against Activation Patching (AP) and DePass with respect to the number of attributed components $M$.

For a dense attribution task, AP requires $M$ independent forward passes. Because each pass incurs the $O(S^2)$ time complexity of the attention mechanism (where $S$ is the sequence length), its $O(M \cdot S^2)$ scaling becomes computationally prohibitive for long-context tasks. To mitigate this, methods like DePass leverage frozen activations through a modified forward pass, but still suffer from an $O(M)$ complexity bottleneck. DePass explicitly decomposes hidden representations into $M$ independent contribution vectors. For large models (e.g., Llama-3.1-8B-Instruct with $N_{\text{mlp}} \approx 14{,}000$), processing these simultaneously is intractable, necessitating sequential batching that yields $\lceil M / B \rceil$ forward passes (where $B$ is the batch size).

In contrast, LTP adopts a target-centric formulation that is completely independent of $M$. By propagating a single effective target vector backward from the output, LTP computes the attribution score for any component via a simple dot product with its cached contribution. Consequently, LTP requires $0$ additional forward passes and only a single backward pass, achieving a rigorous $O(1)$ time complexity with respect to $M$. This decoupling effectively eliminates the prohibitive scaling overhead inherent in forward decomposition-based methods.

\begin{table}[!tp]
    \centering
    \small
    \setlength{\tabcolsep}{5pt}
    \renewcommand{\arraystretch}{0.9}
    \begin{tabular}{lcc}
        \toprule
        \multirow{2}{*}{\textbf{Methods}} & \textbf{Complexity} & \textbf{Fwd passes} \\
        & \small{(w.r.t.\ $M$)} & \\
        \midrule
        Activation Patching (AP) & $O(M)$ & $M$ \\
        DePass & $O(M)$ & $\lceil M / B \rceil$ \\
        \midrule
        \textbf{LTP} & \textbf{O(1)} & \textbf{0} \\
        \bottomrule
    \end{tabular}
    \caption{
    Computational complexity comparison with respect to the number of attributed components $M$.
    \hbox{Activation} Patching and DePass require forward passes that scale linearly with $M$, whereas LTP performs attribution with no additional forward passes and a single backward pass.
    }
    \label{tab:complexity_comparison}
    \vspace{-2ex}
\end{table}

\subsection{Empirical Measurements Analysis}
We provide comprehensive empirical runtime and peak memory measurements across four distinct SwiGLU-based LLMs (representing the LLaMA, Qwen, and Mistral families) in Table~\ref{tab:empirical_measurements}. To ensure a strictly fair comparison, all evaluations were executed under identical hardware configurations, utilizing the same GPU setup and batch size. Furthermore, peak memory allocation was explicitly tracked using PyTorch's native CUDA memory allocator. This approach guarantees the precise measurement of tensor-level VRAM consumption directly managed by the model, effectively isolating the metric from OS-level caching overheads or external system processes.

The empirical results demonstrate that our LTP framework significantly outperforms existing baseline methods in computational speed across all scenarios. Even on short-context datasets like Known 1000, LTP operates consistently faster than Activation Patching (AP) and DePass across all evaluated models. Furthermore, LTP exhibits exceptional scalability and efficiency on longer context lengths, such as the SQuAD v2.0 dataset. For instance, on the Llama-3.1-8B-Instruct model, LTP completed the attribution calculations in just 129.05 seconds. In stark contrast, both AP and DePass required approximately 1.5 hours (5370.65 and 5452.28 seconds, respectively) for the same task, representing a massive 40$\times$ speedup for LTP. 

Crucially, LTP achieves these drastically faster runtimes while maintaining a peak memory footprint that is highly comparable to the baseline methods across all tested models. This confirms that LTP is a profoundly practical, robust, and resource-efficient solution for real-world tasks involving long context sequences, successfully overcoming the severe computational bottlenecks of traditional intervention-based methods.


\begin{table*}[t]
\centering
\resizebox{\textwidth}{!}{%
\begin{tabular}{lcccc}
\toprule
\multirow{2.5}{*}{\textbf{Methods}} & \multicolumn{2}{c}{\textbf{Known 1000}} & \multicolumn{2}{c}{\textbf{SQuAD v2.0}} \\
\cmidrule(lr){2-3} \cmidrule(lr){4-5}
& \textbf{Runtime (s)$\downarrow$} & \textbf{Peak Memory (MB)$\downarrow$}
& \textbf{Runtime (s)$\downarrow$} & \textbf{Peak Memory (MB)$\downarrow$} \\
\midrule

\multicolumn{5}{c}{Llama-3.1-8B-Instruct} \\
\midrule
Activation Patching (AP) & 14.88 & 20983.48 & 5370.65 & 15649.38 \\
DePass & 14.91 & 20663.91 & 5452.28 & 17210.68 \\
\textbf{Layer-wise Target Propagation (LTP)} & \textbf{4.98} & 21603.86 & \textbf{129.05} & 17512.21 \\
\midrule

\multicolumn{5}{c}{Llama-2-7B-Chat} \\
\midrule
Activation Patching (AP) & 17.58 & 24642.69 & 7229.39 & 13505.32 \\
DePass & 15.74 & 17321.58 & 6987.46 & 14976.61 \\
\textbf{Layer-wise Target Propagation (LTP)} & \textbf{5.33} & 19449.77 & \textbf{141.58} & 15026.57 \\
\midrule

\multicolumn{5}{c}{Qwen3-4B-Instruct-2507} \\
\midrule
Activation Patching (AP) & 10.32 & 15354.79 & 7336.73 & 7863.08 \\
DePass & 9.04 & 13866.60 & 4905.13 & 8591.08 \\
\textbf{Layer-wise Target Propagation (LTP)} & \textbf{4.00} & 13985.03 & \textbf{119.43} & 9571.57 \\
\midrule

\multicolumn{5}{c}{Mistral-7B-Instruct-v0.3} \\
\midrule
Activation Patching (AP) & 17.33 & 18395.77 & 9603.84 & 13973.31 \\
DePass & 16.87 & 19795.21 & 10808.99 & 14814.81 \\
\textbf{Layer-wise Target Propagation (LTP)} & \textbf{5.55} & 21182.21 & \textbf{170.92} & 16191.69 \\

\bottomrule
\end{tabular}%
}
\caption{Empirical runtime and peak memory measurements across various SwiGLU-based LLMs (representing the LLaMA, Qwen, and Mistral families) on datasets with different context lengths. Notably, LTP achieves vastly superior computational efficiency, delivering drastically faster runtimes while maintaining a comparable peak memory footprint to the baseline methods. It maintains highly robust performance even on the long-context SQuAD v2.0 dataset, successfully overcoming the severe computational bottlenecks of AP and DePass.}
\label{tab:empirical_measurements}
\end{table*}

\section{Algorithm of Layer-wise Target Propagation}~\label{app:dpa_algorithm}
The complete procedure for Layer-wise Target Propagation (LTP) is formalized in Algorithm~\ref{alg:dpa}. The framework operates in a two-stage process: a single forward pass for activation caching followed by a recursive top-down target propagation. By decoupling the attribution cost from the number of attributed components $M$, LTP achieves $O(1)$ complexity, making it highly scalable for dense attribution tasks in large-scale transformers.

\begin{algorithm*}[t]
    \caption{The algorithm of Layer-wise Target Propagation (LTP).}
    \label{alg:dpa}
    \setstretch{1.1}
    \centering
    \begin{minipage}{0.90\textwidth}
        \SetKwInOut{Input}{Input}
        \SetKwInOut{Output}{Output}
        \SetKwComment{Comment}{// }{}

        \Input{Input tokens $X$, target logit direction $t^{(L)}$, sensitivity parameter $p$}
        \Output{Attribution scores $S_t$ for all components}

        \BlankLine
        \Comment{Stage 1: Forward Pass and Activation Caching}
        Execute a single forward pass through $L$ layers\;
        \For{each layer $l = 1$ \KwTo $L$}{
            Cache attention weights $\alpha^{(h)}$ and normalized inputs $\tilde{x}$\;
            Cache GLU pre-activations $s^{(n)}$ and gate activations $\alpha^{(n)}$\;
        }

        \BlankLine
        \Comment{Stage 2: Top-Down Target Propagation}
        Initialize effective target $t^{(L)} \leftarrow W_{U}^{(t)}$\;
        \For{each layer $l = L$ \KwTo 1}{
            \Comment{1. Propagate through Gated Linear Unit (GLU)}
            Compute $g_{up}^{(n)}(t)$ and $g_{gate}^{(n)}(t)$ using linearized inverse rules\;
            $t_{mid}^{(l)} \leftarrow t^{(l)} + \sum_{n} (\mu_{up} \cdot g_{up}^{(n)}(t^{(l)}) + \mu_{gate} \cdot g_{gate}^{(n)}(t^{(l)}))$\;
            
            \BlankLine
            \Comment{2. Propagate through Multi-Head Self-Attention (MHSA)}
            \For{each head $h$}{
                Compute value path $g_{V}^{(h)}(t_{mid}^{(l)})$ and control paths $g_{Q}^{(h)}, g_{K}^{(h)}$\;
                $g_{head}^{(h)} \leftarrow \mu_{Q}g_{Q}^{(h)} + \mu_{K}g_{K}^{(h)} + \mu_{V}g_{V}^{(h)}$\;
            }
            $t^{(l-1)} \leftarrow t_{mid}^{(l)} + \sum_{h} g_{head}^{(h)}$\;
        }

        \BlankLine
        \Comment{Stage 3: Score Calculation}
        \Comment{Calculate scores for each hierarchical level}
        $S_{t}(x_i^{(0)}) \leftarrow (x_i^{(0)})^\top t_i^{(0)}$ \Comment*[r]{Input Token Attribution}
        $S_{t}(h^{(l)}) \leftarrow \text{Attn}^{(h)}(\tilde{x}^{(l-1)}) \cdot t_{mid}^{(l)}$ \Comment*[r]{Attention Head Attribution}
        $S_{t}(n^{(l)}) \leftarrow \text{GLU}^{(n)}(\tilde{x}_{mid}^{(l)}) \cdot t^{(l)}$ \Comment*[r]{MLP Neuron Attribution}
        \Return $S_t$\;
    \end{minipage}
\end{algorithm*}
\section{Derivations of LTP Target Tracing}~\label{app:dpa_derivations}
In this section, we provide the complete mathematical derivations for the target propagation rules underlying Layer-wise Target Propagation (LTP). By propagating a single effective target vector backward through the network, LTP analytically computes the gradient of the attribution score with respect to the intermediate representations. Below, we derive these exact target propagation formulas for the two primary computational blocks of the Transformer architecture: the Gated Linear Unit (GLU) and the Multi-Head Self-Attention (MHSA) module.
\subsection{Gated Linear Unit (GLU) Target Propagation}
\paragraph{Forward Pass Definition.} Consider a single neuron $n$ within the GLU layer. The contribution of this neuron to the residual stream is defined as:
$$\text{GLU}^{(n)}(\tilde{\mathbf{x}}) = \left( \sigma(\tilde{\mathbf{x}} \mathbf{w}_G^{(n)}) \odot (\tilde{\mathbf{x}} \mathbf{w}_U^{(n)}) \right) \mathbf{w}_D^{(n)}$$
where:
\begin{itemize}
    \item $\sigma(\cdot)$ is the SiLU (Swish) activation function.
    \item $\mathbf{w}_G^{(n)}, \mathbf{w}_U^{(n)}$ are the column vectors of the gate and up projection matrices.
    \item $\mathbf{w}_D^{(n)}$ is the row vector of the down projection matrix.
    \item $\odot$ denotes element-wise multiplication (scalar here for a single neuron).
\end{itemize}
The attribution of this neuron to a global target $\mathbf{t}$ is given by:$$S^{(n)} = \text{GLU}^{(n)}(\tilde{\mathbf{x}}) \cdot \mathbf{t}$$$$S^{(n)} = \underbrace{\sigma(\tilde{\mathbf{x}} \mathbf{w}_G^{(n)})}_{\text{activation } \alpha} \cdot \underbrace{(\tilde{\mathbf{x}} \mathbf{w}_U^{(n)})}_{\text{content } v} \cdot \underbrace{(\mathbf{w}_D^{(n)} \mathbf{t})}_{\text{scalar } \lambda}$$

\subsection{Up-Projection (Value) Pathway}
To identify the contribution of the information-carrying pathway, we linearize the operation by freezing the gate activation $\alpha^{(n)}$. The attribution becomes linear with respect to $\tilde{\mathbf{x}}$:
$$S^{(n)}_{up} \approx \alpha^{(n)} \cdot (\tilde{\mathbf{x}} \mathbf{w}_U^{(n)}) \cdot \lambda$$
The gradient of this score with respect to the input $\tilde{\mathbf{x}}$ (which represents the propagated target) is:
$$\mathbf{g}_{up}^{(n)}(\mathbf{t}) = \nabla_{\tilde{\mathbf{x}}} S^{(n)}_{up} = \alpha^{(n)} \lambda \mathbf{w}_U^{(n)}$$
Incorporating the RMSNorm scaling parameters $\boldsymbol{\gamma}$ into the effective weight $\tilde{\mathbf{w}}$, we get the paper's formulation:
$$\boxed{\mathbf{g}_{up}^{(n)}(\mathbf{t}) = (\tilde{\mathbf{w}}_{U}^{(n)}) \cdot \alpha^{(n)} \cdot (\mathbf{w}_{D}^{(n)} \mathbf{t})}$$

\subsection{Gate-Projection Pathway}
For the gating pathway, we freeze the content term $v = \tilde{\mathbf{x}} \mathbf{w}_U^{(n)}$ and attribute through the non-linear activation $\sigma(s)$, where $s = \tilde{\mathbf{x}} \mathbf{w}_G^{(n)}$.$$S^{(n)}_{gate} \approx \sigma(s) \cdot v \cdot \lambda$$
Using the linear approximation $\sigma(s) \approx s \cdot \frac{\sigma(s)}{s} = s \cdot \frac{\alpha^{(n)}}{s}$, we rewrite the score:$$S^{(n)}_{gate} \approx (\tilde{\mathbf{x}} \mathbf{w}_G^{(n)}) \cdot \frac{\alpha^{(n)}}{s} \cdot v \cdot \lambda$$
Taking the gradient with respect to $\tilde{\mathbf{x}}$:$$\mathbf{g}_{gate}^{(n)}(\mathbf{t}) = \nabla_{\tilde{\mathbf{x}}} S^{(n)}_{gate} = \mathbf{w}_G^{(n)} \cdot \frac{\alpha^{(n)}}{s} \cdot v \cdot \lambda$$
Grouping terms yields the final propagation formula:$$\boxed{\mathbf{g}_{gate}^{(n)}(\mathbf{t}) = (\tilde{\mathbf{w}}_{G}^{(n)}) \cdot \frac{\alpha^{(n)}}{s} \cdot \left( v \cdot \mathbf{w}_{D}^{(n)} \mathbf{t} \right)}$$

\subsection{Multi-Head Self-Attention (MHSA) Target Propagation}
\paragraph{Forward Pass Definition.} The output of an attention head $h$ at query position $i$ is:
$$\mathbf{y}_i^{(h)} = \sum_{j \le i} \alpha_{ij} (\tilde{\mathbf{x}}_j \mathbf{W}_V^{(h)}) \mathbf{W}_O^{(h)}$$
where the attention weights $\alpha_{ij}$ are computed via Softmax over the scaled dot-product of Rotary Positional Embedding (RoPE) transformed queries and keys:$$\alpha_{ij} = \text{Softmax}_j(s_{ij})$$$$s_{ij} = \frac{R_Q(\tilde{\mathbf{x}}_i \mathbf{W}_Q^{(h)}) \cdot R_K(\tilde{\mathbf{x}}_j \mathbf{W}_K^{(h)})}{\sqrt{d_h}}$$
The attribution to the target $\mathbf{t}$ is $S = \mathbf{y}_i^{(h)} \cdot \mathbf{t}$.

\paragraph{Value Pathway.} We assume the attention pattern $\alpha_{ij}$ is fixed. The operation becomes linear with respect to the input $\tilde{\mathbf{x}}_j$ via the value matrix $\mathbf{W}_V$:$$S_{value} = \sum_{j \le i} \alpha_{ij} (\tilde{\mathbf{x}}_j \mathbf{W}_V^{(h)}) (\mathbf{W}_O^{(h)} \mathbf{t})$$
The propagated target to token $j$ is derived by taking the gradient w.r.t $\tilde{\mathbf{x}}_j$:$$\boxed{\mathbf{g}_{V, j}^{(h)}(\mathbf{t}) = \tilde{\mathbf{W}}_V^{(h)} \left( \sum_{i \ge j} \alpha_{ij} (\mathbf{W}_O^{(h)} \mathbf{t}) \right)}$$

\paragraph{Query and Key Pathways (Softmax Decomposition).} To propagate through the Query and Key paths, we must differentiate the Softmax function. Let the projected target vector be $\mathbf{t}' = \mathbf{W}_O^{(h)} \mathbf{t}$ and the value vector be $\mathbf{v}_j = \tilde{\mathbf{x}}_j \mathbf{W}_V^{(h)}$. The attribution score is:$$S = \sum_{k} \alpha_{ik} (\mathbf{v}_k \cdot \mathbf{t}')$$
We first find the sensitivity of $S$ with respect to the pre-softmax logits $s_{ij}$. Using the derivative of Softmax, $\frac{\partial \alpha_{ik}}{\partial s_{ij}} = \alpha_{ij}(\delta_{jk} - \alpha_{ik})$:$$\begin{aligned} \frac{\partial S}{\partial s_{ij}} &= \sum_{k} \frac{\partial \alpha_{ik}}{\partial s_{ij}} (\mathbf{v}_k \cdot \mathbf{t}') \\ &= \sum_{k} \alpha_{ij}(\delta_{jk} - \alpha_{ik}) (\mathbf{v}_k \cdot \mathbf{t}') \\ &= \alpha_{ij} \left( \mathbf{v}_j \cdot \mathbf{t}' - \left(\sum_{k} \alpha_{ik} \mathbf{v}_k\right) \cdot \mathbf{t}' \right) \end{aligned}$$
Defining the expected head output as $\boldsymbol{\mu}_i^{(h)} = \sum_{k} \alpha_{ik} \mathbf{v}_k$, we obtain the scalar interaction term $\delta_{ij}$:$$\delta_{ij} = \alpha_{ij} \big(\mathbf{v}_j - \boldsymbol{\mu}_i^{(h)} \big) \cdot (\mathbf{W}_O^{(h)} \mathbf{t})$$
Now we propagate $\delta_{ij}$ through the dot product $s_{ij}$.

\paragraph{Query Path.} The gradient of $s_{ij}$ with respect to the query input vector $\mathbf{q}_i = \tilde{\mathbf{x}}_i \mathbf{W}_Q$ is proportional to the key vector $\mathbf{k}_j$.$$\nabla_{\tilde{\mathbf{x}}_i} s_{ij} = \mathbf{W}_Q^{(h)} R_{Q,i}^{-1} \left( \frac{R_K(\tilde{\mathbf{x}}_j \mathbf{W}_K^{(h)})}{\sqrt{d_h}} \right)$$
Aggregating over all $j$, the target for the query input at position $i$ is:$$\boxed{\mathbf{g}_{Q, i}^{(h)}(\mathbf{t}) = \tilde{\mathbf{W}}_Q^{(h)} \sum_{j \le i} \delta_{ij} \cdot R_{Q,i}^{-1} \left( \frac{R_{K}(\tilde{\mathbf{x}}_j \mathbf{W}_K^{(h)})}{\sqrt{d_h}} \right)}$$

\paragraph{Key Path.} Similarly, the gradient of $s_{ij}$ with respect to the key input vector $\mathbf{k}_j = \tilde{\mathbf{x}}_j \mathbf{W}_K$ is proportional to the query vector $\mathbf{q}_i$.$$\nabla_{\tilde{\mathbf{x}}_j} s_{ij} = \mathbf{W}_K^{(h)} R_{K,j}^{-1} \left( \frac{R_Q(\tilde{\mathbf{x}}_i \mathbf{W}_Q^{(h)})}{\sqrt{d_h}} \right)$$
Aggregating over all $i$ that attend to $j$ (i.e., $i \ge j$ in causal attention), the target for the key input at position $j$ is:$$\boxed{\mathbf{g}_{K, j}^{(h)}(\mathbf{t}) = \tilde{\mathbf{W}}_K^{(h)} \sum_{i \ge j} \delta_{ij} \cdot R_{K,j}^{-1} \left( \frac{R_{Q}(\tilde{\mathbf{x}}_i \mathbf{W}_Q^{(h)})}{\sqrt{d_h}} \right)}$$
\section{Dataset Configurations}~\label{app:dataset_configurations}
In this section, we provide detailed information regarding the licenses, usage, and preprocessing steps for the datasets used to evaluate the Layer-wise Target Propagation (LTP) framework. We conduct experiments across four representative tasks: factual knowledge, reading comprehension, sentiment analysis, and circuit discovery.
\subsection{Licenses and Usage}
All datasets utilized in this study are publicly accessible and were used in strict accordance with their respective licensing terms for research purposes.
\begin{itemize}
    \item \textbf{Known 1000.} This dataset is distributed under the MIT License. It consists of 1,000 factual statements designed to evaluate how language models represent and utilize factual associations.
    \item \textbf{SQuAD v2.0.} The Stanford Question Answering Dataset is licensed under Creative Commons Attribution-ShareAlike 4.0 International (CC BY-SA 4.0). It is a standard benchmark for reading comprehension, requiring the model to identify answer spans within a given context or determine if a question is unanswerable.
    \item \textbf{IMDb Sentiment.} Usage of this dataset follows the IMDb Non-Commercial Licensing terms, which allow for limited personal and non-commercial use. In compliance with these terms, we acknowledge that the information is courtesy of IMDb (https://www.imdb.com) and is used with permission. The data was sourced from official contributor datasets to ensure adherence to copyright and conditions of use. It contains movie reviews labeled for binary sentiment analysis.
    \item \textbf{Indirect Object Identification (IOI).} This dataset is distributed under the MIT License. It serves as a specialized benchmark for component-level circuit discovery.
\end{itemize}

\subsection{Dataset Preprocessing Details}
To maintain consistent evaluation signals and manage computational complexity, we used a combined dataset of the Test and Dev sets from the experiments. We used 1,000 samples only for the IMDB dataset, while using the total combined sample size for the rest. We performed systematic preprocessing as follows:
\begin{itemize}
    \item \textbf{Known 1000.} We specifically selected instances where the model provided a correct baseline prediction for the target entity to ensure that the attribution was grounded in existing model knowledge. The data was then formatted as a prefix-completion task, ensuring that target entities consisted of short token spans to facilitate precise attribution measurement.
    \item \textbf{SQuAD v2.0.} Our processing focused exclusively on answerable questions to evaluate the model's ability to attribute information to the provided context. The preprocessing involved normalizing whitespace and special characters while ensuring all questions concluded with a proper question mark. Furthermore, we utilized the answer\_start index to establish exact ground-truth tokens and filtered for answers consisting of no more than two words to ensure the localization signal remained clear and unambiguous.
    \item \textbf{IMDb Sentiment.} We sampled a balanced subset of 1,000 reviews, comprising 500 positive and 500 negative samples to prevent sentiment bias. The text was cleaned by removing HTML tags and collapsing redundant spaces to prevent tokenization artifacts. To ensure sufficient context for attribution while maintaining computational efficiency, we filtered reviews to maintain a character length between 200 and 500.
    \item \textbf{Indirect Object Identification (IOI).} This dataset was constructed using fixed linguistic templates to isolate the specific circuit responsible for identifying indirect objects. Our preprocessing focused on identifying the precise token positions of the subjects and indirect objects within these templates, which allowed us to measure the direct and indirect effects of internal components with high resolution.
\end{itemize}
\section{Experiment Reproducibility}~\label{app:reproducibility}
\begin{figure*}[!t]
    \centering
    \subfigure{
        \includegraphics[width=.71\linewidth]{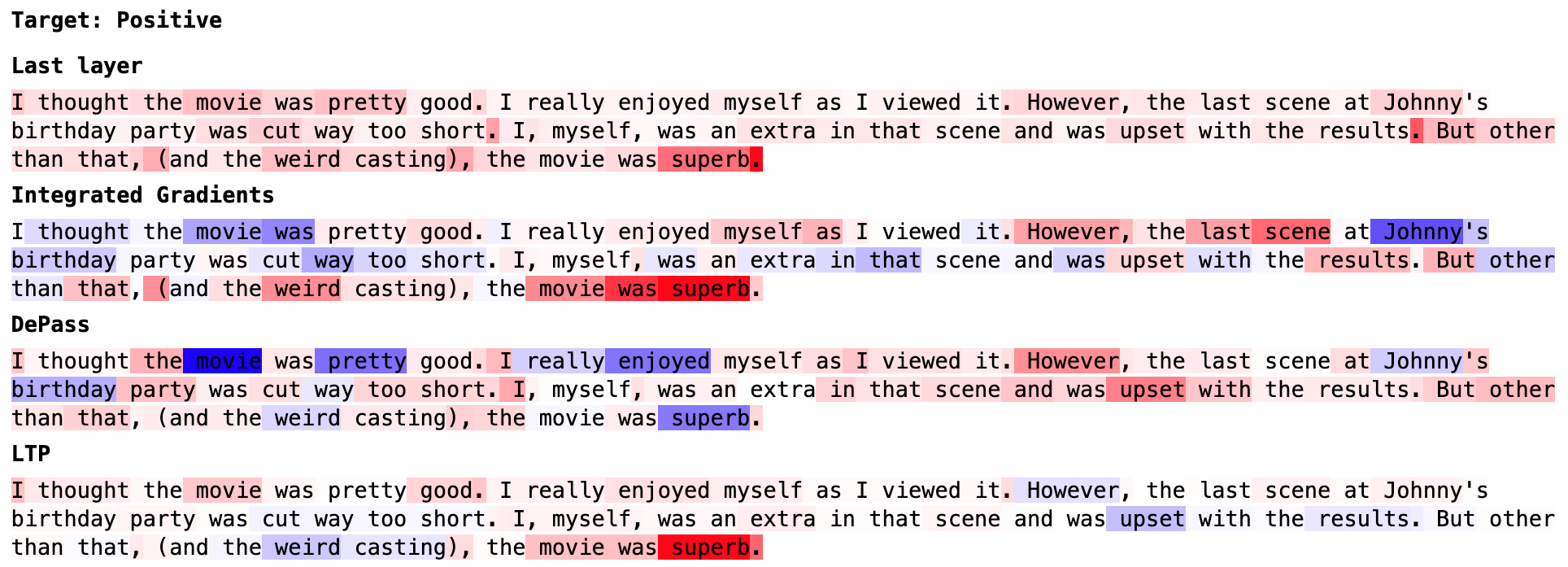}
    }
    \subfigure{
        \includegraphics[width=.26\linewidth]{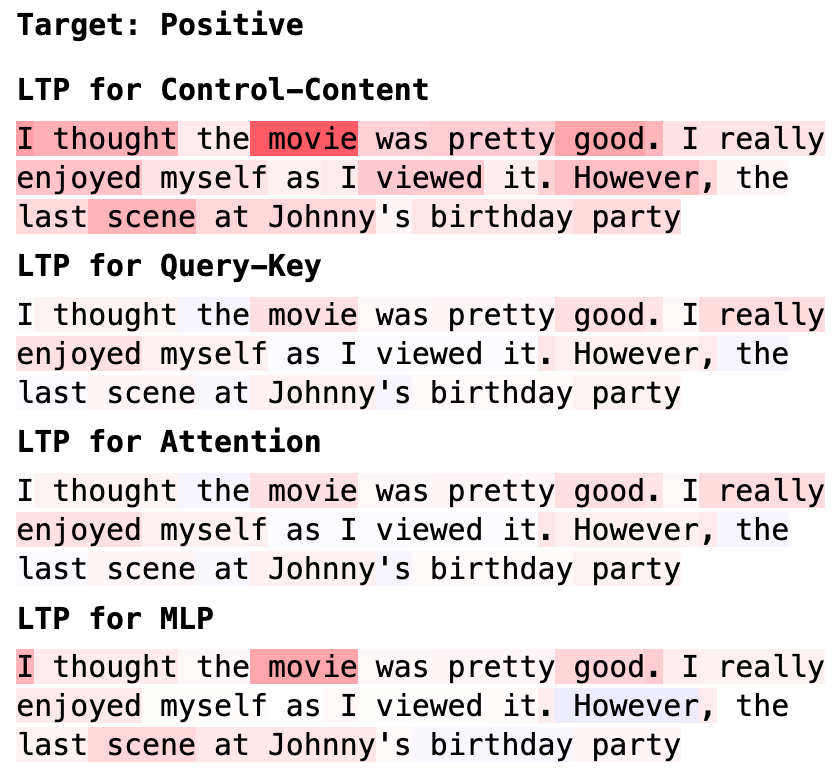}
    }
    \vspace{-.1in}
    \caption{Qualitative attribution examples. Red highlights denote tokens that positively contribute to the target, while blue denotes tokens that hinder it. As shown on the IMDb dataset (left), LTP more precisely distinguishes between helpful and harmful context compared to baseline methods. The right panel illustrates LTP's attribution across different configurations.}
    \label{fig:attribution_examples}
\end{figure*}
\begin{table*}[!tp]
    \centering
    \small
    \setlength{\tabcolsep}{3pt}
    \renewcommand{\arraystretch}{0.9}

\resizebox{\textwidth}{!}{%
    \begin{tabular}{l@{\hskip 10pt}ccccccc@{\hskip 14pt}ccccccc}
        \toprule
        \multirow{2.5}{*}{\textbf{Sensitivity}}
        & \multicolumn{7}{c}{\textbf{IMDb} (input-level, $p$)}
        & \multicolumn{7}{c}{\textbf{IOI} (component-level, $p$)} \\
        \cmidrule(lr){2-8}\cmidrule(lr){9-15}
        & 0.0 & 0.2 & 0.4 & 0.5 & 0.6 & 0.9 & 1.0
        & 0.0 & 0.1 & 0.2 & 0.3 & 0.5 & 0.7 & 1.0 \\
        \midrule
        Control-Content & 6.54 & 24.80 & 36.78 & \textbf{38.26} & 37.64 & 27.72 & 27.44 & 84.10 & 99.70 & 112.95 & \textbf{113.53} & 87.74 & 40.42 & 30.04 \\
        Query-Key & 35.74 & 36.88 & 38.21 & 38.26 & 38.58 & \textbf{40.10} & 39.68 & 91.82 & 97.13 & \textbf{101.98} & 100.50 & 87.74 & 73.01 & 49.87 \\
        Attention & 6.78 & 26.74 & 36.91 & 38.26 & \textbf{38.31} & 32.21 & 28.37 & 54.07 & 53.83 & 60.52 & 70.22 & \textbf{87.74} & 57.35 & 29.14 \\
        MLP & 32.42 & 38.06 & \textbf{38.88} & 38.26 & 36.93 & 29.56 & 26.15 & 144.35 & \textbf{153.37} & 148.56 & 132.52 & 87.74 & 56.83 & 45.04 \\
        \bottomrule
    \end{tabular}%
}
    \caption{Extended sensitivity analysis for \textbf{input-level} attribution on IMDb (Llama-3.1-8B-Instruct) and \textbf{component-level} attribution on the IOI (Llama-3.1-8B-Instruct). We report the Total AUC (recovery $-$ disruption; higher is better) for each weighting family in Table~\ref{tab:sensitivity_analysis}. At $p=0.5$, all families coincide with the balanced configuration. \textbf{Bold} indicates the best result per family and setting.}
    \label{tab:extended_sensitivity}
\end{table*}
All experiments, including model inference and the recursive propagation of Layer-wise Target Propagation (LTP) targets, were conducted on a high-performance computing cluster to ensure a consistent and high-throughput execution environment.  The specific hardware and software configurations used for our empirical evaluations are as follows :
\begin{itemize}
    \item \textbf{Hardware Resources.} We utilized an NVIDIA DGX system equipped with NVIDIA A100-SXM4-40GB GPUs. Each GPU provides 40 GB of HBM2 memory, which is essential for caching the high-dimensional activations required for attribution across large-scale models like Llama-3-8B-Instruct. The high memory bandwidth of the A100 architecture facilitated efficient processing of dense attribution tasks.
    \item \textbf{Software Environment.} The system operated on CUDA Version 12.2 with NVIDIA Driver Version 535.161.08. This environment supports the latest optimization libraries required to execute transformer-based models and perform efficient tensor operations during the target propagation stage.
    \item \textbf{Implementation Details.} The LTP framework and all comparative baselines, such as Integrated Gradients and DePass, were implemented using the PyTorch library, and the results were described based on a single execution.
\end{itemize}
\section{Qualitative Attribution Examples}~\label{app:qualitative}
To complement the quantitative faithfulness results in Section~\ref{sec:experiments}, we provide qualitative attribution visualizations that illustrate the behavior of Layer-wise Target Propagation (LTP) on natural language inputs. Figure~\ref{fig:attribution_examples} presents two complementary views.
\paragraph{Token-level Attribution on IMDb (left).}
The left panel of Figure~\ref{fig:attribution_examples} compares LTP against representative baselines on an IMDb sentiment classification example. Red highlights mark tokens that positively contribute to the predicted sentiment, while blue marks tokens whose presence pushes the prediction in the opposite direction. LTP produces sharper polarity contrasts and more selectively highlights sentiment-bearing tokens---such as explicit affective adjectives and negation cues---than baselines that diffuse importance across syntactic context. This aligns with the quantitative observation in Table~\ref{tab:input_faithfulness} that attribution to complex semantic concepts like sentiment exposes the limitations of attention-only approaches: while raw attention captures token-to-token mixing, it does not by itself signal the semantic polarity that drives the final prediction.

\paragraph{Pathway Sensitivity (right).}
The right panel compares LTP's attribution maps on the same IMDb input under the four parameter combinations summarized in Table~\ref{tab:sensitivity_analysis}---Control--Content, Query--Key, Attention, and MLP---each evaluated at the endpoint $p = 1$, which routes attribution through the corresponding pathway in its purest form. Despite sharing the same input and target, the four maps highlight visibly different token subsets, making the relative contribution of each pathway concrete and complementing the quantitative trends in Figure~\ref{fig:sensitivity_analysis}.

\section{Extended Sensitivity Analysis}~\label{app:extended_sensitivity}
The choice of the bilinear weighting parameters $\mu$ is first and foremost motivated by the conservation property, which requires that the relevances of the inputs of a bilinear function sum to the relevance at its output. Among the values satisfying this property, we adopt $p=0.5$ because it corresponds to balanced propagation across both inputs and is therefore the natural choice absent any task-specific prior. The sensitivity analysis in Section~\ref{sec:sensitivity_analysis} supports this choice post-hoc; in this section, we strengthen the argument by extending the analysis to fundamentally different tasks.

Concretely, we run the full $p$-sweep ($p \in [0, 1]$) on two settings that differ from the main-paper analysis in both task and granularity: IMDb (input-level attribution task) and IOI (component-level attribution task). Following the weighting families defined in Table~\ref{tab:sensitivity_analysis}, we report the Total AUC (recovery - disruption; higher is better) per family. Table~\ref{tab:extended_sensitivity} presents the results for both settings.

The results first support $p=0.5$ as a solid choice regardless of the task: the balanced configuration remains at or near the peak of every family in the input-level setting and avoids the sharp degradation observed at the extremes of each sweep. Beyond this, the extended experiments reveal task-consistent properties. For input-level attribution, a slightly higher $p$ on the Query-Key interaction leads to the best results, whereas for component-level attribution, the best results are achieved with a lower $p$ on the MLP gate-up interaction. These gains, which are substantial especially for component-level attribution, suggest that task-adaptive weighting is a promising direction for future work.

\section{Generalizability}~\label{app:extended_results}
In this section, we evaluate the generalizability of Layer-wise Target Propagation (LTP) across different model architectures to ensure our findings are robust and not strictly tied to a single model family.

\subsection{Generalization within the LLaMA Family}~\label{app:llama2_results}
We initially extended our experiments to Llama-2-7B-Chat not only Llama-3.1-8B-Instruct. The empirical results demonstrate that LTP maintains consistent attribution accuracy and computational efficiency within the LLaMA architecture.

\begin{figure*}[!t]
    \centering
        \includegraphics[width=2\columnwidth]{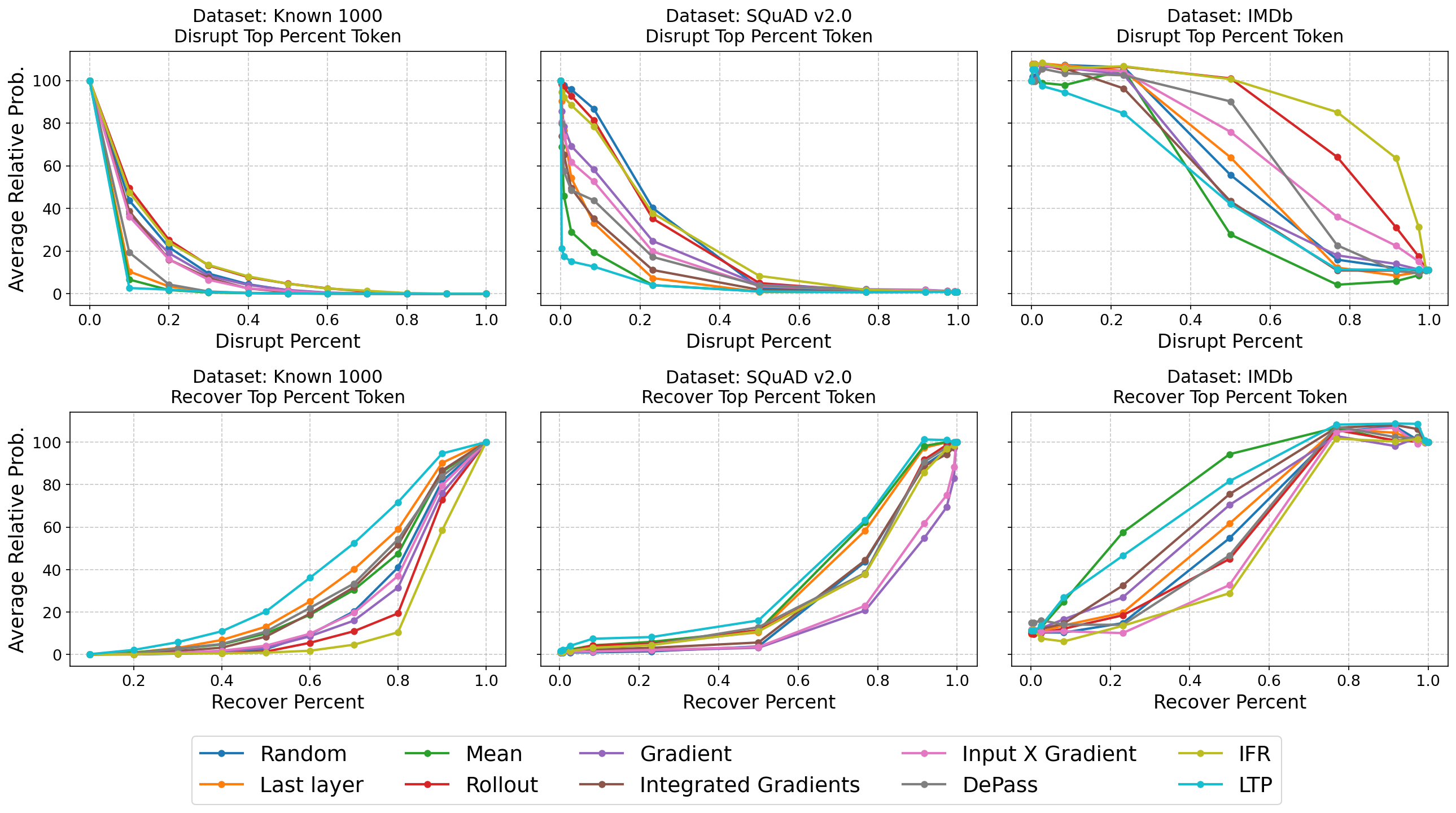}
    \vspace{-.1in}
    \caption{Performance comparison with baseline attribution methods on \textbf{Llama-2-7B-Chat} under different top input token masking strategies. Our approach shows \emph{lower} Top-$k$ disruption and \emph{higher} Top-$k$ recovery, indicating more accurate identification of important components.}
    ~\label{fig:input_faithfulness_llama2}
    \vspace{-.1in}
\end{figure*}

\begin{figure*}[!tp]
    \centering
        
        \includegraphics[width=2\columnwidth]{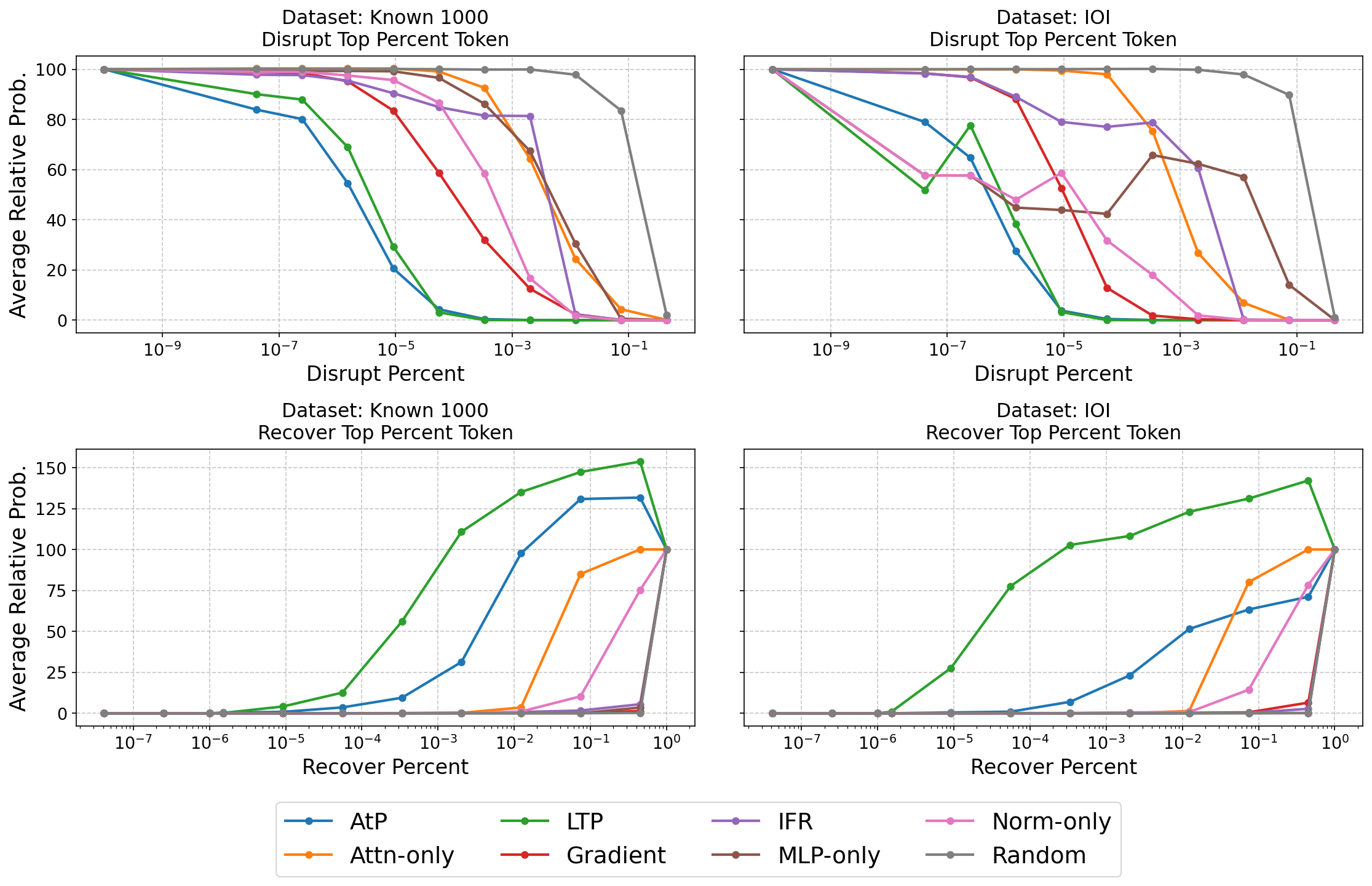}
    \caption{Performance comparison with baseline attribution methods on \textbf{Llama-2-7B-Chat} under different top model component masking strategies. Our approach shows lower Top-k disruption and higher Top-k recovery, indicating more accurate identification of important components.}
    ~\label{fig:component_faithfulness_llama2}
    \vspace{-.2in}
\end{figure*}

The input and component attribution results for Llama-2-7B-Chat are shown in Table~\ref{tab:input_faithfulness_llama2}, and ~\ref{tab:component_faithfulness_llama2} (with corresponding Figures~\ref{fig:input_faithfulness_llama2} and \ref{fig:component_faithfulness_llama2}).

\subsection{Generalization to Non-LLaMA SwiGLU Architectures}
To address concerns regarding generalization to other model families, we conducted additional experiments on the Qwen and Mistral model families, specifically Qwen3-4B-Instruct-2507, Qwen2.5-32B-Instruct, and Mistral-7B-Instruct-v0.3. While these models share a SwiGLU-based Transformer backbone with the LLaMA series, they are fundamentally distinct. They feature structural variations in the attention mechanism (e.g., Sliding Window Attention in Mistral), utilize different tokenizers, and undergo entirely different pre- and post-training pipelines.

\begin{figure*}[!t]
    \centering
        
        \includegraphics[width=2\columnwidth]{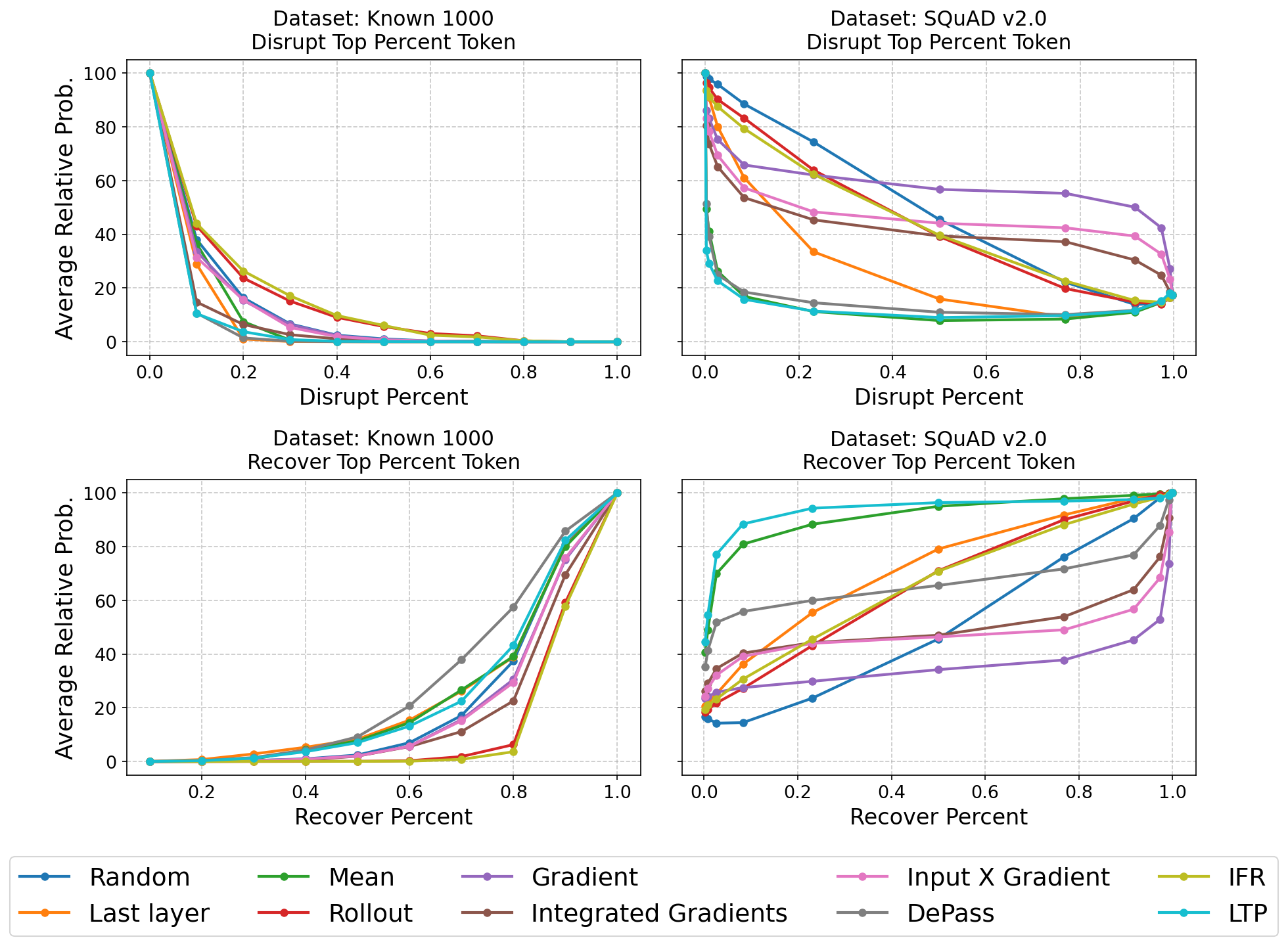}
    \vspace{-.1in}
    \caption{Performance comparison with baseline attribution methods on \textbf{Qwen3-4B-Instruct-2507} under different top input token masking strategies. Our approach shows \emph{lower} Top-$k$ disruption and \emph{higher} Top-$k$ recovery, indicating more accurate identification of important components.}
    ~\label{fig:input_faithfulness_qwen3}
    \vspace{-.1in}
\end{figure*}

\begin{figure*}[!tp]
    \centering
        
        \includegraphics[width=2\columnwidth]{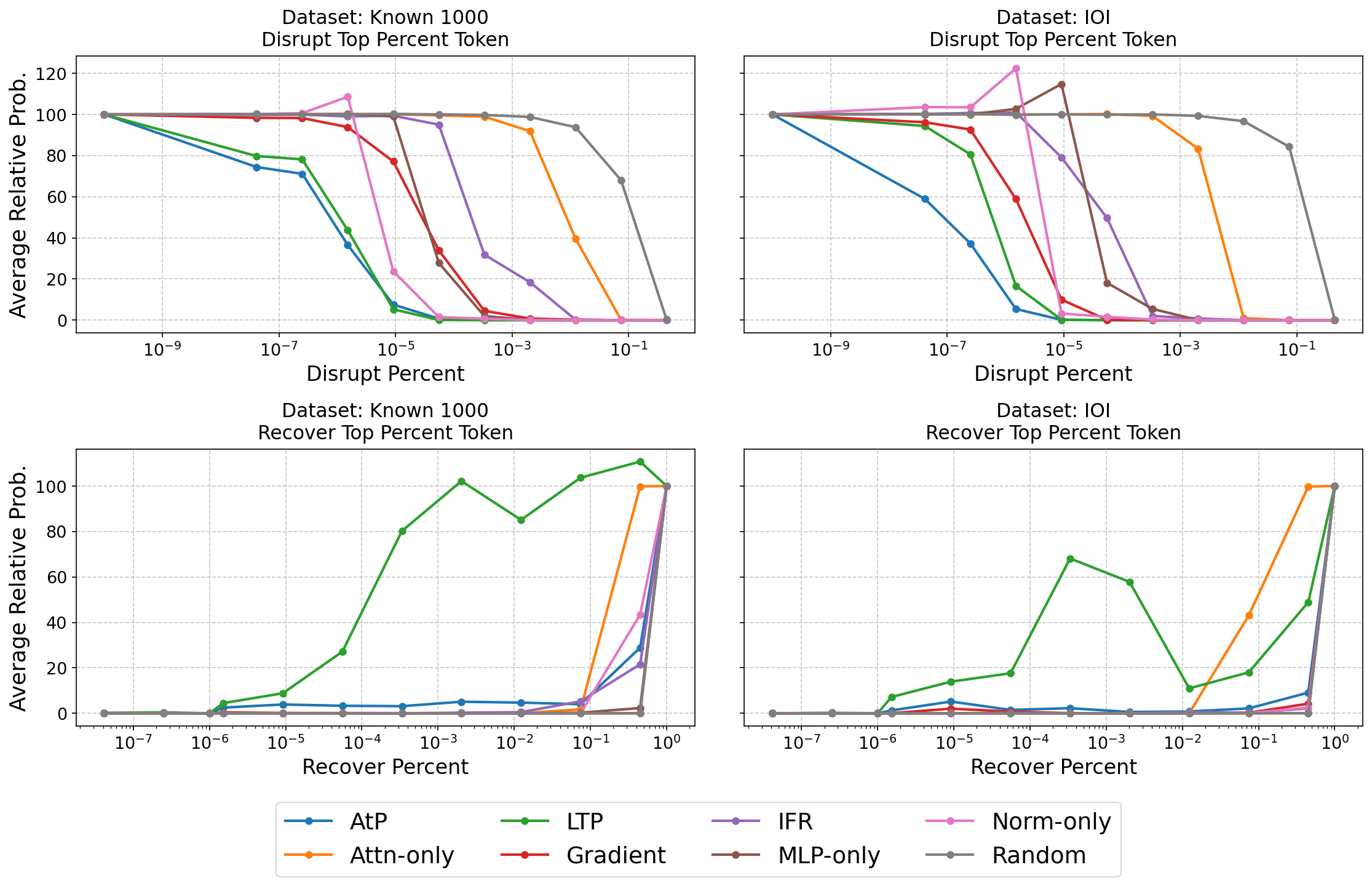}
    \caption{Performance comparison with baseline attribution methods on \textbf{Qwen3-4B-Instruct-2507} under different top model component masking strategies. Our approach shows lower Top-k disruption and higher Top-k recovery, indicating more accurate identification of important components.}
    ~\label{fig:component_faithfulness_qwen3}
    \vspace{-.2in}
\end{figure*}

\begin{figure*}[!t]
    \centering
        
        \includegraphics[width=2\columnwidth]{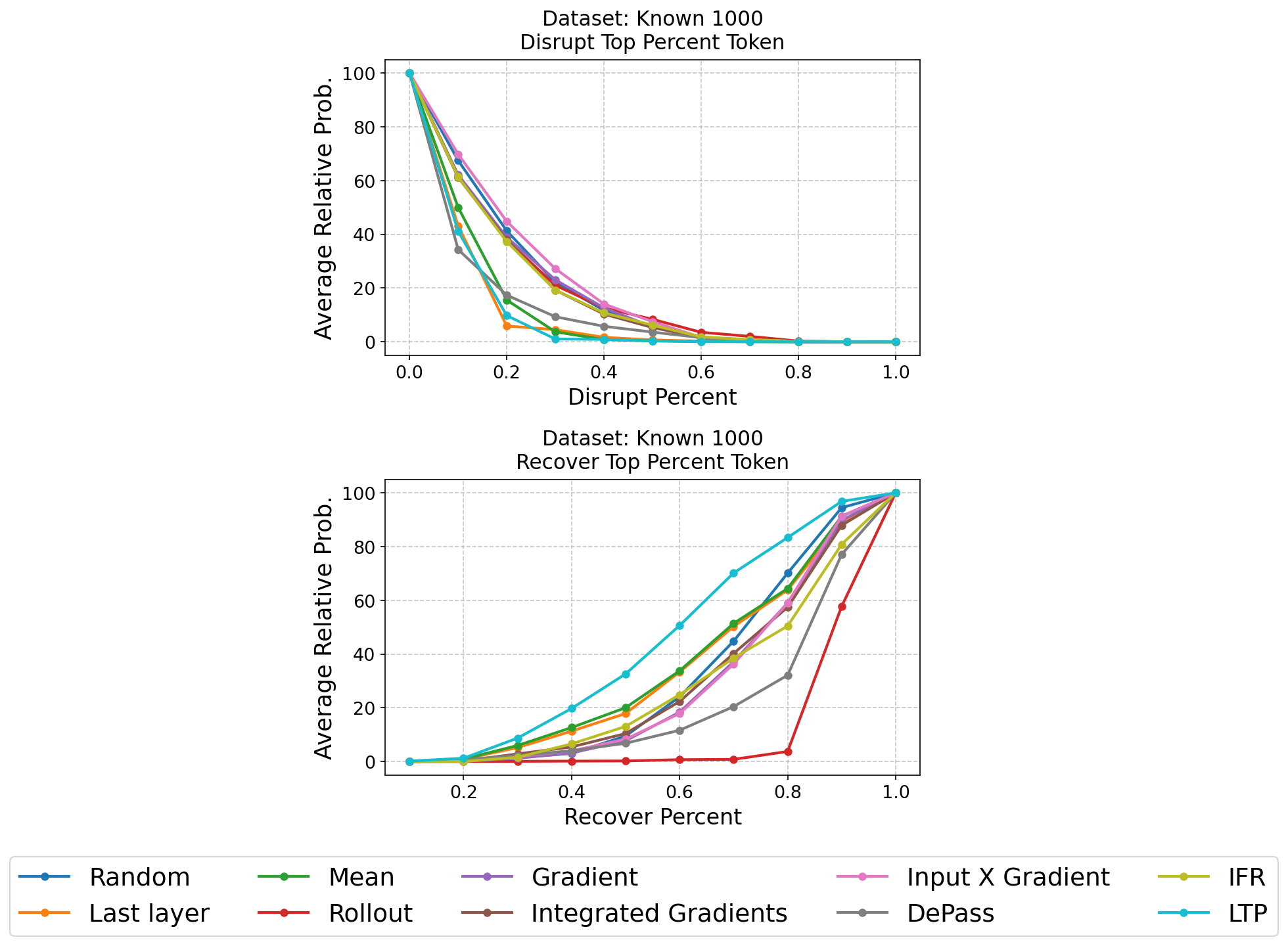}
    \vspace{-.1in}
    \caption{Performance comparison with baseline attribution methods on \textbf{Qwen2.5-32B-Instruct} under different top input token masking strategies. Our approach shows \emph{lower} Top-$k$ disruption and \emph{higher} Top-$k$ recovery, indicating more accurate identification of important components.}
    ~\label{fig:input_faithfulness_qwen2.5}
    \vspace{-.1in}
\end{figure*}

\begin{figure*}[!tp]
    \centering
        
        \includegraphics[width=2\columnwidth]{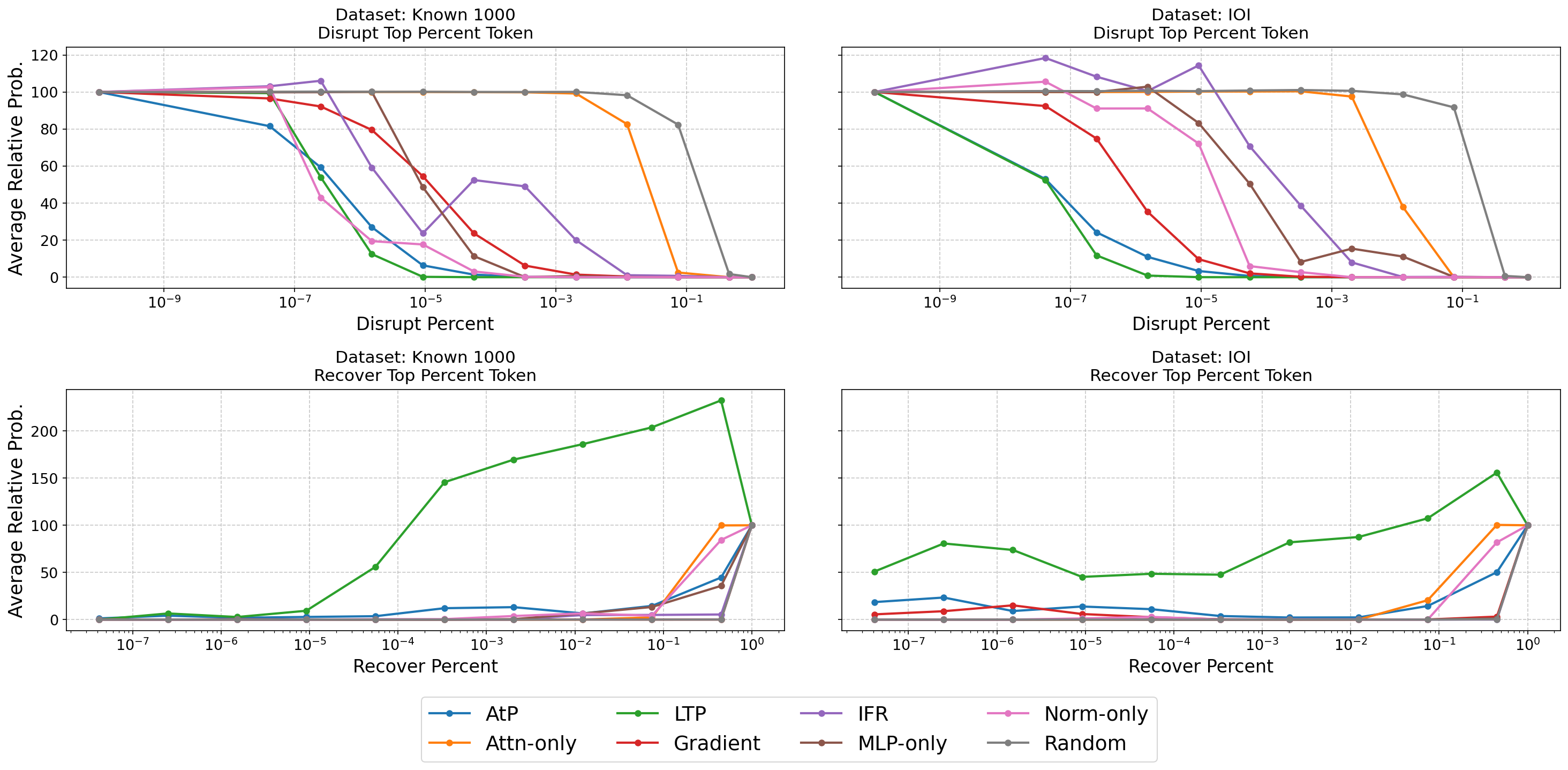}
    \caption{Performance comparison with baseline attribution methods on \textbf{Qwen2.5-32B-Instruct} under different top model component masking strategies. Our approach shows lower Top-k disruption and higher Top-k recovery, indicating more accurate identification of important components.}
    ~\label{fig:component_faithfulness_qwen2.5}
    \vspace{-.2in}
\end{figure*}

\begin{figure*}[!t]
    \centering
        
        \includegraphics[width=2\columnwidth]{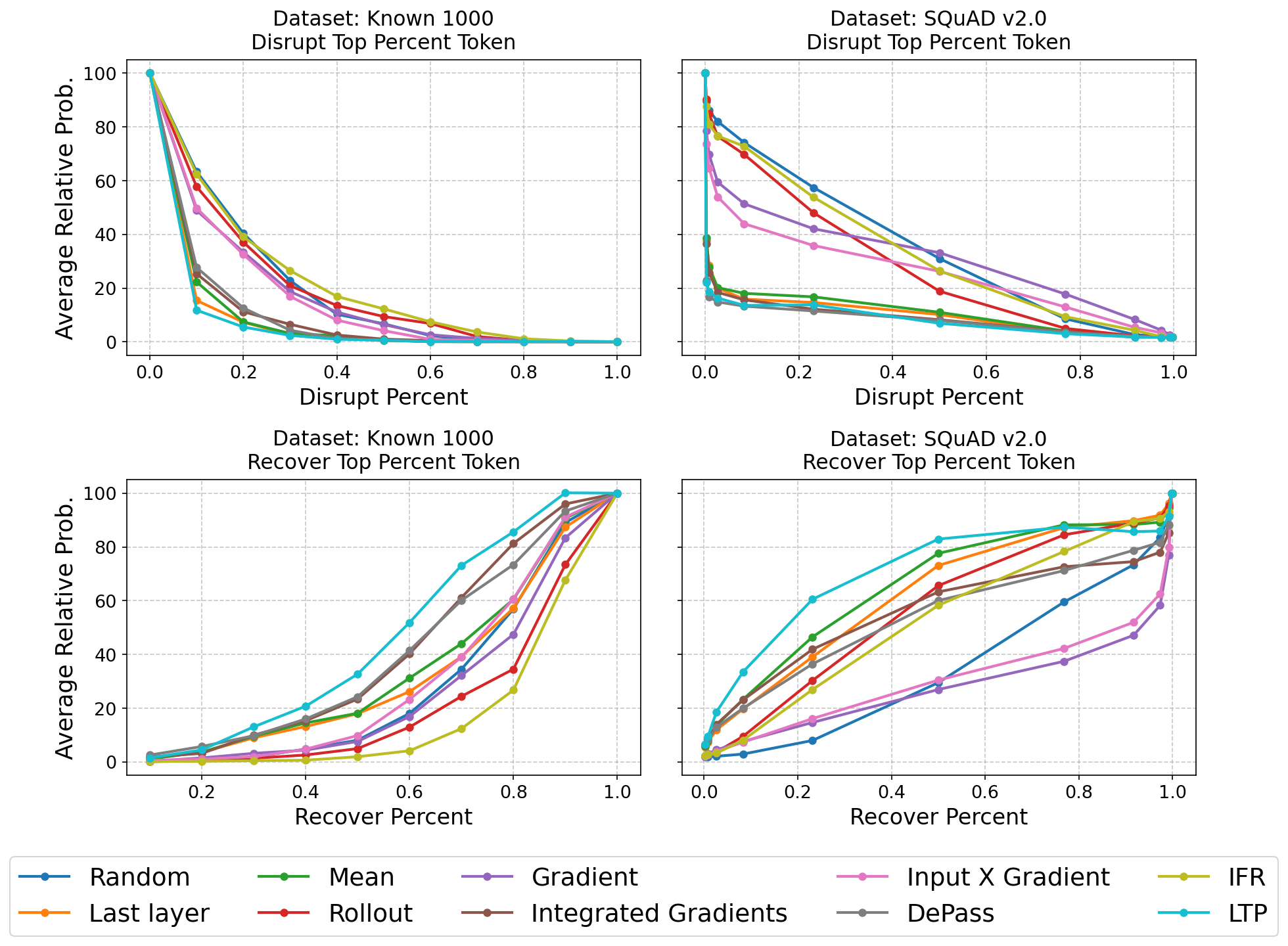}
    \vspace{-.1in}
    \caption{Performance comparison with baseline attribution methods on \textbf{Mistral-7B-Instruct-v0.3} under different top input token masking strategies. Our approach shows \emph{lower} Top-$k$ disruption and \emph{higher} Top-$k$ recovery, indicating more accurate identification of important components.}
    ~\label{fig:input_faithfulness_mistral}
    \vspace{-.1in}
\end{figure*}

\begin{figure*}[!tp]
    \centering
        \includegraphics[width=2\columnwidth]{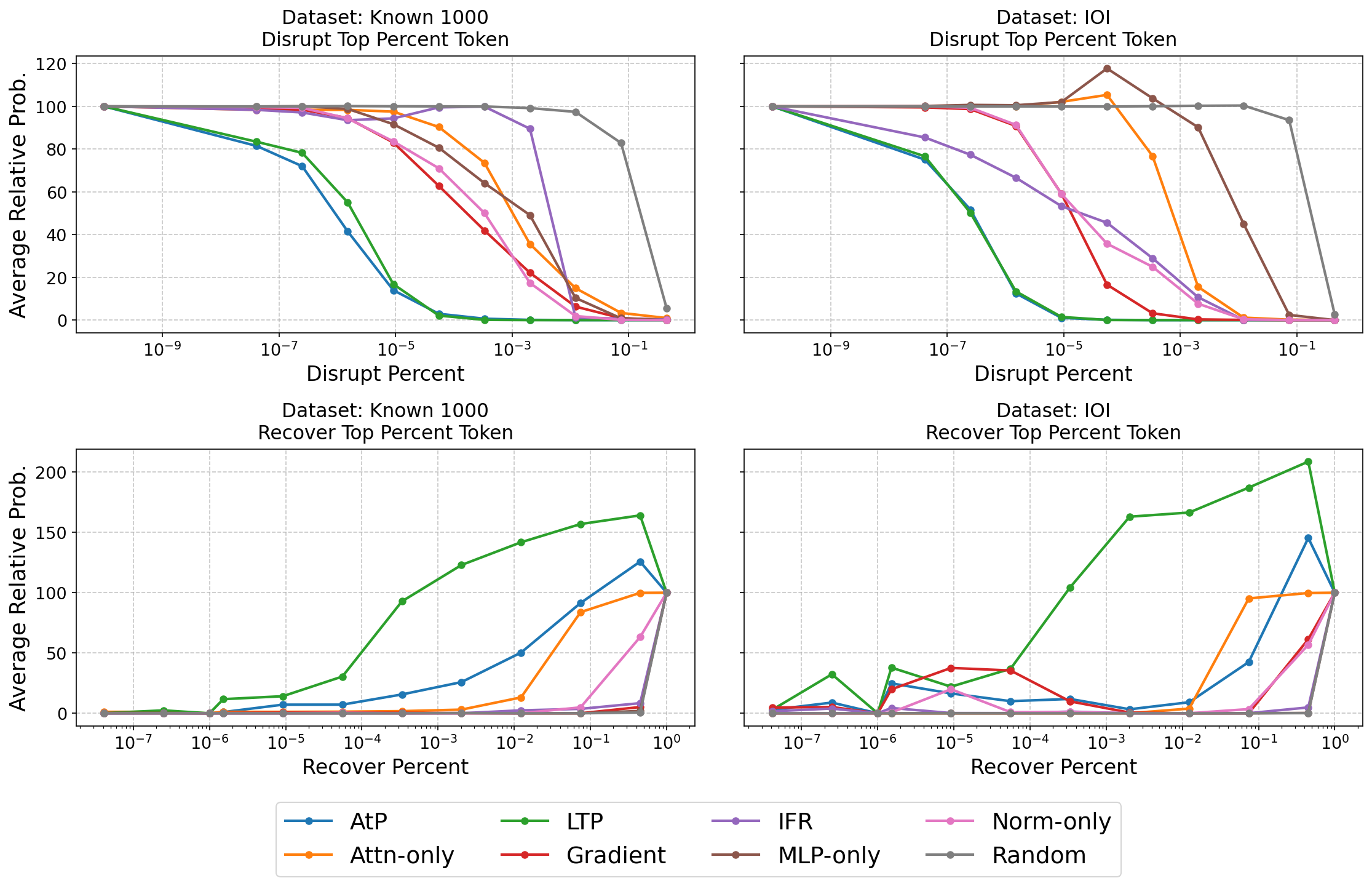}
    \caption{Performance comparison with baseline attribution methods on \textbf{Mistral-7B-Instruct-v0.3} under different top model component masking strategies. Our approach shows lower Top-k disruption and higher Top-k recovery, indicating more accurate identification of important components.}
    ~\label{fig:component_faithfulness_mistral}
    \vspace{-.2in}
\end{figure*}

We present quantitative results and visualizations for Qwen3-4B-Instruct-2507 in Tables~\ref{tab:input_faithfulness_qwen3} and \ref{tab:component_faithfulness_qwen3} (with corresponding Figures~\ref{fig:input_faithfulness_qwen3} and \ref{fig:component_faithfulness_qwen3}). Similarly, results for Qwen2.5-32B-Instruct are detailed in Tables~\ref{tab:input_faithfulness_qwen2.5} and \ref{tab:component_faithfulness_qwen2.5} (Figures~\ref{fig:input_faithfulness_qwen2.5} and \ref{fig:component_faithfulness_qwen2.5}), and for Mistral-7B-Instruct-v0.3 in Tables~\ref{tab:input_faithfulness_mistral} and \ref{tab:component_faithfulness_mistral} (Figures~\ref{fig:input_faithfulness_mistral} and \ref{fig:component_faithfulness_mistral}).
Empirical results demonstrate that LTP exhibits robust superiority over existing baseline methods across these architectural variations. By consistently achieving state-of-the-art performance in the majority of evaluations, LTP confirms its broad applicability and the stability of its performance.

\subsection{Generalization to Non-SwiGLU Architectures}~\label{app:gpt2_results}
\begin{figure*}[!t]
    \centering
        \includegraphics[width=1.8\columnwidth]{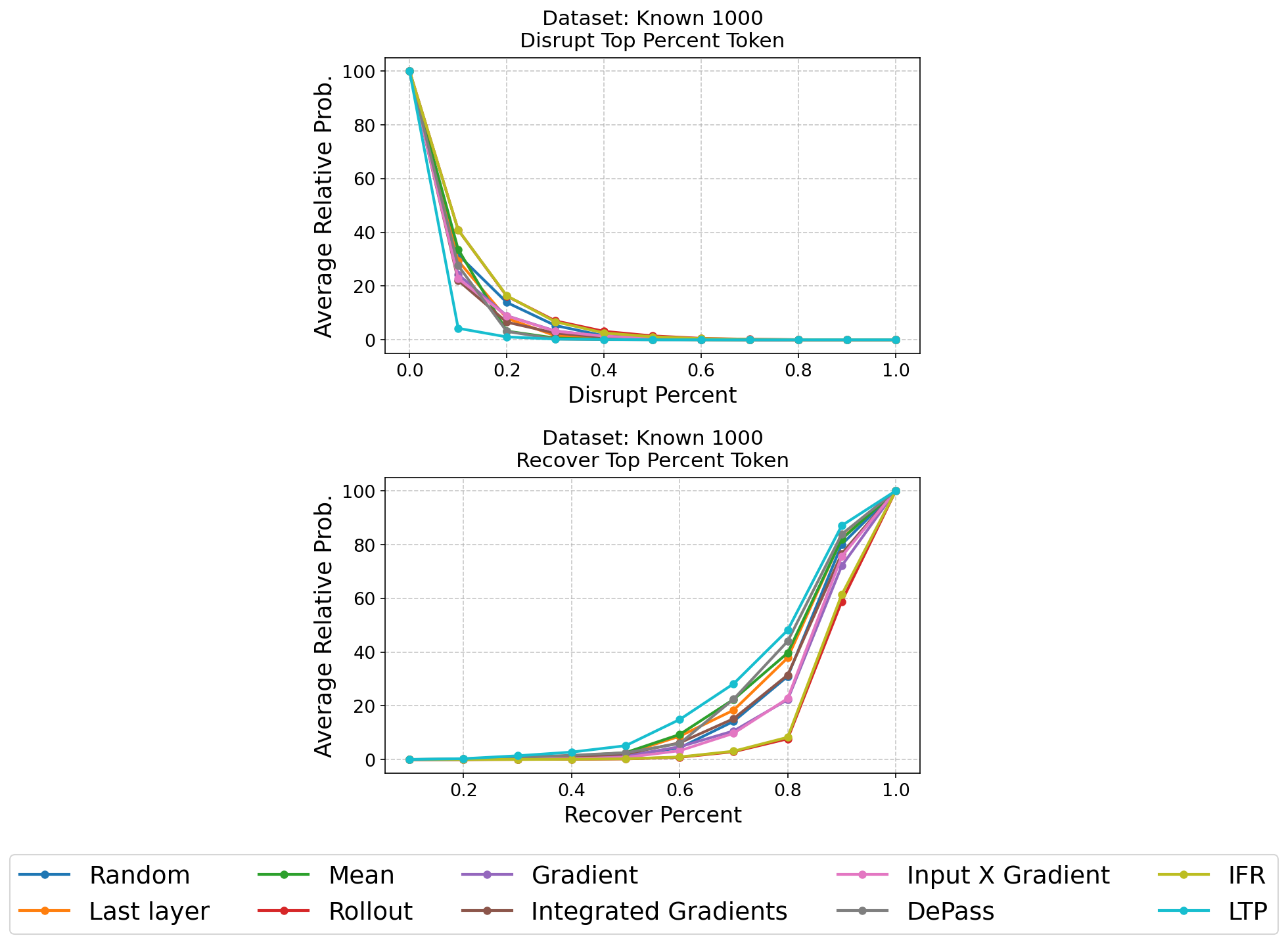}
    \caption{Performance comparison with baseline attribution methods on \textbf{GPT-2 Large} under different top input token masking strategies. Our approach shows \emph{lower} Top-$k$ disruption and \emph{higher} Top-$k$ recovery, indicating more accurate identification of important input tokens.}
    ~\label{fig:input_faithfulness_gpt2}
\end{figure*}

\begin{figure*}[!tp]
    \centering

        \includegraphics[width=2\columnwidth]{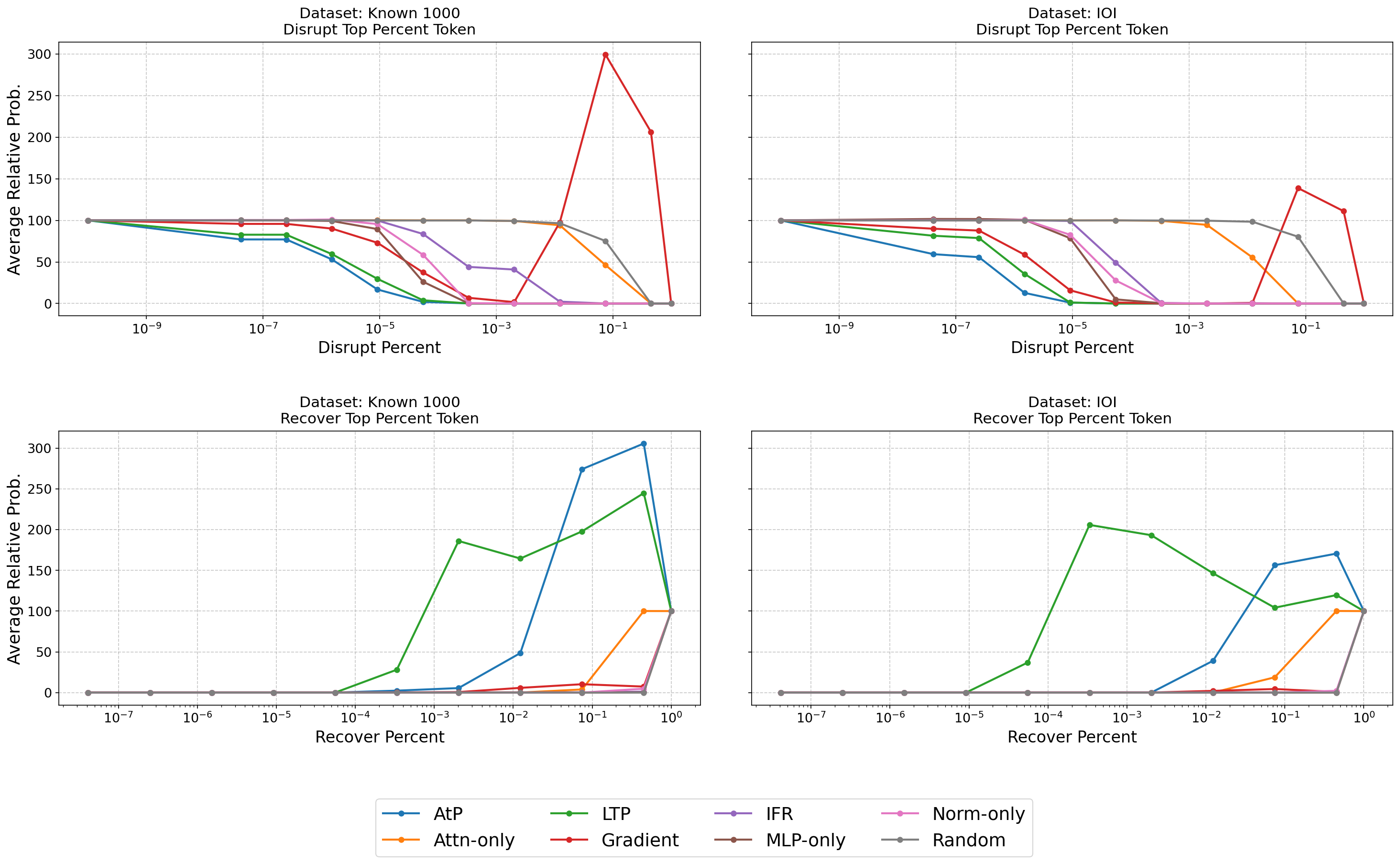}
    \caption{Performance comparison with baseline attribution methods on \textbf{GPT-2 Large} under different top model component masking strategies. LTP shows low Top-$k$ disruption and the earliest Top-$k$ recovery, indicating more accurate identification of important components.}
    ~\label{fig:component_faithfulness_gpt2}
    \vspace{-.2in}
\end{figure*}

While our derivation focuses on the SwiGLU Transformer, the underlying target-propagation framework is not tied to gated MLPs. To probe this, we additionally evaluate LTP on GPT-2 Large, a classical Transformer whose MLP block employs a GELU activation instead of a gated linear unit. Table~\ref{tab:input_faithfulness_gpt2} and Figure~\ref{fig:input_faithfulness_gpt2} report the input attribution results on Known 1000, while Table~\ref{tab:component_faithfulness_gpt2} and Figure~\ref{fig:component_faithfulness_gpt2} report the component attribution results on Known 1000 and IOI.

For input attribution, LTP achieves the best disruption, recovery, and total scores among all evaluated methods. For component attribution, AtP and LTP both attain zero disruption on both benchmarks, while AtP achieves the higher recovery AUC. Notably, however, the recovery curves in Figure~\ref{fig:component_faithfulness_gpt2} show that LTP recovers the original prediction probability with the fewest components, rising substantially earlier than AtP, whose recovery overtakes only at the larger component ratios that dominate the AUC. These results indicate that LTP transfers to non-SwiGLU architectures without sacrificing its precise localization of causally relevant components.
\begin{table*}[!t]
    \centering
    \footnotesize
    \renewcommand{\arraystretch}{0.9}

\resizebox{\linewidth}{!}{%
    \begin{tabular}{l@{\hskip 14pt}ccc@{\hskip 14pt}ccc@{\hskip 14pt}ccc}
        \toprule
        \multirow{2.5}{*}{\textbf{Methods}} 
        & \multicolumn{3}{c}{\textbf{Known 1000}} 
        & \multicolumn{3}{c}{\textbf{SQuAD v2.0}} 
        & \multicolumn{3}{c}{\textbf{IMDb}} \\
        \cmidrule(lr){2-4}\cmidrule(lr){5-7}\cmidrule(lr){8-10}
        & dis.\,$\downarrow$ & rec.\,$\uparrow$ & total\,$\uparrow$
        & dis.\,$\downarrow$ & rec.\,$\uparrow$ & total\,$\uparrow$
        & dis.\,$\downarrow$ & rec.\,$\uparrow$ & total\,$\uparrow$ \\
        \midrule
        \multicolumn{10}{c}{\textit{Reference}} \\
        \midrule
        Random
        & 13.17\nd & 20.64\nd & 7.47\nd
        & 23.74\nd & 25.04\nd & 1.30\nd
        & 58.98\nd & 57.80\nd & -1.18\nd \\
        \midrule
        \multicolumn{10}{c}{\textit{Attention-based}} \\
        \midrule
        Last layer
        & 6.53\nd & 28.85$^{\dagger}$ & 22.32$^{\dagger}$
        & 8.90\nd & 31.95\nd & 23.05\nd
        & 59.81\nd & 60.94\nd & 1.13\nd \\
        Mean
        & 5.97$^{\dagger}$ & 25.10\nd & 19.13\nd
        & 5.48$^{\dagger}$ & 33.35$^{\dagger}$ & 27.87$^{\dagger}$
        & \textbf{46.75}\nd & \textbf{78.60}\nd & \textbf{31.85}\nd \\
        Rollout
        & 15.45\nd & 16.17\nd & 0.72\nd
        & 22.64\nd & 27.35\nd & 4.71\nd
        & 83.31\nd & 55.54\nd & -27.77\nd \\
        \midrule
        \multicolumn{10}{c}{\textit{Gradient-based}} \\
        \midrule
        Gradient
        & 12.14\nd & 18.75\nd & 6.61\nd
        & 16.94\nd & 15.55\nd & -1.39\nd
        & 55.16\nd & 63.90\nd & 8.74\nd \\
        Input$\times$Gradient
        & 11.31\nd & 20.31\nd & 9.00\nd
        & 14.73\nd & 17.16\nd & 2.43\nd
        & 69.39\nd & 50.85\nd & -18.54\nd \\
        Integrated Gradients
        & 11.58\nd & 25.30\nd & 13.72\nd
        & 9.87\nd & 26.08\nd & 16.21\nd
        & 52.30\nd & 68.19\nd & 15.89\nd \\
        \midrule
        \multicolumn{10}{c}{\textit{Decomposition-based}} \\
        \midrule
        DePass
        & 7.55\nd & 26.29\nd & 18.74\nd
        & 12.62\nd & 27.49\nd & 14.87\nd
        & 68.19\nd & 56.13\nd & -12.06\nd \\
        \midrule
        \multicolumn{10}{c}{\textit{Contextual mixing-based}} \\
        \midrule
        IFR
        & 15.22\nd & 12.76\nd & -2.46\nd
        & 12.62\nd & 26.29\nd & 2.65\nd
        & 91.62\nd & 48.67\nd & -42.95\nd \\
        \midrule
        \multicolumn{10}{c}{\textit{Proposed Method}} \\
        \midrule
        \textbf{Layer-wise Target Propagation (LTP)}
        & \textbf{5.60}\nd & \textbf{34.41}\nd & \textbf{28.81}\nd
        & \textbf{3.73}\nd & \textbf{36.08}\nd & \textbf{32.35}\nd
        & 48.21$^{\dagger}$ & 74.53$^{\dagger}$ & 26.32$^{\dagger}$ \\
        \bottomrule
    \end{tabular}
}
    \caption{
    Faithfulness evaluation of crucial token attribution on \textbf{Llama-2-7B-Chat} across \textbf{Known 1000} (factual knowledge), \textbf{SQuAD v2.0} (reading comprehension), and \textbf{IMDb} (sentiment analysis).
    Lower disruption (\textit{dis.}) and higher recovery (\textit{rec.}) indicate more faithful token localization; \textit{total} denotes their aggregate AUC score.
    \textbf{Bold} indicates the best result, and $^{\dagger}$ the second best.
    Layer-wise Target Propagation consistently achieves the highest faithfulness across datasets.
    }
    \label{tab:input_faithfulness_llama2}
\end{table*}

\begin{table*}[!t]
    \centering
    \small
    \renewcommand{\arraystretch}{0.9}
    
    \begin{tabular}{l@{\hskip 36pt}ccc@{\hskip 36pt}ccc}
        \toprule
        \multirow{2.5}{*}{\textbf{Methods}} 
        & \multicolumn{3}{c}{\textbf{Known 1000}} 
        & \multicolumn{3}{c}{\textbf{IOI}} \\
        \cmidrule(lr){2-4}\cmidrule(lr){5-7}
        & dis.\,$\downarrow$ & rec.\,$\uparrow$ & total\,$\uparrow$ 
        & dis.\,$\downarrow$ & rec.\,$\uparrow$ & total\,$\uparrow$ \\
        \midrule
        \multicolumn{7}{c}{\textit{Reference}} \\
        \midrule
        Random
        & 22.89\nd & 27.65\nd & 4.76\nd
        & 24.07\nd & 27.67\nd & 3.60\nd \\
        \midrule
        \multicolumn{7}{c}{\textit{Baselines}} \\
        \midrule
        Attn-only
        & 2.34\nd & 92.58\nd & 90.24\nd
        & 0.53\nd & 91.39$^{\dagger}$ & 90.86$^{\dagger}$ \\
        MLP-only
        & 1.74\nd & 29.16\nd & 27.42\nd
        & 5.59\nd & 27.58\nd & 21.99\nd \\
        Norm-only
        & 0.24\nd & 64.68\nd & 64.44\nd
        & 0.05\nd & 66.98\nd & 66.93\nd \\
        \midrule
        \multicolumn{7}{c}{\textit{Gradient-based}} \\
        \midrule
        Gradient
        & 0.22$^{\dagger}$ & 28.21\nd & 27.99\nd
        & 0.01$^{\dagger}$ & 30.69\nd & 30.68\nd \\
        AtP
        & \textbf{0.00}\nd & 120.80$^{\dagger}$ & 120.80$^{\dagger}$
        & \textbf{0.00}\nd & 76.30\nd & 76.30\nd \\
        \midrule
        \multicolumn{7}{c}{\textit{Contextual mixing-based}} \\
        \midrule
        IFR
        & 0.69\nd & 30.49\nd & 29.80\nd
        & 0.48\nd & 28.86\nd & 28.38\nd \\
        \midrule
        \multicolumn{7}{c}{\textit{Proposed Method}} \\
        \midrule
        \textbf{Layer-wise Target Propagation (LTP)}
        & \textbf{0.00}\nd & \textbf{136.45}\nd & \textbf{136.45}\nd
        & \textbf{0.00}\nd & \textbf{127.16}\nd & \textbf{127.16}\nd \\
        \bottomrule
    \end{tabular}

    \caption{
    Component-level attribution performance on \textbf{Llama-2-7B-Chat} across \textbf{Known 1000} (factual knowledge) and \textbf{IOI} (Indirect Object Identification) benchmarks.
    Lower disruption (\textit{dis.}) and higher recovery (\textit{rec.}) indicate more faithful localization of causally relevant components; the \textit{total} score aggregates both effects.
    \textbf{Bold} denotes the best result, and $^{\dagger}$ indicates the second best.
    Layer-wise Target Propagation consistently achieves the most faithful component localization across benchmarks.
    }
    \label{tab:component_faithfulness_llama2}
\end{table*}

\begin{table*}[!t]
    \centering
    \small
    \renewcommand{\arraystretch}{0.9}
    
    \begin{tabular}{l@{\hskip 36pt}ccc@{\hskip 36pt}ccc}
        \toprule
        \multirow{2.5}{*}{\textbf{Methods}} 
        & \multicolumn{3}{c}{\textbf{Known 1000}} 
        & \multicolumn{3}{c}{\textbf{SQuAD v2.0}} \\
        \cmidrule(lr){2-4}\cmidrule(lr){5-7}
        & dis.\,$\downarrow$ & rec.\,$\uparrow$ & total\,$\uparrow$
        & dis.\,$\downarrow$ & rec.\,$\uparrow$ & total\,$\uparrow$\\
        \midrule
        \multicolumn{7}{c}{\textit{Reference}} \\
        \midrule
        Random
        & 11.52\nd & 19.64\nd & 8.12\nd
        & 48.89\nd & 50.00\nd & 1.11\nd \\
        \midrule
        \multicolumn{7}{c}{\textit{Attention-based}} \\
        \midrule
        Last layer
        & 8.03\nd & 22.82$^{\dagger}$ & 14.79\nd
        & 26.04\nd & 72.41\nd & 46.37\nd \\
        Mean
        & 9.50\nd & 22.35\nd & 12.85\nd
        & \textbf{11.69}\nd & 91.55$^{\dagger}$ & 79.86$^{\dagger}$ \\
        Rollout
        & 15.25\nd & 11.79\nd & -3.46\nd
        & 43.81\nd & 66.12\nd & 22.31\nd \\
        \midrule
        \multicolumn{7}{c}{\textit{Gradient-based}} \\
        \midrule
        Gradient
        & 10.88\nd & 18.07\nd & 7.19\nd
        & 57.82\nd & 35.49\nd & -22.33\nd \\
        Input$\times$Gradient
        & 10.54\nd & 17.95\nd & 7.41\nd
        & 46.23\nd & 47.40\nd & 1.17\nd \\
        Integrated Gradients
        & 7.55\nd & 16.18\nd & 8.63\nd
        & 41.39\nd & 49.99\nd & 8.60\nd \\
        \midrule
        \multicolumn{7}{c}{\textit{Decomposition-based}} \\
        \midrule
        DePass
        & \textbf{6.24}\nd & \textbf{26.74}\nd & \textbf{20.50}\nd
        & 13.74\nd & 66.19\nd & 52.45\nd \\
        \midrule
        \multicolumn{7}{c}{\textit{Contextual mixing-based}} \\
        \midrule
        IFR
        & 15.80\nd & 11.28\nd & -4.52\nd
        & 43.74\nd & 66.46\nd & 22.72\nd \\
        \midrule
        \multicolumn{7}{c}{\textit{Proposed Method}} \\
        \midrule
        \textbf{Layer-wise Target Propagation (LTP)}
        & 6.54$^{\dagger}$ & 22.37\nd & 15.83$^{\dagger}$
        & 11.93$^{\dagger}$ & \textbf{93.86}\nd & \textbf{81.93}\nd \\
        \bottomrule
    \end{tabular}
    \caption{
    Faithfulness evaluation of crucial token attribution on \textbf{Qwen3-4B-Instruct-2507} across \textbf{Known 1000} (factual knowledge), and \textbf{SQuAD v2.0} (reading comprehension).
    Lower disruption (\textit{dis.}) and higher recovery (\textit{rec.}) indicate more faithful token localization; \textit{total} denotes their aggregate AUC score.
    \textbf{Bold} indicates the best result, and $^{\dagger}$ the second best.
    Layer-wise Target Propagation consistently achieves the highest faithfulness across datasets.
    }
    \label{tab:input_faithfulness_qwen3}
\end{table*}

\begin{table*}[!t]
    \centering
    \small
    \renewcommand{\arraystretch}{0.9}
    
    \begin{tabular}{l@{\hskip 36pt}ccc@{\hskip 36pt}ccc}
        \toprule
        \multirow{2.5}{*}{\textbf{Methods}} 
        & \multicolumn{3}{c}{\textbf{Known 1000}} 
        & \multicolumn{3}{c}{\textbf{IOI}} \\
        \cmidrule(lr){2-4}\cmidrule(lr){5-7}
        & dis.\,$\downarrow$ & rec.\,$\uparrow$ & total\,$\uparrow$ 
        & dis.\,$\downarrow$ & rec.\,$\uparrow$ & total\,$\uparrow$ \\
        \midrule
        \multicolumn{7}{c}{\textit{Reference}} \\
        \midrule
        Random
        & 19.00\nd & 27.53\nd & 8.53\nd
        & 22.63\nd & 27.55\nd & 4.92\nd \\
        \midrule
        \multicolumn{7}{c}{\textit{Baselines}} \\
        \midrule
        Attn-only
        & 2.13\nd & 74.12$^{\dagger}$ & 71.99$^{\dagger}$
        & 0.66\nd & \textbf{83.11}\nd & \textbf{82.45}\nd \\
        MLP-only
        & 0.01$^{\dagger}$ & 28.66\nd & 28.65\nd
        & 0.01$^{\dagger}$ & 27.56\nd & 27.55\nd \\
        Norm-only
        & \textbf{0.00}\nd & 47.64\nd & 47.64\nd
        & \textbf{0.00}\nd & 28.68\nd & 28.68\nd \\
        \midrule
        \multicolumn{7}{c}{\textit{Gradient-based}} \\
        \midrule
        Gradient
        & 0.03\nd & 27.54\nd & 27.51\nd
        & \textbf{0.00}\nd & 29.54\nd & 29.54\nd \\
        AtP
        & \textbf{0.00}\nd & 41.98\nd & 41.98\nd
        & \textbf{0.00}\nd & 32.25\nd & 32.25\nd \\
        \midrule
        \multicolumn{7}{c}{\textit{Contextual mixing-based}} \\
        \midrule
        IFR
        & 0.17\nd & 38.68\nd & 38.51\nd
        & 0.02\nd & 28.59\nd & 28.57\nd \\
        \midrule
        \multicolumn{7}{c}{\textit{Proposed Method}} \\
        \midrule
        \textbf{Layer-wise Target Propagation (LTP)}
        & \textbf{0.00}\nd & \textbf{105.21}\nd & \textbf{104.35}\nd
        & \textbf{0.00}\nd & 54.91$^{\dagger}$ & 54.91$^{\dagger}$ \\
        \bottomrule
    \end{tabular}

    \caption{
    Component-level attribution performance on \textbf{Qwen3-4B-Instruct-2507} across \textbf{Known 1000} (factual knowledge) and \textbf{IOI} (Indirect Object Identification) benchmarks.
    Lower disruption (\textit{dis.}) and higher recovery (\textit{rec.}) indicate more faithful localization of causally relevant components; the \textit{total} score aggregates both effects.
    \textbf{Bold} denotes the best result, and $^{\dagger}$ indicates the second best.
    Layer-wise Target Propagation consistently achieves the most faithful component localization across benchmarks.
    }
    \label{tab:component_faithfulness_qwen3}
\end{table*}

\begin{table*}[!t]
    \centering
    \small
    \renewcommand{\arraystretch}{0.9}
    
    \begin{tabular}{l@{\hskip 72pt}ccc}
        \toprule
        \multirow{2.5}{*}{\textbf{Methods}} 
        & \multicolumn{3}{c}{\textbf{Known 1000}} \\
        \cmidrule{2-4}
        & dis.\,$\downarrow$ & rec.\,$\uparrow$ & total\,$\uparrow$\\
        \midrule
        \multicolumn{4}{c}{\textit{Reference}} \\
        \midrule
        Random
        & 20.00\nd & 29.79\nd & 9.79\nd \\
        \midrule
        \multicolumn{4}{c}{\textit{Attention-based}} \\
        \midrule
        Last layer
        & 10.63$^{\dagger}$ & 32.18\nd & 21.55$^{\dagger}$ \\
        Mean
        & 12.06\nd & 33.07$^{\dagger}$ & 21.01\nd \\
        Rollout
        & 19.70\nd & 11.37\nd & -8.33\nd \\
        \midrule
        \multicolumn{4}{c}{\textit{Gradient-based}} \\
        \midrule
        Gradient
        & 19.44\nd & 26.67\nd & 7.23\nd \\
        Input$\times$Gradient
        & 21.52\nd & 26.92\nd & 5.40\nd \\
        Integrated Gradients
        & 18.60\nd & 27.70\nd & 9.10\nd \\
        \midrule
        \multicolumn{4}{c}{\textit{Decomposition-based}} \\
        \midrule
        DePass
        & 12.29\nd & 20.53\nd & 8.24\nd \\
        \midrule
        \multicolumn{4}{c}{\textit{Contextual mixing-based}} \\
        \midrule
        IFR
        & 18.75\nd & 26.61\nd & 7.86\nd \\
        \midrule
        \multicolumn{4}{c}{\textit{Proposed Method}} \\
        \midrule
        \textbf{Layer-wise Target Propagation (LTP)}
        & \textbf{10.32}\nd & \textbf{41.35}\nd & \textbf{31.03}\nd \\
        \bottomrule
    \end{tabular}
    \caption{
    Faithfulness evaluation of crucial token attribution for \textbf{Qwen2.5-32B-Instruct} on \textbf{Known 1000} (factual knowledge).
    Lower disruption (\textit{dis.}) and higher recovery (\textit{rec.}) indicate more faithful token localization; \textit{total} denotes their aggregate AUC score.
    \textbf{Bold} indicates the best result, and $^{\dagger}$ the second best.
    Layer-wise Target Propagation consistently achieves the highest faithfulness across datasets.
    }
    \label{tab:input_faithfulness_qwen2.5}
\end{table*}

\begin{table*}[!t]
    \centering
    \small
    \renewcommand{\arraystretch}{0.9}
    
    \begin{tabular}{l@{\hskip 24pt}ccc@{\hskip 24pt}ccc}
        \toprule
        \multirow{2.5}{*}{\textbf{Methods}}
        & \multicolumn{3}{c}{\textbf{Known 1000}}
        & \multicolumn{3}{c}{\textbf{IOI}} \\
        \cmidrule(lr){2-4}\cmidrule(lr){5-7}
        & dis.\,$\downarrow$ & rec.\,$\uparrow$ & total\,$\uparrow$
        & dis.\,$\downarrow$ & rec.\,$\uparrow$ & total\,$\uparrow$ \\
        \midrule
        \multicolumn{7}{c}{\textit{Reference}} \\
        \midrule
        Random
        & 22.59\nd & 27.55\nd & 4.96\nd
        & 24.45\nd & 27.68\nd & 3.23\nd \\
        \midrule
        \multicolumn{7}{c}{\textit{Baselines}} \\
        \midrule
        Attn-only
        & 4.25\nd & 74.29\nd & 70.04\nd
        & 2.07\nd & 78.50$^{\dagger}$ & 76.43$^{\dagger}$ \\
        MLP-only
        & 0.01$^{\dagger}$ & 47.25\nd & 47.24\nd
        & 0.55\nd & 28.17\nd & 27.62\nd \\
        Norm-only
        & \textbf{0.00}\nd & 67.75$^{\dagger}$ & 67.75$^{\dagger}$
        & 0.01$^{\dagger}$ & 65.45\nd & 65.44\nd \\
        \midrule
        \multicolumn{7}{c}{\textit{Gradient-based}} \\
        \midrule
        Gradient
        & 0.05\nd & 27.56\nd & 27.51\nd
        & \textbf{0.00}\nd & 29.01\nd & 29.01\nd \\
        AtP
        & \textbf{0.00}\nd & 51.69\nd & 51.69\nd
        & \textbf{0.00}\nd & 54.06\nd & 54.06\nd \\
        \midrule
        \multicolumn{7}{c}{\textit{Contextual mixing-based}} \\
        \midrule
        IFR
        & 0.38\nd & 31.29\nd & 30.91\nd
        & 0.14\nd & 27.56\nd & 27.42\nd \\
        \midrule
        \multicolumn{7}{c}{\textit{Proposed Method}} \\
        \midrule
        \textbf{Layer-wise Target Propagation (LTP)}
        & \textbf{0.00}\nd & \textbf{187.56}\nd & \textbf{187.56}\nd
        & \textbf{0.00}\nd & \textbf{126.85}\nd & \textbf{126.85}\nd \\
        \bottomrule
    \end{tabular}

    \caption{
    Component-level attribution performance for \textbf{Qwen2.5-32B-Instruct} across \textbf{Known 1000} (factual knowledge) and \textbf{IOI} (Indirect Object Identification) benchmarks.
    Lower disruption (\textit{dis.}) and higher recovery (\textit{rec.}) indicate more faithful localization of causally relevant components; the \textit{total} score aggregates both effects.
    \textbf{Bold} denotes the best result, and $^{\dagger}$ indicates the second best.
    Layer-wise Target Propagation consistently achieves the most faithful component localization across benchmarks.
    }
    \label{tab:component_faithfulness_qwen2.5}
\end{table*}

\begin{table*}[!t]
    \centering
    \small
    \renewcommand{\arraystretch}{0.9}
    
    \begin{tabular}{l@{\hskip 36pt}ccc@{\hskip 36pt}ccc}
        \toprule
        \multirow{2.5}{*}{\textbf{Methods}} 
        & \multicolumn{3}{c}{\textbf{Known 1000}} 
        & \multicolumn{3}{c}{\textbf{SQuAD v2.0}} \\
        \cmidrule(lr){2-4}\cmidrule(lr){5-7}
        & dis.\,$\downarrow$ & rec.\,$\uparrow$ & total\,$\uparrow$
        & dis.\,$\downarrow$ & rec.\,$\uparrow$ & total\,$\uparrow$\\
        \midrule
        \multicolumn{7}{c}{\textit{Reference}} \\
        \midrule
        Random
        & 19.67\nd & 26.38\nd & 6.71\nd
        & 34.63\nd & 34.68\nd & 0.05\nd \\
        \midrule
        \multicolumn{7}{c}{\textit{Attention-based}} \\
        \midrule
        Last layer
        & 7.80$^{\dagger}$ & 30.41\nd & 22.61\nd
        & 9.79\nd & 62.84\nd & 53.05\nd \\
        Mean
        & 8.61\nd & 32.27\nd & 23.66\nd
        & 10.78\nd & 65.99$^{\dagger}$ & 55.21$^{\dagger}$ \\
        Rollout
        & 19.79\nd & 20.46\nd & 0.67\nd
        & 27.94\nd & 56.88\nd & 28.94\nd \\
        \midrule
        \multicolumn{7}{c}{\textit{Gradient-based}} \\
        \midrule
        Gradient
        & 17.31\nd & 24.66\nd & 7.35\nd
        & 31.19\nd & 27.56\nd & -3.63\nd \\
        Input$\times$Gradient
        & 16.31\nd & 28.18\nd & 11.87\nd
        & 25.65\nd & 30.44\nd & 4.79\nd \\
        Integrated Gradients
        & 9.72\nd & 38.14$^{\dagger}$ & 28.42$^{\dagger}$
        & 8.67\nd & 55.98\nd & 47.31\nd \\
        \midrule
        \multicolumn{7}{c}{\textit{Decomposition-based}} \\
        \midrule
        DePass
        & 9.64\nd & 37.53\nd & 27.89\nd
        & \textbf{7.92}\nd & 53.93\nd & 46.01\nd \\
        \midrule
        \multicolumn{7}{c}{\textit{Contextual mixing-based}} \\
        \midrule
        IFR
        & 21.97\nd & 16.43\nd & -5.54\nd
        & 32.56\nd & 52.78\nd & 20.22\nd \\
        \midrule
        \multicolumn{7}{c}{\textit{Proposed Method}} \\
        \midrule
        \textbf{Layer-wise Target Propagation (LTP)}
        & \textbf{7.16}\nd & \textbf{43.23}\nd & \textbf{36.07}\nd
        & 8.06$^{\dagger}$ & \textbf{70.97}\nd & \textbf{62.91}\nd \\
        \bottomrule
    \end{tabular}
    \caption{
    Faithfulness evaluation of crucial token attribution on \textbf{Mistral-7B-Instruct-v0.3} across \textbf{Known 1000} (factual knowledge), and \textbf{SQuAD v2.0} (reading comprehension).
    Lower disruption (\textit{dis.}) and higher recovery (\textit{rec.}) indicate more faithful token localization; \textit{total} denotes their aggregate AUC score.
    \textbf{Bold} indicates the best result, and $^{\dagger}$ the second best.
    Layer-wise Target Propagation consistently achieves the highest faithfulness across datasets.
    }
    \label{tab:input_faithfulness_mistral}
\end{table*}

\begin{table*}[!t]
    \centering
    \small
    \renewcommand{\arraystretch}{0.9}
    
    \begin{tabular}{l@{\hskip 36pt}ccc@{\hskip 36pt}ccc}
        \toprule
        \multirow{2.5}{*}{\textbf{Methods}} 
        & \multicolumn{3}{c}{\textbf{Known 1000}} 
        & \multicolumn{3}{c}{\textbf{IOI}} \\
        \cmidrule(lr){2-4}\cmidrule(lr){5-7}
        & dis.\,$\downarrow$ & rec.\,$\uparrow$ & total\,$\uparrow$ 
        & dis.\,$\downarrow$ & rec.\,$\uparrow$ & total\,$\uparrow$ \\
        \midrule
        \multicolumn{7}{c}{\textit{Reference}} \\
        \midrule
        Random
        & 23.39\nd & 27.84\nd & 4.45\nd
        & 25.27\nd & 27.60\nd & 2.33\nd \\
        \midrule
        \multicolumn{7}{c}{\textit{Baselines}} \\
        \midrule
        Attn-only
        & 1.76\nd & 92.60\nd & 90.84\nd
        & 0.30\nd & 94.67\nd & 94.37\nd \\
        MLP-only
        & 0.94\nd & 28.44\nd & 27.50\nd
        & 2.82\nd & 27.66\nd & 24.84\nd \\
        Norm-only
        & 0.26\nd & 57.87\nd & 57.61\nd
        & 0.10\nd & 54.59\nd & 54.49\nd \\
        \midrule
        \multicolumn{7}{c}{\textit{Gradient-based}} \\
        \midrule
        Gradient
        & 0.64\nd & 29.94\nd & 29.30\nd
        & 0.02$^{\dagger}$ & 56.00\nd & 55.98\nd \\
        AtP
        & 0.01$^{\dagger}$ & 107.73$^{\dagger}$ & 107.72$^{\dagger}$
        & \textbf{0.00}\nd & 104.50$^{\dagger}$ & 104.50$^{\dagger}$ \\
        \midrule
        \multicolumn{7}{c}{\textit{Contextual mixing-based}} \\
        \midrule
        IFR
        & 0.91\nd & 32.31\nd & 31.40\nd
        & 0.13\nd & 29.81\nd & 29.68\nd \\
        \midrule
        \multicolumn{7}{c}{\textit{Proposed Method}} \\
        \midrule
        \textbf{Layer-wise Target Propagation (LTP)}
        & \textbf{0.00}\nd & \textbf{143.76}\nd & \textbf{143.76}\nd
        & \textbf{0.00}\nd & \textbf{172.15}\nd & \textbf{172.15}\nd \\
        \bottomrule
    \end{tabular}

    \caption{
    Component-level attribution performance on \textbf{Mistral-7B-Instruct-v0.3} across \textbf{Known 1000} (factual knowledge) and \textbf{IOI} (Indirect Object Identification) benchmarks.
    Lower disruption (\textit{dis.}) and higher recovery (\textit{rec.}) indicate more faithful localization of causally relevant components; the \textit{total} score aggregates both effects.
    \textbf{Bold} denotes the best result, and $^{\dagger}$ indicates the second best.
    Layer-wise Target Propagation consistently achieves the most faithful component localization across benchmarks.
    }
    \label{tab:component_faithfulness_mistral}
\end{table*}

\begin{table*}[!t]
    \centering
    \small
    \renewcommand{\arraystretch}{0.9}

    \begin{tabular}{l@{\hskip 64pt}ccc}
        \toprule
        \multirow{2.5}{*}{\textbf{Methods}}
        & \multicolumn{3}{c}{\textbf{Known 1000}} \\
        \cmidrule{2-4}
        & dis.\,$\downarrow$ & rec.\,$\uparrow$ & total\,$\uparrow$ \\
        \midrule
        \multicolumn{4}{c}{\textit{Reference}} \\
        \midrule
        Random
        & 10.31\nd & 18.19\nd & 7.88\nd \\
        \midrule
        \multicolumn{4}{c}{\textit{Attention-based}} \\
        \midrule
        Last layer
        & 8.95\nd & 20.05\nd & 11.09\nd \\
        Mean
        & 8.80\nd & 20.87\nd & 12.07\nd \\
        Rollout
        & 11.97\nd & 12.08\nd & 0.12\nd \\
        \midrule
        \multicolumn{4}{c}{\textit{Gradient-based}} \\
        \midrule
        Gradient
        & 8.87\nd & 16.21\nd & 7.34\nd \\
        Input$\times$Gradient
        & 8.72\nd & 16.30\nd & 7.57\nd \\
        Integrated Gradients
        & 8.29\nd & 18.34\nd & 10.05\nd \\
        \midrule
        \multicolumn{4}{c}{\textit{Decomposition-based}} \\
        \midrule
        DePass
        & 8.11$^{\dagger}$ & 21.16$^{\dagger}$ & 13.05$^{\dagger}$ \\
        \midrule
        \multicolumn{4}{c}{\textit{Contextual mixing-based}} \\
        \midrule
        IFR
        & 11.84\nd & 12.43\nd & 0.59\nd \\
        \midrule
        \multicolumn{4}{c}{\textit{Proposed Method}} \\
        \midrule
        \textbf{Layer-wise Target Propagation (LTP)}
        & \textbf{5.58}\nd & \textbf{23.81}\nd & \textbf{18.23}\nd \\
        \bottomrule
    \end{tabular}

    \caption{
    Faithfulness evaluation of crucial token attribution on \textbf{GPT-2 Large} on the \textbf{Known 1000} (factual knowledge) benchmark.
    Lower disruption (\textit{dis.}) and higher recovery (\textit{rec.}) indicate more faithful token localization; \textit{total} denotes their aggregate AUC score.
    \textbf{Bold} indicates the best result, and $^{\dagger}$ the second best.
    Layer-wise Target Propagation achieves the highest faithfulness across all metrics.
    }
    \label{tab:input_faithfulness_gpt2}
\end{table*}

\begin{table*}[!t]
    \centering
    \small
    \renewcommand{\arraystretch}{0.9}

    \begin{tabular}{l@{\hskip 24pt}ccc@{\hskip 24pt}ccc}
        \toprule
        \multirow{2.5}{*}{\textbf{Methods}}
        & \multicolumn{3}{c}{\textbf{Known 1000}}
        & \multicolumn{3}{c}{\textbf{IOI}} \\
        \cmidrule(lr){2-4}\cmidrule(lr){5-7}
        & dis.\,$\downarrow$ & rec.\,$\uparrow$ & total\,$\uparrow$
        & dis.\,$\downarrow$ & rec.\,$\uparrow$ & total\,$\uparrow$ \\
        \midrule
        \multicolumn{7}{c}{\textit{Reference}} \\
        \midrule
        Random
        & 20.66\nd & 27.54\nd & 6.88\nd
        & 21.83\nd & 27.57\nd & 5.75\nd \\
        \midrule
        \multicolumn{7}{c}{\textit{Baselines}} \\
        \midrule
        Attn-only
        & 14.25\nd & 74.69\nd & 60.44\nd
        & 2.70\nd & 77.97\nd & 75.28\nd \\
        MLP-only
        & 0.02\nd & 27.70\nd & 27.68\nd
        & 0.02\nd & 27.54\nd & 27.52\nd \\
        Norm-only
        & 0.01$^{\dagger}$ & 29.75\nd & 29.74\nd
        & 0.01$^{\dagger}$ & 28.55\nd & 28.54\nd \\
        \midrule
        \multicolumn{7}{c}{\textit{Gradient-based}} \\
        \midrule
        Gradient
        & 107.68\nd & 33.41\nd & -74.26\nd
        & 51.18\nd & 28.98\nd & -22.21\nd \\
        AtP
        & \textbf{0.00}\nd & \textbf{230.65}\nd & \textbf{230.65}\nd
        & \textbf{0.00}\nd & \textbf{142.03}\nd & \textbf{142.03}\nd \\
        \midrule
        \multicolumn{7}{c}{\textit{Contextual mixing-based}} \\
        \midrule
        IFR
        & 0.39\nd & 28.11\nd & 27.72\nd
        & 0.03\nd & 28.06\nd & 28.03\nd \\
        \midrule
        \multicolumn{7}{c}{\textit{Proposed Method}} \\
        \midrule
        \textbf{Layer-wise Target Propagation (LTP)}
        & \textbf{0.00}\nd & 191.11$^{\dagger}$ & 191.11$^{\dagger}$
        & \textbf{0.00}\nd & 112.31$^{\dagger}$ & 112.31$^{\dagger}$ \\
        \bottomrule
    \end{tabular}

    \caption{
    Component-level attribution performance for \textbf{GPT-2 Large} across \textbf{Known 1000} (factual knowledge) and \textbf{IOI} (Indirect Object Identification) benchmarks.
    Lower disruption (\textit{dis.}) and higher recovery (\textit{rec.}) indicate more faithful localization of causally relevant components; the \textit{total} score aggregates both effects.
    \textbf{Bold} denotes the best result, and $^{\dagger}$ indicates the second best.
    While AtP attains the higher recovery AUC, LTP matches its zero disruption and recovers the original prediction probability with the fewest components (cf. Figure~\ref{fig:component_faithfulness_gpt2}).
    }
    \label{tab:component_faithfulness_gpt2}
\end{table*}

\end{document}